\documentclass[nohyperref]{article}

\usepackage{microtype}
\usepackage{graphicx}
\usepackage{subfigure}
\usepackage{booktabs} 
\usepackage[math]{alp}

\usepackage{makecell}
\usepackage{pifont}
\usepackage{hyperref}

\usepackage[accepted]{icml2023}

\usepackage{multirow}

\usepackage{amsmath}
\usepackage{amssymb}
\usepackage{mathtools}
\usepackage{amsthm}
\usepackage{graphics, setspace, calc}
\usepackage{abstract}
\usepackage{enumitem}
\usepackage{paralist}
\usepackage{bbm}
\usepackage[makeroom]{cancel}

\usepackage[capitalize,noabbrev]{cleveref}

\theoremstyle{plain}

\theoremstyle{definition}

\theoremstyle{remark}

\usepackage[textsize=tiny]{todonotes}

\definecolor{sky}{RGB}{0, 230, 230}

\icmltitlerunning{Efficient and Degree-Guided Graph Generation via Discrete Diffusion Modeling}

\begin{document}

\twocolumn[
\icmltitle{Efficient and Degree-Guided Graph Generation via Discrete Diffusion Modeling}




\begin{icmlauthorlist}
\icmlauthor{Xiaohui Chen}{tufts}
\icmlauthor{Jiaxing He}{tufts}
\icmlauthor{Xu Han}{tufts}
\icmlauthor{Li-Ping Liu}{tufts}
\end{icmlauthorlist}

\icmlaffiliation{tufts}{Department of Computer Science, Tufts University, Medford, MA, USA}
\icmlcorrespondingauthor{Xiaohui Chen}{xiaohui.chen@tufts.edu}
\icmlcorrespondingauthor{Li-Ping Liu}{liping.liu@tufts.edu}

\icmlkeywords{Machine Learning, ICML}

\vskip 0.3in
]



\printAffiliationsAndNotice{}  

\begin{abstract}

Diffusion-based graph generative models are effective in generating high-quality small graphs. However, it is hard to scale them to large graphs that contain thousands of nodes. In this work, we propose EDGE, a new diffusion-based graph generative model that addresses generative tasks for large graphs. The model is developed by reversing a discrete diffusion process that randomly removes edges until obtaining an empty graph.  It leverages graph sparsity in the diffusion process to improve computational efficiency. In particular, EDGE only focuses on a small portion of graph nodes and only adds edges between these nodes. Without compromising modeling ability, it makes much fewer edge predictions than previous diffusion-based generative models. Furthermore, EDGE can explicitly model the node degrees of training graphs and then gain performance improvement in capturing graph statistics. The empirical study shows that EDGE is much more efficient than competing methods and can generate large graphs with thousands of nodes. It also outperforms baseline models in generation quality: graphs generated by the proposed model have graph statistics more similar to those of training graphs.

\end{abstract}
\section{Introduction}

There is a long history of using random graph models \citep{newman2002random} to model large graphs. Traditional models such as \erdosrenyi~(ER) model~\citep{erdos1960evolution}, Stochastic-Block Model~(SBM)~\citep{holland1983stochastic}, and Exponential-family Random Graph Models~\citep{lusher2013exponential} are often used to model existing graph data and focus on prescribed graph structures. Besides modeling existing data, one interesting problem is to generate new graphs to simulate existing ones \citep{ying2009graph}, which has applications such as network data sharing. In generative tasks~\citep{chakrabarti2006graph}, traditional models often fall short in describing complex structures. A promising direction is to use deep neural models to generate large graphs.

There are only a few deep generative models designed for generating large graphs: NetGAN ~\citep{bojchevski2018netgan} and CELL~ \citep{rendsburg2020netgan} are two examples. However, recent research~\citep{chanpuriya2021power} shows that these two models are edge-independent models and have a theoretical limitation: they cannot reproduce several important statistics (e.g. triangle counts and clustering coefficient) in their generated graphs unless they memorize the training graph. A list of other models \citep{chanpuriya2021power} including Variational Graph Autoencoders (VGAE) \citep{kipf2016variational} and GraphVAE \citep{simonovsky2018graphvae} are also edge-independent models and share the same limitation.  

Diffusion-based generative models~\citep{liu2019graph, niu2020permutation, jo2022score, chen2022nvdiff} have gained success in modeling small graphs. These models generate a graph in multiple steps and are NOT edge-independent because edges generated in later steps depend on previously generated edges. They are more flexible than one-shot models \citep{kipf2016variational,madhawa2019graphnvp,lippe2020categorical}, which directly predict an adjacency matrix in one step. They also have an advantage over auto-regressive graph models~\citep{you2018graphrnn, liao2019efficient}, as diffusion-based models are invariant to node permutations and do not have long-term memory issues. However, diffusion-based models are only designed for tasks with small graphs (usually with less than one hundred nodes). 

This work aims to scale diffusion-based generative models to large graphs. The major issue of a diffusion-based model is that it must compute a latent vector or a probability for each node pair in a graph at each diffusion step~\citep{niu2020permutation, jo2022score} -- the computation cost is $O(T N^2)$ if the model generates a graph with $N$ nodes using $T$ steps. The learning task becomes challenging when $N$ is large. At the same time, large graphs increase the difficulties for a model to capture global graph statistics such as clustering coefficients. As a result, the model performance degrades when the training graphs' sizes scale up. 

We propose \textit{\underline{E}fficient and \underline{D}egree-guided graph \underline{GE}nerative model} ({EDGE}) based on a discrete diffusion process. The development of EDGE has three innovations. First, we encourage the sparsity of graphs in the diffusion process by setting the empty graph as the convergent ``distribution''. Then the diffusion process only removes edges and can be viewed as an edge-removal process. The increased sparsity in graphs in the process dramatically reduces the computation -- this is because the message-passing neural network (MPNN)~\citep{kipf2016semi} used in the generative model needs to run on these graphs, and their runtime is linear in the number of edges.  
Second, the generative model, which is the reverse of the edge-removal process, only predicts edges for a small portion of ``active nodes'' that have edge changes in the original edge-removal process. This strategy decreases the number of predictions of MPNN and also its computation time. More importantly, this new design is naturally derived from the aforementioned edge-removal process without modifying its forward transition probabilities. Third, we model the node degrees of training graphs explicitly. By characterizing the node degrees, the statistics of the generated graphs are much closer to training graphs. While other diffusion-based graph models struggle to even train or sample on large graphs, our approach can efficiently generate large graphs with desired statistical properties. We summarize our contributions as follows:
\begin{compactitem}
    \item we use empty graphs as the convergent distribution in a discrete diffusion process to reduce computation;
    \item we propose a new generative process that only predicts edges between a fraction of nodes in graphs;
    \item we explicitly model node degrees in the probabilistic framework to improve graph statistics of generated graphs; and 
    \item we conduct an extensive empirical study and show that our method can efficiently generate large graphs with desired statistics.
\end{compactitem}

\begin{figure*}[t]
    \centering
    \includegraphics[width=\textwidth]{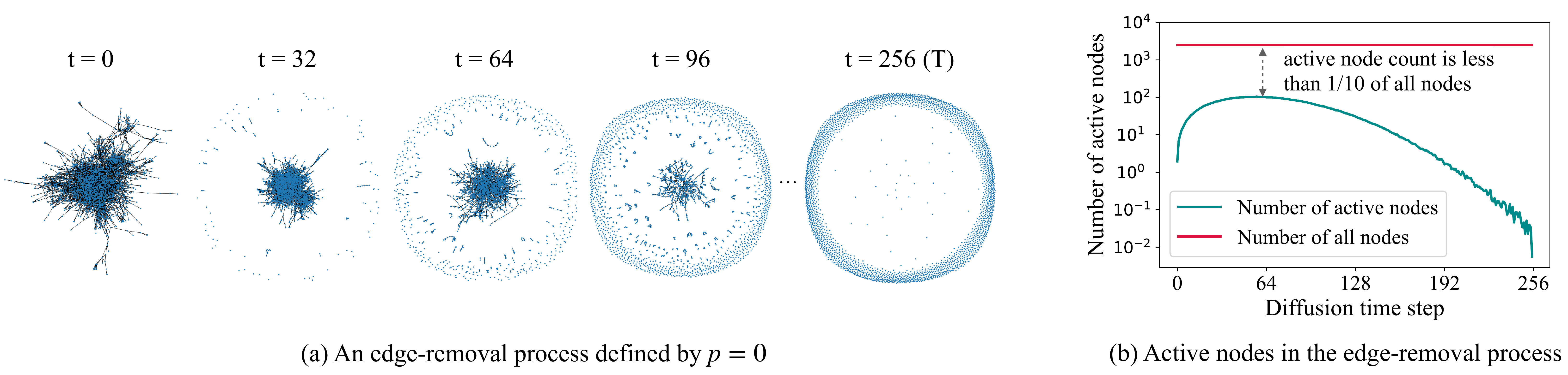}
    \vspace{-1.2em}
    \caption{Dynamics of a discrete diffusion process with $p=0$ and ``active'' nodes in the process on the Cora dataset: (a) the diffusion process with $p=0$ is an edge-removal process. The reverse of it is a generative procedure that constructs a graph by gradually adding edges to an empty graph. (b) under linear noise scheduling, the number of ``active'' nodes (that have their edges removed at a time step) is less than one-tenth of the total number of nodes.}
    \label{fig:degree_change}
\end{figure*}

\vspace{-0.5em}
\section{Background}

This work considers graph generative models that sample adjacency matrices to generate graphs. Let $\calA^N$ denote the space of adjacency matrices of size $N$. We consider simple graphs without self-loops or multi-edges, so an adjacency matrix $\bA\in\calA^N$ is a binary symmetric matrix with a zero diagonal. A generative model defines a distribution over $\calA^N$.  

In this work, we construct a generative model based on a discrete diffusion process \citep{austin2021structured,hoogeboom2021argmax,vignac2022digress}. Let $\bA^0$ denote a graph from the data, then the diffusion process defined by $q(\bA^t|\bA^{t-1})$ corrupts $\bA^0$ in $T$ steps and forms a trajectory $(\bA^0, \bA^1, \ldots, \bA^T)$. We treat $(\bA^{1}, \ldots, \bA^T)$ as latent variables, then $q(\bA^{1}, \ldots, \bA^T | \bA^0) = \prod_{t=1}^T q(\bA^t|\bA^{t-1})$. As $T \rightarrow \infty$, $q(\bA^T)$ approaches a convergent distribution, which is often a simple one with easy samples. We often choose a large enough $T$ so that $q(\bA^{T})$ is a good approximation of the convergent distribution. 

We model these trajectories with a denoising model $p_\theta(\bA^{t-1} | \bA^{t})$ parameterized by $\theta$, then the model has a joint $p_\theta(\bA^{0:T}) = p(\bA^T)\prod_{t=1}^T p_\theta(\bA^{t-1}|\bA^t)$ and a marginal $p_\theta(\bA^0)$ that describes the data distribution. Here $p(\bA^T)$ is the convergent distribution in $q$.  

Usually $q(\bA^t|\bA^{t-1})$ needs easy probability calculations. One choice is to treat each edge independently, and 
\begin{align}
    q(\bA^t|\bA^{t-1}) &= \prod_{i,j: i<j} \calB(\bA^t_{i,j}; (1-\beta_t) \bA^{t-1}_{i,j} + \beta_t p) \label{eq:forward-kernel}\\
     &:= \calB(\bA^t;(1-\beta_t)\bA^{t-1} + \beta_t p). \nonumber 
\end{align}
Here $\calB(x; \mu)$ represents the Bernoulli distribution over $x$ with probability $\mu$. We also use $\calB(\bA; \bmu)$ to represent the probability of independent Bernoulli variables arranged in a matrix. The diffusion rate $\beta_t$ determines the probability of resampling the entry $\bA^t_{i,j}$ from a Bernoulli distribution with probability $p$, instead of keeping the entry $\bA^{t-1}_{i,j}$. 

This diffusion process requires two special properties for model fitting. First, we can sample $\bA^t$ at any time step $t$ directly from $\bA^0$. Let $\alpha_t = 1-\beta_t$ and $\bar{\alpha}_t=\prod_{\tau=1}^t\alpha_\tau$,
\begin{align}
    q(\bA^t|\bA^0) = \calB(\bA^t;\bar{\alpha}_{t}\bA^0 + (1-\bar{\alpha}_{t}) p).
\end{align}
The diffusion rates $\beta_t$-s are defined in a way such that $\bar{\alpha}_T$ is almost $0$, then $\bA^T$ is almost independent from $\bA^0$, i.e., $q(\bA^T|\bA^0) \approx p(\bA^T) \equiv \calB(\bA^T;p)$. The configuration of $\beta_t$-s is called \textit{noise scheduling}. In the context of graph generation, $p(\bA^T)$ is the \erdosrenyi~graph model $G(N, p)$~\citep{erdos1960evolution},  with $p$ being the probability of forming an edge between two nodes. 

Second, we can compute the posterior of the forward transition when conditioning on $\bA^0$: 
\begin{align}
    q(\bA^{t-1}|\bA^t,\bA^0)=\frac{q(\bA^{t}|\bA^{t-1})q(\bA^{t-1}|\bA^0)}{q(\bA^{t}|\bA^0)}.\label{eq:post}
\end{align}
Since all the terms on the right-hand side are known, the posterior can be computed analytically.

The generative model $p_\theta(\bA^{0:T})$ is trained by maximizing a variational lower bound of $\log p_\theta(\bA^0)$~\citep{ho2020denoising, hoogeboom2021argmax,austin2021structured}. In an intuitive understanding, $p_\theta(\bA^{t-1}|\bA^t)$ is learned to match the posterior of the forward transition$q(\bA^{t-1}|\bA^t,\bA^0)$. 

During generation, we sample $\bA^T \sim p(\bA^T)$ and then ``denoise'' it iteratively with $p_\theta(\bA^{t-1}|\bA^t)$ to get an $\bA^0$ sample.

\vspace{-0.5em}
\section{Method}
\vspace{-0.2em}
\subsection{Diffuse graphs to empty graphs -- a motivation} 
\label{sec:g(n,0)}
\vspace{-0.2em}
With the main purpose of computation efficiency, we advocate setting $p=0$ and using $G(N, 0)$ as the convergent distribution. This configuration improves the sparsity of the adjacency matrices in diffusion trajectories, thus reducing computation. We consider the amount of computation in the denoising model $p_\theta(\bA^{t-1} | \bA^{t})$  from two aspects: the computation on the input $\bA^t$ and the number of entries to be predicted in the output $\bA^{t-1}$.

We first consider the computation on the input side. We assume that the denoising model $p_\theta(\bA^{t-1} | \bA^{t})$ is constructed with an MPNN. Suppose the input graph $\bA^{t}$ has $M^t$ edges, then a typical MPNN needs to perform $O(M^t)$ message-passing operations to compute node vectors -- here we treat hidden sizes and the number of network layers as constants. The total number of message-passing operations over the trajectory is $O(\sum_{t=1}^T M^t)$. After some calculations, we show that
\begin{align}
    \sum_{t=1}^T M^t = M^0\sum_{t=1}^T \bar{\alpha}_t + \frac{N(N-1)p}{2}\sum_{t=1}^T 1-\bar{\alpha}_t.
\end{align}
By setting $p=0$, we eliminate the second term and reduce the number of edges in graphs in the diffusion trajectory by a significant factor, then the MPNN will have much fewer message-passing operations. 

We then analyze the number of entries we need to predict in the output $\bA^{t-1}$. When $p=0$, the forward process is an edge-removal process, and the degree of a node is non-increasing for any forward transition. A node with a degree change from $t-1$ to $t$ is considered ``active''. When a node is inactive at $t-1$, all edges incident to this node is kept at $t$. Figure~\ref{fig:degree_change} shows the average number of active nodes for each forward transition. We  observe that active nodes only take a small fraction of the total when the convergent distribution is $G(N,0)$. 

While a previous diffusion-based model makes predictions for all node pairs, the observation above indicates that we can save computation by making predictions only for pairs of active nodes. In particular, the denoising model can first infer which nodes are active in each step and then only predict edges between active nodes. Below we will develop such a model and only consider the diffusion process with $p = 0$.

\begin{figure*}[t]
\centering
    \includegraphics[width=\textwidth]{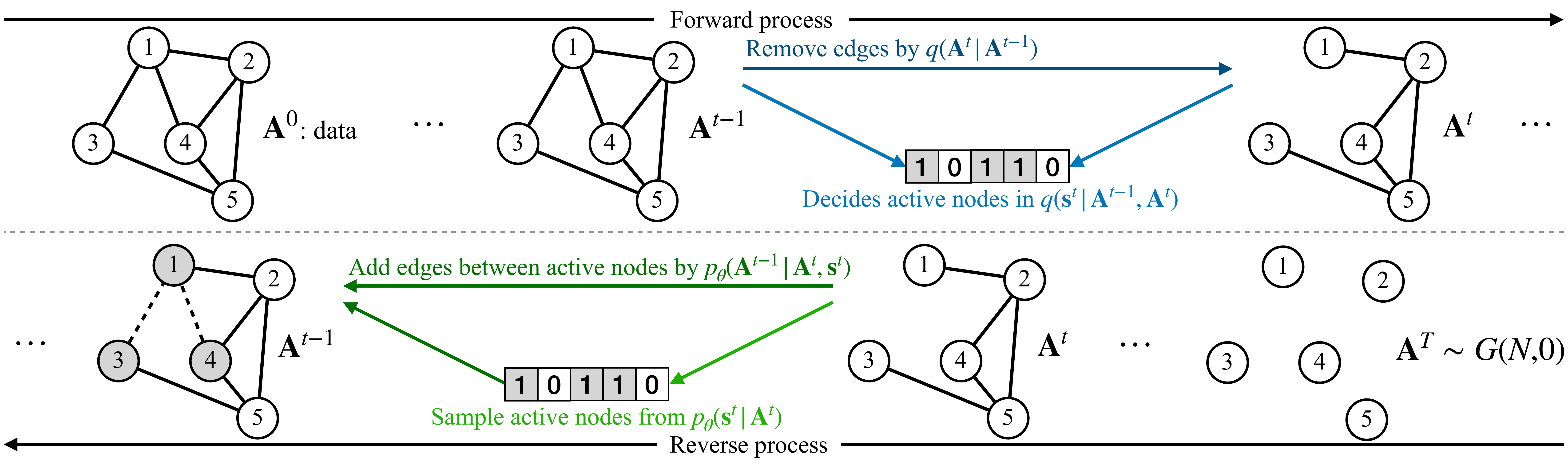}
\vspace{-2em}
    \caption{Forward and reverse processes. For the forward process, $\bA^t$ is sampled from $q(\bA^t|\bA^{t-1})$, then $\bs^t$ is deterministically generated given $\bA^{t-1}$ and $\bA^t$. For the reverse process, $\bs^t$ is first sampled from a node selection distribution $p_\theta(\bs^t|\bA^t)$, then $\bA^{t-1}$ is sampled from the parameterized distribution $p_\theta(\bA^{t-1}|\bA^t, \bs^t)$.}
\vspace{-1.5em}
    \label{fig:framework}
\end{figure*}

\subsection{A diffusion-based model that explicitly models active nodes}
\label{sec:framework}
We treat the ``active nodes'' as latent variables $\bs^{1:T}$ and incorporate them into both the forward and reverse processes. Let $\bd^t = \mathrm{deg}(\bA^t)$ be the node degree vector of $\bA^t$, then $\bs^t:= \mathbbm{1}[\bd^{t-1} \neq \bd^t]$ is a binary vector indicating whether nodes are active (having degree change from $t-1$ to $t$) or not from $t - 1$ to $t$. In the following, we redefine the forward and reverse processes. 

\vspace{-0.5em}
\paragraph{Forward process.}
With latent variables $\bs^{1:T}$, we show that the forward process can be rewritten into the following decomposition:
\begin{align}
\!\!\!\!q(\bA^{1:T}, \bs^{1:T}|\bA^0)\!=\!\prod_{t=1}^{T}q(\bA^t|\bA^{t-1})q(\bs^t|\bA^{t-1}, \bA^{t}).
\end{align}
The forward process does not change by including $\bs^{1:T}$ since the value of $\bs^t$ is determined by $\bA^{t-1}$ and $\bA^{t}$. This allows us to use still the forward transition $q(\bA^t|\bA^{t-1})$ to draw the entire sequence.

\vspace{-0.5em}
\paragraph{Reverse process.} We decompose the denoising model as follows:
\begin{align}
\!\!\!\!p_\theta(\bA^{0:T}\!,\bs^{1:T})\!=\!p(\bA^T)\prod_{t=1}^{T}\! p_\theta(\bA^{t-1}|\bA^t\!,\bs^t)p_\theta(\bs^t|\bA^t).\!\!
\label{eq:gen-model}
\end{align}
Here both $p_\theta(\bA^{t-1}|\bA^t,\bs^t)$ and $p_\theta(\bs^t|\bA^t)$  are learnable distributions. Intuitively, the denoising model first predicts which nodes are active $(\bs^t)$ and then generates edges between them to obtain $\bA^{t-1}$. Since we only predict edges between active nodes indicated by $\bs^t$, all edges that incident inactive nodes are carried from $\bA^{t}$ to $\bA^{t-1}$ directly. 

Our EDGE model is specified by \eqref{eq:gen-model}. The generative framework is demonstrated in Figure~\ref{fig:framework}.

\subsection{Learning the reverse process}
\label{sec:obj}

We optimize the model parameters $\theta$ by maximizing the variational lower bound (VLB) of $\log{p(\bA^0)}$. Following~\citet{sohl2015deep, ho2020denoising}, the VLB is: 
\begin{align}
&\calL(\bA^0;\theta) = \bbE_{q}\Big[\log \frac{p_\theta(\bA^{0:T}, \bs^{1:T})}{q(\bA^{1:T}, \bs^{1:T} | \bA^0)}\Big]  \label{eq:L1} \\
&=\mathbb{E}_q\Bigg[\log{\frac{p(\bA^T)}
{q(\bA^T|\bA^0)}}+\underbrace{\log{p_\theta(\bA^0|\bA^1\!,\bs^1)}}_{\text{reconstruction term}~\calL_\mathrm{rec}}+ \nonumber\\
&\sum_{t=2}^T\underbrace{\log{\frac{p_\theta(\bA^{t-1}|\bA^t,\bs^t)}{q(\bA^{t-1}|\bA^t,\bs^t,\bA^0)}}}_{\text{edge prediction term}~\calL_\text{edge}(t)}+\sum_{t=1}^T\!\!\underbrace{\log{\frac{p_\theta(\bs^t|\bA^t)}{q(\bs^t|\bA^t,\bA^0)}}}_{\text{node selection term}~\calL_\text{node}(t)}\!\!\Bigg].\nonumber
\end{align}
Appendix~\ref{deria} shows detailed derivation. The first term contains no learnable parameters. The second term measures the reconstruction likelihood. For the edge prediction term $\calL_\text{edge}(t)$, unlike~\citet{sohl2015deep, ho2020denoising}, the posterior $q(\bA^{t-1}|\bA^{t},\bs^t,\bA^0)$ is hard to compute, and there is not a closed-form for this term. Since the entropy $\mathbb{H}[q(\bA^{t-1}|\bA^t,\bs^t,\bA^0)]$ is a constant, we only optimize the cross entropy term in $\calL_\text{edge}(t)$ via Monte Carlo estimates. We leave the work of variance reduction to the future. 

For the node selection term $\calL_\text{node}(t)$, we show that $q(\bs^t|\bA^t,\bA^0)$ has closed-form expression. In particular, we first derive the posterior of the node degree distribution $q(\bd^t|\bA^t,\bA^0)$ as follows:
\begin{align}
    &q(\bd^{t-1}|\bA^t,\bA^{0})=q(\bd^{t-1}|\bd^t,\bd^{0})=\prod_{i=1}^Nq(\bd^{t-1}_i|\bd^t_i,\bd^{0}_i),\nonumber\\
    &\text{where}~q(\bd^{t-1}_i|\bd^t_i,\bd^0_i)=\mathrm{Bin}(k=\Delta^t_i,n=\Delta^0_i,p=\gamma_t),\nonumber\\
    &\text{with}~\Delta^t_i=\bd^{t-1}_i\!-\!\bd^t_i,~\Delta^0_i=\bd^0_i\!-\!\bd^t_i,~\gamma_t=\frac{\beta_{t}\bar{\alpha}_{t-1}}{1-\bar{\alpha}_{t}}.
\end{align}
Here $\mathrm{Bin}(k; n, p)$ is a binomial distribution parameterized by $n$ and $p$. Intuitively, a node degree $\bd^{t-1}_i$ is only relevant to the node's degrees $\bd^0_i$ and $\bd^t_i$ at steps 0 and $t$. The actual edges do not affect the degree probability since each edge is added or removed independently. We provide formal proof and discuss the forward node degree distribution in Appendix~\ref{degree_change_process}. 

Since $\bs_i^t = \mathbbm{1}[\bd^{t-1}_i\neq\bd^{t}_i]$, we can compute the probability $q(\bs_i^t=1|\bd^t_i,\bd^0_i)$, which is $1-q(\bd^{t-1}_i=\bd^t_i|\bd^t_i,\bd^0_i)$. Finally, we obtain the closed-form posterior:
\begin{align}
    q(\bs^t|\bd^t, \bd^0)=\prod^N_{i=1}q(\bs^t_i|\bd^t_i, \bd^0_i),~\text{where}\label{eq:q-degree-dist}\\
    q(\bs^t_i|\bd^t_i, \bd^0_i)=\calB\big(\bs^t_i;1-(1-\gamma_t)^{\Delta^0_i}\big).\nonumber
\end{align}
The KL divergence $\calL_\text{node}(t)$ turns out to be comparisons between Bernoulli distributions. 

\subsection{Degree-guided graph generation}

\label{sec:deg-seq}
A graph's node degrees are often strongly correlated to its other statistics, so it is important for a generative model to capture the node degrees of training graphs. Here we directly incorporate degree information in the proposed generative model.

We explicitly model node degrees $\bd^0$ of a graph $\bA^0$ as a random variable, then the forward process becomes 
\begin{align}
 q(\bA^{1:T}|\bA^0) =q(\bA^{1:T}|\bA^0)q(\bd^0|\bA^0).
\end{align}
Here $q(\bd^0|\bA^0) = 1$ because $\bd^0$ is determined by $\bA^0$. We also include $\bd^0$ into the generative model $p(\bA^0, \bd^0)$. If the model guarantees that $\bd^0$ is the node degrees of $\bA^0$, 
then  $p_\theta(\bA^0) = p_\theta(\bA^0, \bd^0)$ still models graph data $\bA^0$. Even if $p_\theta(\bA^0, \bd^0)$ allows  
$\bd^0$ to differ a little from the true node degrees, it is still valid to maximize the likelihood $p_\theta(\bA^0, \bd^0 = \bA^0 \bone)$ because model training will encourage the model to generate $\bA^0$ and $\bd^0$ to match each other. We decompose the model by:
\begin{align}
p_\theta(\bA^0, \bd^0) = p_\theta(\bd^0) p_\theta(\bA^{0} | \bd^0). 
\end{align}
 With this decomposition, we first sample arbitrary node degrees $\bd^0$ from $p_\theta(\bd^0)$, then generate a graph with the degree constraint (See Alg.~\ref{alg:deg-generation}). Correspondingly, the denoising model becomes
\begin{align}
p_\theta(\bA^{0:T}, \bs^{1:T}, \bd^0) = p_\theta(\bd^0) p_\theta(\bA^{0:T}, \bs^{1:T} | \bd^0).
\end{align}
We separate the optimizations for the node degree model $p_\theta(\bd^0)$ and the graph denoising model $p_\theta(\bA^{0:T}, \bs^{1:T} | \bd^0)$.  The entire training objective is 
\begin{align} 
\calL(\bA^0\!,\!\bd^0;\!\theta)\!=\!\mathbb{E}_q\bigg[\underbrace{\log{p_\theta(\bd^0)}}_{\calL(\bd^0;\theta)}+\underbrace{\log{\frac{p_\theta(\bA^{0:T}\!,\bs^{1:T}|\bd^0)}{q(\bA^{1:T}\!,\bs^{1:T}|\bA^0)}}}_{\calL(\bA^0|\bd^0;\theta)}\bigg].\nonumber
\end{align}
(See Appendix~\ref{obj2} for detailed derivation.) For $\calL(\bd^0;\theta)$, we treat the learning of node degree distribution as a sequence modeling task. The decomposition of $\calL(\bA^0|\bd^0;\theta)$ remains the same as Eqn.~\eqref{eq:L1}, except that all terms related to the graph denoising model are now conditioning on $\bd^0$. In particular, for the node selection distribution, we consider a special parameterization by setting $p_\theta(\bs^t|\bA^t,\bd^0):=q(\bs^t|\bd^t,\bd^0)$ in Eqn.~\eqref{eq:q-degree-dist}. Note that now the node selection distribution contains no learnable parameters. Moreover, since the KL divergence $\calL_\text{node}(t)$ is now zero, we can further simplify the $\calL(\bA^0|\bd^0;\theta)$ into
\begin{multline} 
   \!\!\! \calL(\bA^0 |\bd^0;\!\theta)\!=\!\mathbb{E}_q\bigg[\!\log{\!\frac{p(\bA^T)}{q(\bA^T|\bA^0)}}\!+\!\log{p_\theta(\bA^0|\bA^1\!,\bs^1,\bd^0)}\!\! \\ 
    +\sum_{t=2}^T\log{\frac{p_\theta(\bA^{t-1}|\bA^t,\bs^t,\bd^0)}{q(\bA^{t-1}|\bA^t,\bs^t,\bA^0)}}\bigg].\label{eq:L2} 
\end{multline} 
In our framework,  the node degree constraint $\bd^0$ is mainly imposed on $p_\theta(\bs^t|\bA^t,\bd^0)$ -- only nodes with a degree below the specified degree $\bd^0$ may be selected to participate in the edge prediction. On the other hand, though the exact probability $p_\theta(\bA^{t-1} | \bA^t, \bs^{t}, \bd^0)$ includes information about the maximum number of edges ($\bd^0-\bd^t$) that can be added to nodes, this can be not easy to track during the edge formation. Here we consider simply augmenting the inputs to the neural network with $\bd^0$. In practice, we found that the specified node degrees $\bd^0$ can accurately control the actual node degrees of the generated graphs.

\begin{algorithm}[t]
  \caption{Degree-guided graph generation}
  \label{alg:deg-generation}
  \begin{algorithmic}
    \STATE {\bfseries Input:} Empty graph $\bA^T$, graph model $p_\theta(\bA^{t-1}|\bA^t, \bs^t)$, degree sequence model $p_\theta(\bd^0)$, and diffusion steps $T$.
    \STATE {\bfseries Output:} Generated graph $\bA^0$
    \STATE Draw $\bd^0\sim p_\theta(\bd^0)$
    \FOR{$t=T,\ldots, 1$}
    \STATE Draw $\bs^t \sim q(\bs^t|\mathrm{deg}(\bA^t), \bd^0)$.
    \STATE Draw $\bA^{t-1} \sim p_\theta(\bA^{t-1}|\bA^t, \bs^t)$.
    \ENDFOR
  \end{algorithmic}
\end{algorithm}

Degree-guided generation turns out to be very useful in modeling large graphs. We reason that the $\bd^0$ significantly reduces the possible trajectories a graph can evolve along, thus reducing the modeling complexity.

\subsection{Implementation}

We briefly describe the implementation of $p_\theta(\bs^t | \bA^{t})$, $p_\theta(\bA^{t-1} | \bA^t, \bs^t)$, and $p_\theta(\bd^0)$. Note we use the same network architecture for $p_\theta(\bA^{t-1} | \bA^t, \bs^t)$ and $p_\theta(\bA^{t-1} | \bA^t, \bs^t, \bd^0)$, except the inputs to the latter includes $\bd^0$.  We treat $p_\theta(\bs^t | \bA^{t})$ as a node classification problem and $p_\theta(\bA^{t-1}|\bA^t,\bs^t)$ as an link prediction problem. Both components share the same MPNN that takes $\bA^t$ as the input and computes node representations $\bZ^t \in \bbR^{N \times d_\mathrm{h}}$ for all nodes. The hidden dimension $d_\mathrm{h}$ is a hyper-parameter here. Then a network head uses $\bZ^t$ to predict $\bs^t$, and another one uses $\bZ^t[\bs^t]$ to predict links between active nodes indicated by $\bs^t$. For the node degree model $p_\theta(\bd^0)$, if there are multiple graphs in the dataset, we use a recurrent neural network~(RNN) to fit the histogram of node degrees. If there is only one graph with node degrees $\bd^0$, then we set $p_\theta(\bd^0) = 1$ directly. Implementation details are in Appendix~\ref{app:arch}.

\subsection{Model analysis}
\label{sec:complexity}
\paragraph{Complexity analysis.}
Let integer $M$ represent the number of edges in a graph, and $K$ be the maximum number of active nodes during the reverse process. In each generation step $t$, the MPNN needs $O(M)$ operations to compute node representations, $O(N)$ operations to predict $\bs^t$, and $O(K^2)$ operations to predict links between $K$ active nodes. The factor $K$ is relevant to noise scheduling: we find that $K$ is smaller than $N$ by at least one order of magnitude when the noise scheduling is linear. In a total of $T$ generation steps, the overall running time $O\big(T\max(K^2, M)\big)$. As a comparison, previous diffusion-based models need running time $O(TN^2)$ because they need to make $O(N^2)$ link predictions at each time step. 
\paragraph{Expressivity analysis.}
EDGE modifies a graph for multiple iterations to generate a sample. In each iteration, it adds new edges to the graph based on the graph structure in the prior iteration. Therefore, EDGE is NOT an edge-independent model and does not have the limitation analyzed by \citet{chanpuriya2021power}, thus it has a theoretical advantage over previous one-shot generative models. 

The ability of EDGE might be affected by the underlying MPNN, which may not be able to distinguish different graph structures due to expressivity issues \citep{xu2018powerful}. However, this issue can be overcome by choosing more expressive GNNs \citep{sato2020survey}. We defer such discussion to future work.

\section{Related Work}

Edge-independent models, which assume edges are formed independently with some probabilities, are prevalent in probabilistic models for large networks. These models include classical models such as ER graph models~\citep{erdos1960evolution}, SBMs~\citep{holland1983stochastic}, and recent neural models such as variational graph auto-encoders~\citep{kipf2016variational,mehta2019stochastic, li2020dirichlet, chen2022interpretable}, NetGAN and its variant~\citep{bojchevski2018netgan, rendsburg2020netgan}. Recent works show that these models can not reproduce desiring statistics of the target network, such as triangle counts, clustering coefficient, and square counts~\citep{seshadhri2020impossibility, chanpuriya2021power}.

Deep auto-regressive (AR) graph models~\citep{li2018learning, you2018graphrnn, liao2019efficient, zang2020moflow, han2023fitting} generate graph edges by sequentially filling up an adjacency matrix to generate a graph. These algorithms are slow because they need to make $N^2$ predictions. \citet{dai2020scalable} proposes a method to leverage graph sparsity and predict only non-zero entries in the adjacency matrix. Long-term memory is a typical issue of these sequential models, so it is hard for them to model global graph properties. Moreover, these models are not invariant with respect to node orders of training graphs, and special techniques \citep{chen2021order, han2023fitting} are often needed to train these models.

Diffusion-based generative models are shown to be powerful in generating high-quality graphs~\citep{niu2020permutation, liu2019graph, jo2022score, haefeli2022diffusion, chen2022nvdiff, vignac2022digress,kongautoregressive}. By ``tailoring'' a graph with multiple steps, these models can model edge correlations. They overcome the limitations of auto-regressive modes as well. However, all previous diffusion-based models focus on generation tasks with small graphs. This work aims to scale diffusion-based models to large graphs.


\section{Experiments}

We empirically evaluate our proposed approach from two perspectives: whether it can capture statistics of training graphs and whether it can generate graphs efficiently. 

\subsection{Experimental setup}
\paragraph{Datasets.} We conduct experiments on both generic graph datasets and large networks. The generic graph datasets consist of multiple graphs of varying sizes. Here we consider Community and Ego datasets~\citep{you2018graphrnn}, all of which contain graphs with hundreds of nodes. We also consider four real-world networks, Polblogs~\citep{adamic2005political}, Cora~\citep{sen2008collective}, Road-Minnesota~\citep{rossi2015network}, and PPI~\citep{stark2010biogrid}. Each of these networks contains thousands of nodes. We also use the QM9 dataset~\citep{ramakrishnan2014quantum} to demonstrate that EDGE can be easily extended to generate graphs with attributes. The statistics of the datasets are shown in Table~\ref{tab:data-stats}. 

\begin{table}[t]
    \small
    \centering
    \begin{tabular}{lcccc}\toprule
         &  \#nodes & \#edges & \#graphs & feature\\\midrule
        Community\!& [60, 160] & [231, 1,965] & 510\\
        Ego & [50, 399] & [57, 1,071] & 757\\
        QM9& [1,9] & [0, 28] & 133,885 & \checkmark\\
        Polblogs & 1,222 & 16,714 & 1\\
        Cora & 2,485 & 5,069 & 1\\
        Road-MN & 2,640 &6,604 & 1\\
        PPI& 3,852 & 37,841 & 1\\\bottomrule
    \end{tabular}
    \vspace{-0.5em}
    \caption{Dataset statistics}
    \vspace{-1.5em}
    \label{tab:data-stats}
\end{table}

\begin{table*}[t]
    \centering
    \small
    \begin{tabular}{lccccccccccc}\toprule
 & \multicolumn{5}{c}{Community} & & \multicolumn{5}{c}{Ego} 
 \\
 & \multicolumn{3}{c}{Structure-based (MMD)} & \multicolumn{2}{c}{Neural-based} && \multicolumn{3}{c}{Structure-based (MMD)} & \multicolumn{2}{c}{Neural-based}
 \\
 & Deg. & Clus. & Orb. & FID & RBF MMD && Deg. & Clus. & Orb. & FID & RBF MMD
\\\midrule
GRNN & 0.1440 & \underline{0.0535} & \textbf{0.0198} & 8.3869 & 0.1591 && \underline{0.0768} & 1.1456 & 0.1087 & 90.5655 & 0.6827
\\
GRAN & 0.1022 & 0.0894 & \textbf{0.0198} & 64.1145 & 0.0749 && 0.5778 & \underline{0.3360} & \textbf{0.0406} & 489.9598 & {0.2633}
\\\midrule					
GraphCNF & 0.1129 & 1.2882 & \textbf{0.0197} & 29.1526 & 0.1341 && 0.1010 & 0.7654 & {0.0820} & \underline{18.7929} & \underline{0.0896}
\\
GDSS & {0.0535} & 0.2072 & \textbf{0.0196} & 6.5531 & \underline{0.0443} && 0.8189 & 0.6032 & 0.3315 & 60.6100 & 0.4331
\\
DiscDDPM & 0.1238 & 0.6549 & \underline{0.0246} & 8.6321 & 0.0840 && 0.4613 & \underline{0.1681} & \underline{0.0633} & 42.7994 & 0.1561
\\
DiGress & \underline{0.0409} & \textbf{0.0167} & 0.0298 & \underline{3.4261} & \underline{0.0460} && \underline{0.0708} & \textbf{0.0092} & 0.1205 & \underline{18.6794} & \textbf{0.0489}
\\\midrule
EDGE & \textbf{0.0175} & \underline{0.0689} & \textbf{0.0198} & \textbf{2.2378} & \textbf{0.0227} && \textbf{0.0579} & \underline{0.1773} & \textbf{0.0519} & \textbf{15.7614} & \textbf{0.0658} 
\\\bottomrule
    \end{tabular}
    \caption{Generation performance on generic graphs. We used unpaired t-tests to compare the results; the numbers in bold indicate the method is better at the 5\% significance level, and the second-best method is underlined. We provide standard deviation in Appendix~\ref{detail_experiment}.}
    \label{tab:main-generic-graph}
\end{table*}

\begin{table*}[h]
    \centering
    \small
    \begin{tabular}{lccccccccccccc}\toprule
 & \multicolumn{6}{c}{Polblogs} && \multicolumn{6}{c}{Cora}
\\
 & EO & PLE & NTC & CC & CPL & AC && EO & PLE & NTC & CC & CPL & AC

\\\midrule
     
True & 
100 & 1.414 & 1 & 0.226 & 2.738 & -0.221 & &
100 & 1.885 & 1 & 0.090 & 6.311 & -0.071

\\\midrule
OPB & 
24.5 & \underline{1.395} & 0.667 & 0.150 & 2.524 & -0.143 &&
10.9 & \textbf{1.852} & 0.097 & 0.008 & 4.476 & \underline{-0.037}

\\
HDOP & 
16.4 & 1.393 & 0.687 & 0.153 & 2.522  & -0.131 &&
0.9 & \textbf{1.849} & 0.113 & 0.009 & 4.770 & \underline{-0.030}

\\ 
CELL & 
26.8 & 1.385 & 0.810 & \underline{0.211} & 2.534 & \underline{-0.230} &&
10.3 & 1.774 & 0.009 & 0.002 & \underline{5.799} & -0.018

\\
CO & 
20.1 & 1.975 & 0.045 & 0.028 & 2.502 & 0.068 &&
9.7 & 1.776 & 0.009 & 0.002 & 5.653 & 0.010

\\ 
TSVD & 
32.0 & 1.373 & \underline{0.872} & 0.205 & 2.532  & \textbf{-0.216} &&
6.7 & \textbf{1.858} & \underline{0.349} & \underline{0.028}  & 4.908 & -0.006
\\
VGAE & 
3.6 & 1.723 & 0.05 & 0.001 & 2.531 & -0.086 &&
1.5 & 1.717 & 0.120 & 0.220 & 4.934 & 0.002
\\\midrule
GRNN & 
9.6 & 1.333 & 0.354 & 0.095 & \underline{2.566} & 0.096 & &
0.4 & \underline{1.822} & 0.043 & 0.011 & \textbf{6.146} & 0.043

\\\midrule
EDGE & 
16.5 & \textbf{1.398} & \textbf{0.977} & \textbf{0.217} & \textbf{2.647} & \textbf{-0.214} & &
1.1 & 1.755 & \textbf{0.446} & \textbf{0.034} & 4.995 & \textbf{-0.046}
\\\bottomrule\toprule

 & \multicolumn{6}{c}{Road-Minnesota} && \multicolumn{6}{c}{PPI}
\\
 & EO & PLE & NTC & CC & CPL & AC && EO & PLE & NTC & CC & CPL & AC

\\\midrule
     
True & 
100 & 2.147 & 1 & 0.028 & 35.349 & -0.187&&
100 & 1.462 & 1 & 0.092 & 3.095 & -0.099

\\\midrule
OPB & 
29.7 & \textbf{2.188} & 0.083 & 0.002 & 8.036 & 0.009 &&
16.3 & \underline{1.443} & 0.640 & 0.058 & 2.914 & \textbf{-0.089}

\\
HDOP & 
13.2 & \textbf{2.192} & \underline{0.208} & 0.004 & 8.274 & {-0.024} &&
6.9 & \underline{1.444} & 0.638 & 0.058 & 2.917 & \underline{-0.086}

\\ 
CELL & 
30.7 & 2.267 & 0.053 & 0.001 & 10.219 & \textbf{-0.082} &&
6.7 & 1.400 & 0.248 & 0.040 & \textbf{3.108} & 0.176

\\
CO & 
19.8 & \underline{2.044} & 2.845 & \textbf{0.040} & \underline{11.478} & -0.012 & &
9.9 & 1.754 & 0.015 & 0.006 & \underline{3.046} & 0.043

\\ 
TSVD & 
19.4 & \textbf{2.172} & 0.060 & 0.001 & 8.431 & 0.006 &&
13.2 & 1.426 & \underline{0.848} & \underline{0.077} & 2.867 & \textbf{-0.089}

\\
VGAE & 
1.3 & {1.678} & 0.096 & 0.009 & 11.120 & -0.027 &&
0.5 & 1.362 & 0.091 & 0.012 & 2.991 & 0.054
\\\midrule

GRNN &
0.6 & 1.570 & 0.099 & 0.007 & \textbf{11.695} & 0.006 & &
OOM & OOM & OOM & OOM & OOM & OOM

\\\midrule
EDGE & 

0.8 & 1.910 & \textbf{0.962} & \underline{0.011} & {9.125} & \underline{-0.063} &&
7.5 & \textbf{1.449} & \textbf{0.981} & \textbf{0.091} & \underline{3.028} & \textbf{-0.107}
\\\bottomrule

    \end{tabular}
    \caption{Graph statistics of generated large networks. EDGE generates graphs with statistics that are much closer to the ground truths. }
    \label{tab:main-large-network}
\end{table*}

\paragraph{Baselines.} 
For generic graphs, We compare EDGE to six recent deep generative graph models, which include two auto-regressive graph models,  GraphRNN \citep{you2018graphrnn} and GRAN \citep{liao2019efficient},  three diffusion-based models, GDSS~\citep{jo2022score}, DiscDDPM~\citep{haefeli2022diffusion} and DiGress~\citep{vignac2022digress}, and one flow-based model, GraphCNF \citep{lippe2020categorical}. For large networks, we follow \citet{chanpuriya2021power} and use six edge-independent models, which include VGAE~\citep{kipf2016variational}, CELL~\citep{rendsburg2020netgan}, TSVD~\citep{seshadhri2020impossibility}, and three methods proposed by~\citet{chanpuriya2021power} (CCOP, HDOP, Linear). We also include GraphRNN as a baseline because it is still affordable to train it on large networks. For the QM9 dataset, We compare EDGE against GDSS~\citep{jo2022score} and DiGress~\citep{vignac2022digress}. The implementation of our model is available at \href{https://github.com/tufts-ml/graph-generation-EDGE}{github.com/tufts-ml/graph-generation-EDGE}. 

\paragraph{Evaluation.} We examine the generated generic graphs with both structure-based and neural-based metrics. For structured-based metrics, we evaluate the Maximum Mean Discrepancy (MMD)~\citep{gretton2012kernel} between test graphs and generated graphs in terms of degrees, clustering coefficients, and orbit counts~\citep{you2018graphrnn}. For neural-based metrics, we evaluate the FID and the MMD RBF metrics proposed by~\citet{thompson2022evaluation}. All implementations of the evaluation are provided by~\citet{thompson2022evaluation}. For all these metrics, the smaller, the better. 

For each large network, we follow \citet{chanpuriya2021power} and evaluate how well the graph statistics of the generated network can match ground truths, which are statistics computed from training data. We consider the following statistics: power-law exponent of the degree sequence (PLE); normalized triangle counts (NTC); global clustering coefficient (CC)~\citep{chanpuriya2021power}; characteristic path length (CPL); and assortativity coefficient (AC)~\citep{newman2002assortative}. We also report the edge overlap ratio (EO) between the generated network and the original one to check to which degree a model memorizes the graph. A graph generated by a good model should have statistics similar to true values computed from the training graph. At the same time, it should have a small EO with the training network, which means that the model should not simply memorize the input data. 

For the QM9 dataset, we evaluate the Validity, Uniqueness, Fr\'{e}chet ChemNet Distance~\citep{preuer2018frechet} and Scaffold similarity~\citep{bemis1996properties} on the samples generated from baselines and our proposed method. We use molsets library~\citep{polykovskiy2020molecular} to implement the evaluation.

\subsection{Evaluation of sample quality}
\paragraph{Generic graph generation.} Table~\ref{tab:main-generic-graph} summarizes the evaluation of generated graphs on the Community and Ego datasets. Best performances are in bold, and second-best performances are underscored. EDGE outperforms all baselines on 8 out of 10 metrics. For the other two metrics, EDGE only performs slightly worse than the best. We hypothesize that EDGE gains advantages by modeling node degrees because they are informative to the graph structure. 

\paragraph{Large network generation.}  Unlike edge-independent models, the edge overlap ratios in the GraphRNN and our approach are not tunable. To make a fair comparison, we report the performance of the edge-independent models that have a similar or higher EO than GraphRNN and EDGE. Table~\ref{tab:main-large-network} shows the statistics of the network itself  (labeled as ``True'') and statistics computed from generated graphs. The statistics nearest to true values are considered as best performances, which are in bold. Second-best performances are underscored. 

\begin{figure*}[t]
 \begin{tabular}{cc}
 \begin{tabular}{c}
 {\includegraphics[width=0.17\textwidth]{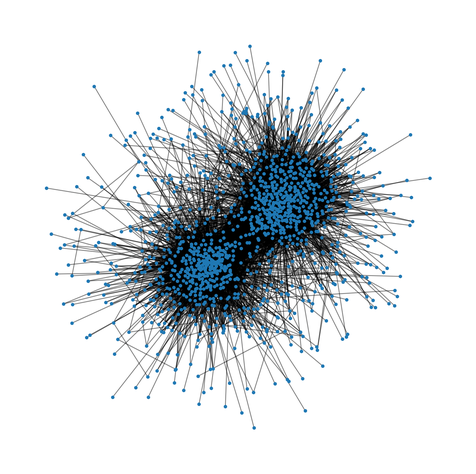}}\\
 Training graph
 \end{tabular}\!&\!
 \begin{tabular}{cccc}
\includegraphics[width=0.17\textwidth]{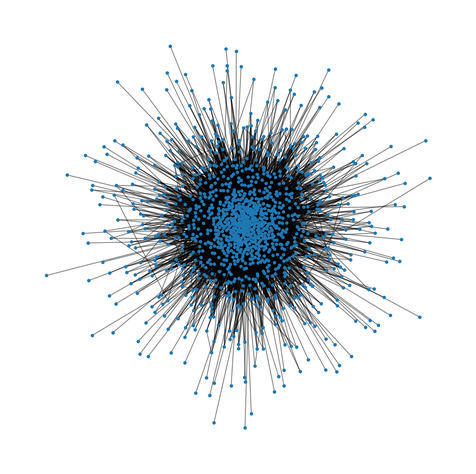}\!\!&\!\!
\includegraphics[width=0.17\textwidth]{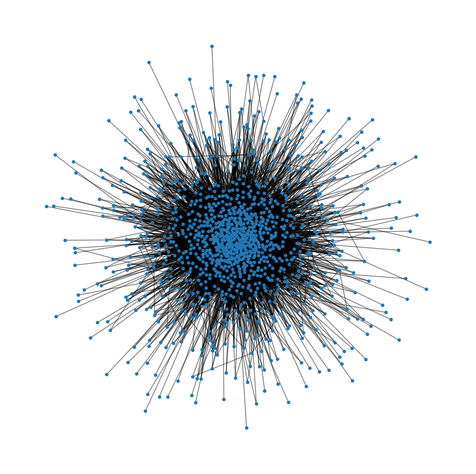}\!\!&\!\!
\includegraphics[width=0.17\textwidth]{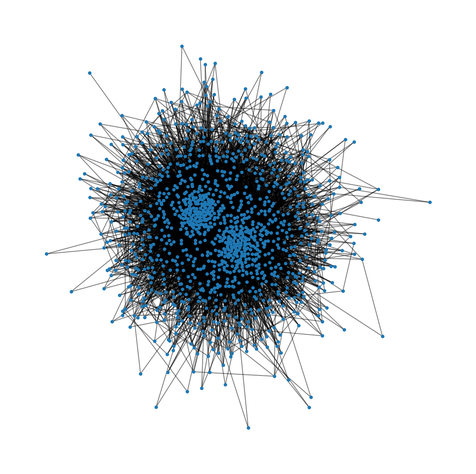}\!\!&\!\!
\includegraphics[width=0.17\textwidth]{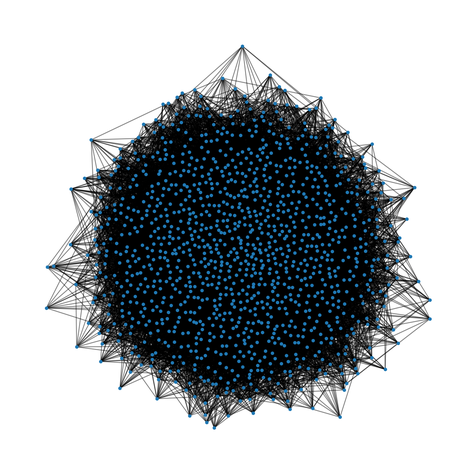}\\
OPB&HDOP&CELL&CO\\
\includegraphics[width=0.17\textwidth]{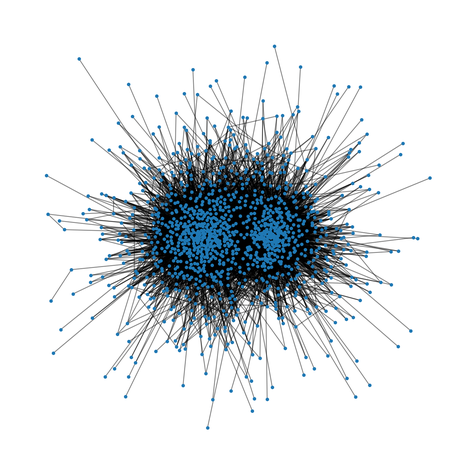}\!\!&\!\!
\includegraphics[width=0.17\textwidth]{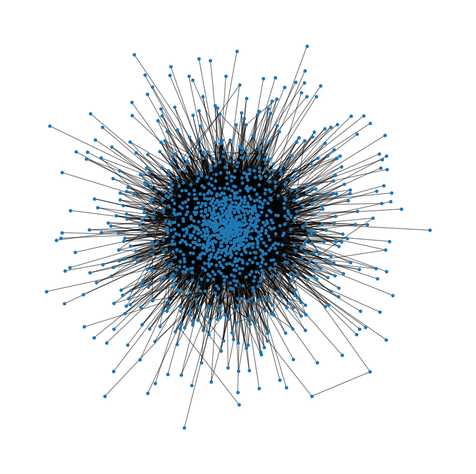}
\!\!&\!\!
\includegraphics[width=0.17\textwidth]{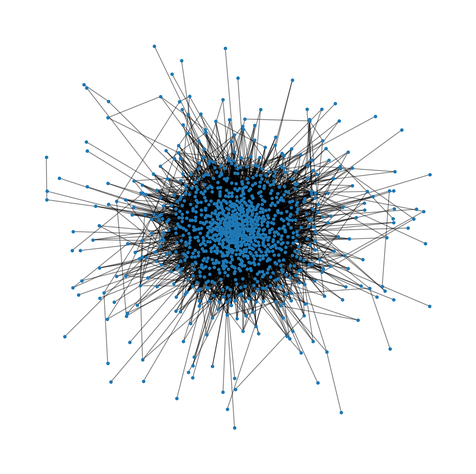}\!\!&\!\!
\includegraphics[width=0.17\textwidth]{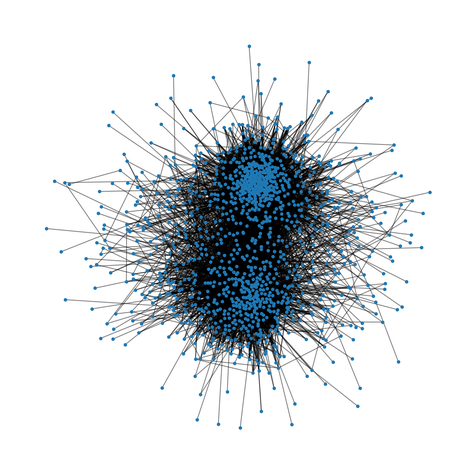} \\
TSVD&VGAE&GRNN&EDGE (ours)
 \end{tabular}
 \end{tabular}
    \caption{Visualization of samples for the Polblogs dataset. We observe that only CELL, TSVD, and EDGE can learn the basic structure of the ground-truth network, while other baselines fail. The network sampled from EDGE appears to be more similar to the training graph.}
    \label{fig:vis-main-polblogs}
\end{figure*}

The proposed approach shows superior performances on all four networks. The improvements on Polblogs and PPI networks are clear. On the Road-Minnesota dataset, EDGE has a much smaller EO than edge-independent models, but its performances in terms of capturing graph statistics are similar to those models. On the Cora dataset, EDGE also has an EO much smaller than edge-independent models, but it slightly improves over these models. Road-Minnesota and Cora networks are both sparse networks -- the message-passing neural model may not work at its full strength. We notice that GraphRNN can not even compete with edge-independent models. We also visualize the generated graphs of Polblogs in~Figure~\ref{fig:vis-main-polblogs}. 

\begin{figure}[t]
    \centering
\scalebox{0.95}{
    \includegraphics[width=0.4\textwidth]{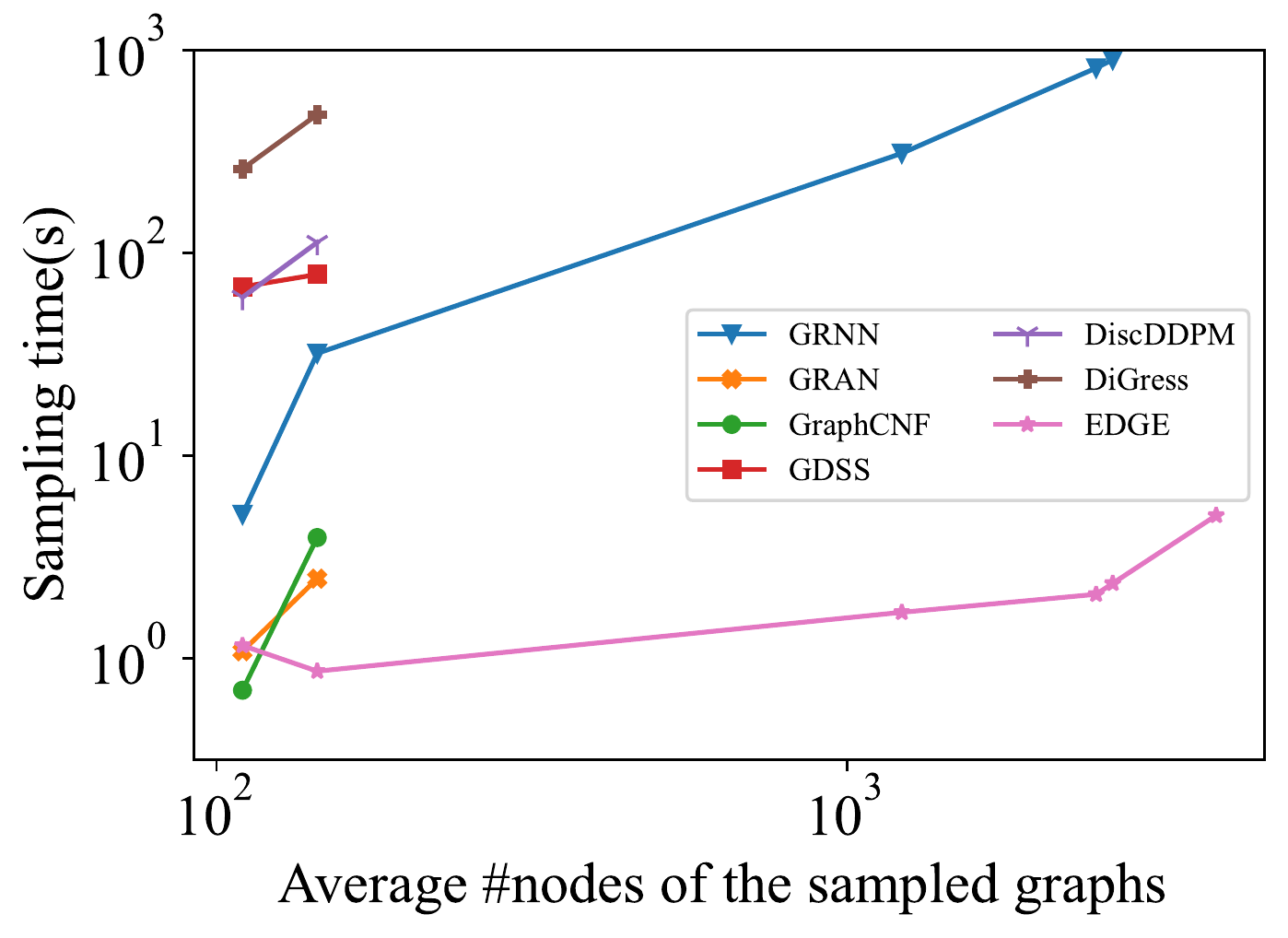}
}
\vspace{-1em}
    \caption{Sampling speed comparison over different models. }
    \vspace{-2em}
    \label{fig:speed}
\end{figure}

\subsection{Efficiency}
\begin{table}[t]
    \centering
    \small
    \begin{tabular}{lcccc}\toprule
&Validity$\uparrow$&Uniqueness$\uparrow$&FCD$\downarrow$&Scaf. Sim.$\uparrow$\\\midrule
GDSS&95.7&98.5&2.9&-\\
DiGress&99.0&\textbf{100}&\textbf{0.151}&\textbf{0.908}\\
EDGE&\textbf{99.1}&\textbf{100}&0.458&0.763\\\bottomrule
    \end{tabular}
    \caption{Generative performance on the QM9 dataset}
    \vspace{-1em}
    \label{tab:qm9-performance}
\end{table}
We compare the sampling efficiency of EDGE against other deep generative graph models. We record the average time for a model to sample one graph to make a consistent comparison over all datasets. The average sampling time for each dataset is averaged over 128 runs. Figure~\ref{fig:speed} shows the relationship between sampling time and graph sizes. Except for GraphRNN, all baseline neural models can only generate graphs for Community and Ego datasets, which contain 110 and 144 nodes on average. Our approach runs only slower than GraphCNF on the Community dataset by 0.5s. On large graphs, our model has a clear advantage in terms of running time. Note that our model spends less time on an Ego graph than a Community graph, though an Ego graph, on average, contains more nodes than a Community graph. This is because the computation of our model scales with the number of edges, and Ego graphs are often sparser than Community graphs. 

\subsection{Generative performance on QM9 dataset}
We further investigate EDGE's ability of generated graphs with node and edge attributes. To include node attributes, we first extend the basic EDGE model with a hierarchical generation process that can also sample node attributes. We put the details of this extension in Appendix~\ref{app:feature-graph-generation}. We evaluate the extended EDGE model on the QM9 dataset and compare it with other neural baselines. The results in Table~\ref{tab:qm9-performance} show that the extended EDGE model has a performance  comparable  with  that of DiGress. Note that DiGress is specially designed for molecule generation, and our model runs much faster than DiGress.  

\begin{table}[t]
    \centering
    \small
    \begin{tabular}{lcccccccc}\toprule
    	\!\!&\!\!\!$\bs^{1:T}$\!\!\!&\!\!\!\! $\bd^0$\!\!\!\!&\!\!PLE\!\!&\!\!NTC\!\!&\!\!CC\!\!&\!\!CPL\!\!&\!\!AC\!\!&\!\!Speed\!\!\\\midrule
\!\!\!True\!\!&\!\!\!\!&\!\!\!\!&\!\!1.414\!\!&\!\!1\!\!&\!\!0.226\!\!&\!\!2.738\!\!&\!\!-0.221\!\!&\\\midrule
\!\!\!G(N,0.5)\!\!&\!\!\!\!&\!\!\!\!&\!\!OOM\!\!&\!\!OOM\!\!&\!\!OOM\!\!&\!\!OOM\!\!&\!\!OOM\!\!&\!\!OOM\\
\!\!\!G(N,0)\!\!&\!\!\!\!&\!\!\!\!&\!\!1.341\!\!&\!\!3.234\!\!&\!\!0.237\!\!&\!\!\textbf{2.747}\!\!&\!\!-0.304\!\!&\!\!15.3s\\
\!\!\!G(N,0)\!\!&\!\!$\checkmark$\!\!&\!\!\!\!&\!\!1.383\!\!&\!\!2.364\!\!&\!\!0.251\!\!&\!\!2.638\!\!&\!\!-0.331\!\!&\!\!2.1s\\
\!\!\!G(N,0)\!\!&\!\!$\checkmark$\!\!&\!\!$\checkmark$\!\!&\!\!\textbf{1.398}\!\!&\!\!\textbf{0.977}\!\!&\!\!\textbf{0.217}\!\!&\!\!2.647\!\!&\!\!\textbf{-0.214}\!\!&\!\!\textbf{1.7s}\\\bottomrule
    \end{tabular}
    \caption{Performance of EDGE's variants on the Polblogs dataset.}
    \vspace{-1em}
    \label{tab:model-variants}
\end{table}

\subsection{Ablation studies}
\paragraph{Diffusion variants.} The random variables $\bs^{1:T}$ and $\bd^0$ play important roles in EDGE's good performances, and we verify that through an ablation study on the Polblogs dataset. We use four diffusion configurations: 1) setting $G(N, 0.5)$ as the convergent distribution and directly using an MPNN as the denoising model $p_\theta(\bA^{t-1} | \bA^{t})$; 2) setting $G(N, 0)$ as the convergent distribution and directly using an MPNN as the denoising model (without modeling active nodes and degree guidance); 3) the EDGE model without degree guidance, and 4) the EDGE model. Table~\ref{tab:model-variants} shows the performances of the four models. If we set the convergent distribution to $G(N,0.5)$, we can not even train such as model since it requires an excessively large amount of GPU memory. This justifies our use of $G(N,0)$ as the convergent distribution. The introduction of $\bs^{1:T}$ (Section~\ref{sec:framework}) significantly improves the sampling speed. Finally, the EDGE approach, which explicitly models node degrees $\bd^0$ and generates graphs with degree guidance, further improves the  generative performance.

\vspace{-0.5em}
\paragraph{Diffusion steps vs. model performance.}
In EDGE, the number of diffusion steps $T$ decides how many nodes would actively participate in the edge prediction. Here we investigate how it affects the model performance under linear noise scheduling. 
\begin{table}[H]
    \centering
    \small
    \begin{tabular}{clcccccc}\toprule
\!&\!\!&\!EO\!&\!PLE\!&\!NTC\!&\!CC\!&\!CPL\!&\!AC\\\midrule
\multirow{6}{1pt}{\rotatebox{90}{Ploblogs}}
&\!True\!&\!100\!&\!1.414\!&\!1\!&\!0.226\!&\!2.738\!&\!-0.221 \\
\!&\!64\!&\!1.8\!&\!1.380\!&\!1.148\!&\!0.235\!&\!2.800\!&\!-0.202\\
\!&\!128$^*$\!&\!14.9\!&\!1.386\!&\!1.030\!&\!0.238\!&\!2.747\!&\!-0.238\\	
\!&\!256$^*$\!&\!16.5\!&\!1.398\!&\!0.977\!&\!0.217\!&\!2.647\!&\!-0.214\\	
\!&\!512$^*$\!&\!15.0\!&\!1.398\!&\!0.923\!&\!0.218\!&\!2.635\!&\!-0.268\\ 
\!&\!1024$^*$\!&\!16.5\!&\!1.400\!&\!0.991\!&\!0.219\!&\!2.665\!&\!-0.246\\\midrule
\multirow{6}{1pt}{\rotatebox{90}{Cora}}
&\!True\!&\!100\!&\!1.885\!&\!1\!&\!0.090\!&\!6.311\!&\!-0.071 \\
\!&\!64$^*$\!&\!0.9\!&\!1.755\!&\!0.446\!&\!0.034\!&\!4.995\!&\!-0.046\\
\!&\!128\!&\!1.1\!&\!1.747\!&\!0.555\!&\!0.042\!&\!5.017\!&\!-0.050\\
\!&\!256\!&\!0.8\!&\!1.753\!&\!0.360\!&\!0.027\!&\!4.818\!&\!-0.041\\
\!&\!512\!&\!0.8\!&\!1.753\!&\!0.360\!&\!0.027\!&\!4.818\!&\!-0.042\\
\!&\!1024\!&\!0.9\!&\!1.762\!&\!0.348\!&\!0.027\!&\!4.778\!&\!-0.034\\\midrule
\multirow{6}{1pt}{\rotatebox{90}{Road-MN}}
&\!True& 100\!&\!2.147\!&\!1\!&\!0.028\!&\!35.349\!&\!-0.187\\
\!&\!64$^*$\!&\!0.8\!&\!1.910\!&\!0.962\!&\!0.011\!&\!9.125\!&\!-0.063\\
\!&\!128\!&\!1.2\!&\!1.803\!&\!1.232\!&\!0.041\!&\!6.501\!&\!-0.030\\	
\!&\!256\!&\!0.8\!&\!1.953\!&\!1.057\!&\!0.014\!&\!7.471\!&\!-0.005\\
\!&\!512\!&\!1.3\!&\!1.965\!&\!1.472\!&\!0.020\!&\!7.710\!&\!-0.006\\
\!&\!1024\!&\!1.2\!&\!1.983\!&\!2.491\!&\!0.035\!&\!7.906\!&\!-0.034\\\midrule

\multirow{6}{1pt}{\rotatebox{90}{PPI}}
&\!True\!&\!100\!&\!1.462\!&\!1\!&\!0.092\!&\!3.095\!&\!-0.099 \\
\!&\!64\!&\!7.4\!&\!1.421\!&\!2.455\!&\!-0.116\!&\!3.498\!&\!-0.116\\
\!&\!128\!&\!6.2\!&\!1.419\!&\!1.503\!&\!0.126\!&\!3.384\!&\!-0.147\\	
\!&\!256$^*$\!&\!7.5\!&\!1.449\!&\!0.981\!&\!0.091\!&\!3.028\!&\!-0.107\\
\!&\!512$^*$\!&\!7.0\!&\!1.438\!&\!1.101\!&\!0.099\!&\!3.244\!&\!-0.107\\
\!&\!1024$^*$\!&\!7.1\!&\!1.441\!&\!0.925\!&\!0.074\!&\!3.150\!&\!-0.101\\\bottomrule
    \end{tabular}
    \caption{Large diffusion steps $T$ does not necessarily improve model performance. Good diffusion steps are labeled with ``*''.}
    \label{tab:step-vs-performance}
\end{table}

Specifically, we train our model on three large networks with $T\in\{64,128,256,512,1024\}$ and report the model performance in Table~\ref{tab:step-vs-performance}. Unlike traditional diffusion models in which more diffusion steps usually yield better performance, a large $T$ for our model does not always improve the performance. For instance, $T=64$ gives the best performance in the Cora  and Road-Minnesota datasets. Our explanation for this observation is the high level of sparsity in training graphs.  If we have a large $T$, the total number of generation steps, the model can only identify a few active nodes and predict edges between them in each time step. The model faces a highly imbalanced classification problem, which may lead to poor model convergence. Such an issue is not observed for relatively denser graphs, e.g. Polblogs and PPI datasets, which require a relatively large $T$ to guarantee good model performances. When $T$ is large enough ($T=128$ for Polbogs and $T=256$ for PPI), further increasing $T$ does not improve the model performance.

\section{Conclusion}
In this work, we propose EDGE, a generative graph model based on a discrete diffusion process. By leveraging the sparsity in the diffusion process, EDGE significantly improves the computation efficiency and scales to graphs with thousands of nodes. By explicitly modeling node degrees, EDGE improves its ability in capturing important statistics of training graphs. Our extensive empirical study shows that EDGE has superior performance in benchmark graph generation in terms of both computational efficiency and generation quality.      
\section*{Acknowledgment}

We thank anonymous reviewers for their valuable feedback. Xiaohui Chen and Li-Ping Liu are partially supported by the National Science Foundation under Grant No.~2239869.


\bibliography{main}
\bibliographystyle{icml2023}

\newpage
\appendix
\onecolumn

\centerline{\Large{\textbf{Appendix}}}
\vspace{1em}

\normalsize
\section{Derivation of the $G(N,0)$ Diffusion Process and the Degree Change Distribution}
We first derive the forward and reverse transition distribution of the $G(N,0)$ diffusion process, then we examine the distribution of node degree changes for both directions.
\subsection{The $G(N,0)$ diffusion process}
\label{derivation of diffusion process}
We consider modeling the upper triangle of the adjacency matrix $\bA^0$. Since we have $p=0$ in our framework, for the forward transition kernel, we have
\begin{align}
    &q(\bA^t|\bA^{t-1})=\prod_{i,j: i<j}q(\bA_{i,j}^t|\bA_{i,j}^{t-1}),~\text{with}~q(\bA_{i,j}^t|\bA_{i,j}^{t-1})=\calB(\bA^t_{i,j};(1-\beta_t)\bA^{t-1}_{i,j}).\\
    &q(\bA^t|\bA^0)=\prod_{i,j: i<j}q(\bA_{i,j}^t|\bA_{i,j}^0),~\text{with}~q(\bA_{i,j}^t|\bA_{i,j}^0)=\calB(\bA^t_{i,j};\bar{\alpha}_t\bA^0_{i,j}).
\end{align}
The posterior $q(\bA^{t-1}|\bA^t,\bA^0)$, whose form is discussed in Eqn.~\eqref{eq:post}, is decomposed into
\begin{align}
    q(\bA^{t-1}|\bA^t,\bA^0)&=\frac{q(\bA^{t}|\bA^{t-1})q(\bA^{t-1}|\bA^0)}{q(\bA^{t}|\bA^0)}\\
    &=\prod_{i,j: i<j}\frac{q(\bA_{i,j}^{t}|\bA_{i,j}^{t-1})q(\bA_{i,j}^{t-1}|\bA_{i,j}^0)}{q(\bA_{i,j}^{t}|\bA_{i,j}^0)}\nonumber\\
    &=\prod_{i,j: i<j} q(\bA_{i,j}^{t-1}|\bA_{i,j}^t,\bA_{i,j}^0)\nonumber
\end{align}

The entry-wise posterior distribution $q(\bA_{i,j}^{t-1}|\bA_{i,j}^t,\bA_{i,j}^0)$ is the key to deriving the distribution of active nodes. Here, we describe the detailed form of this distribution. For any value $p\in[0,1]$, we have the form
\begin{align}
     &q(\bA_{i,j}^{t-1}|\bA_{i,j}^{t},\bA_{i,j}^0)=\calB(\bA_{i,j}^{t-1};\frac{p_1}{p_0+p_1}),\text{where}\\
     &p_1=[(1-\beta_t+\beta_t p){\bA_{i,j}^t}+(\beta_t-\beta_tp)(1-\bA_{i,j}^t)][\bar{\alpha}_{t-1}\bA_{i,j}^0+(1-\bar{\alpha}_{t-1})p]\\
     &p_0=[(\beta_tp){\bA_{i,j}^t} + (1-\beta_tp)(1-\bA_{i,j}^t)][1+\bar{\alpha}_{t-1}p-\bar{\alpha}_{t-1}\bA_{i,j}^0-p]
\end{align}

Note that the posterior derived in~\citet{sohl2015deep, hoogeboom2021argmax} is only applicable to the case where $p=0.5$, the above posterior is more general. In particular, for $p=0$ in our case, the posterior can be simplified into the following three cases
\begin{align}
     q(\bA_{i,j}^{t-1}|\bA_{i,j}^{t},\bA_{i,j}^0)=
     \begin{cases}
     \calB(\bA_{i,j}^{t-1};0), &\bA_{i,j}^0=0\\
     \calB(\bA_{i,j}^{t-1};1), &\bA_{i,j}^0=1, \bA_{i,j}^t=1\\
     \calB(\bA_{i,j}^{t-1};\frac{\beta_t\bar{\alpha}_{t-1}}{1-\bar{\alpha}_t}), &\bA_{i,j}^0=1, \bA_{i,j}^t=0
     \end{cases}\label{eq:post-3cases}
\end{align}

We provide an intuitive interpretation of the three cases. Since we are considering an edge-removing process, for the case where $\bA^0_{i,j} = 0$, the probability an edge $(i,j)$ is formed at timestep $t-1$ is 0 (note that the event $(\bA_{i,j}^0=0,\bA_{i,j}^t=1)$ is unlikely to happen). The case where $\bA^0_{i,j} = \bA^t_{i,j} = 1$ indicates the edge $(i,j)$ is not removed for all timesteps from $0$ to $t$, therefore, $\bA^{t-1}_{i,j}$ always equals to 1. The last case is the only case with uncertainty since an edge $(i,j)$ can be removed at any timestep before $t$. 

\subsection{The distribution of active nodes}
\label{degree_change_process}
We now have the entry-wise forward distribution and posterior distribution. We can  compute the probability that a node has a degree change at each time step. Here we first discuss the form of the forward and posterior degree distributions, which can be directly applied to calculate the degree change distributions:

\textbf{Property 1.} The forward degree distributions have the form
\begin{align}
    &q(\bd^t|\bd^0)=\prod_{i=1}^N q(\bd^t_i|\bd^0_i),~\text{where} ~q(\bd^t_i|\bd^0_i)=\mathrm{Binomial}(k=\bd^t_i, n=\bd^0_i, p=\bar{\alpha}_t).\\
    &q(\bd^t|\bd^{t-1})=\prod_{i=1}^N q(\bd^t_i|\bd^{t-1}_i),~\text{where} ~q(\bd^t_i|\bd^{t-1}_i)=\mathrm{Binomial}(k=\bd^t_i, n=\bd^{t-1}_i, p=1-\beta_t).
\end{align}
Intuitively, for $q(\bd^t|\bd^0)$, there are $\bd_i^0$ edges connected to node $i$, each with probability $\bar{\alpha}_t$ to be kept at time step $t$. The probability the number of remaining edges equals $\bd_i^t$ at time step $t$ is a binomial distribution. A similar statement also holds for the one-step transition $q(\bd^t|\bd^{t-1})$, where an edge will have probability $1-\beta_t$ be kept when transiting from $t-1$ to $t$.


We also need to compute $q(\bd^{t-1}|\bd^t,\bd^0)$, and we show that $q(\bd^{t-1}|\bd^t,\bd^0)=q(\bd^{t-1}|\bA^t,\bA^0)$. It holds because edges are removed independently. Since 
\begin{align}
\bd_{i}^{t-1} = \sum_{j = 1}^{N} \bA^{t-1}_{i,j}, 
\end{align}
and $\bA^{t-1}_{i,j}$-s are independent variables. According to  \eqref{eq:post-3cases}, $\bd_{i}^{t-1}$  is the summation of three types of independent random variables: the first type is always 0, and the second type is always 1. We only need to consider the second and third types of variables, whose counts are respectively $\bd^t_i$ and $(\bd^{0}_i - \bd^{t}_i)$. Then  $\bd^{t-1}$ in $q(\bd^{t-1}|\bA^t,\bA^0)$ is the sum of $\bd^t_i$ and a random variable from $\mathrm{Binomial}\left(n = \bd^{0}_i - \bd^{t}_i, p=\frac{\beta_t \bar{\alpha}_{t-1}}{1 - \bar{\alpha}_t}\right)$. It also indicates that $q(\bd^{t-1}|\bd^t,\bd^0)=q(\bd^{t-1}|\bA^t,\bA^0)$.   

\textbf{Property 2.} The posterior degree distribution $q(\bd^{t-1}|\bd^t,\bd^{0})$ has the form:
\begin{align}
    &q(\bd^{t-1}|\bd^t,\bd^{0})=\prod_{i=1}^N q(\bd^{t-1}_i|\bd^t_i,\bd^{0}_i),~\text{with}\label{eq:p2}\\
    &q(\bd^{t-1}_i|\bd^t_i,\bd^0_i)=\mathrm{Binomial}\left(k=(\bd^{t-1}_i-\bd^t_i); n=\bd^0_i-\bd^t_i,p=\frac{\beta_{t}\bar{\alpha}_{t-1}}{1-\bar{\alpha}_{t}}\right). 
\end{align}

Now we have derived the forward and reverse degree distributions. It's obvious that when a node has no degree change from time step $t-1$ to $t$ or the other way around, we always have $\bd^{t-1}_i=\bd^t_i$. The probabilities of such events can be computed by querying the degree distributions $q(\bd^{t}|\bd^{t-1})$ or $q(\bd^{t-1}|\bd^t,\bd^{0})$. Let $\bs^t_i$ be the random variable that node $i$ has degree change at time step $t$. Below we show the forward and reverse degree change distributions:

\textbf{Property 3.}  At timestep $t$, the forward degree change distribution for node $i$ given $\bd^{t-1}_i$ is
\begin{align}
    q(\bs^t_i|\bd^{t-1}_i)=\calB\big(\bs^t_i;1-(1-\beta_t)^{\bd^{t-1}_i}\big). \label{eq:forward-deg-change}
\end{align}
\textbf{Property 4.} At timestep $t$, the reverse degree change distribution for node $i$ given $\bd^t_i, \bd^0_i$ is
\begin{align}
    q(\bs^t_i|\bd^t_i, \bd^0_i)=\calB\left(\bs^t_i;1-\left(1-\frac{\beta_{t}\bar{\alpha}_{t-1}}{1-\bar{\alpha}_{t}}\right)^{\bd^0_i - \bd^t_i}\right).
\end{align}
The distribution of active nodes from $q(\bs^t_i|\bd^{t-1}_i)$  provides insightful supporting evidence that only a part of nodes may have degree change at each transition, motivating us the develop such a scalable generative framework. Controlling the number of nodes with degree change in the forward process can function as a principle to improve the noise scheduling algorithm. In practice, we only use the reverse degree change distribution when learning the reverse process. The reverse degree distribution is essential in improving the model expressivity since it enables graph generation with degree guidance.

\section{Derivations of the Training Objectives}
\subsection{Derivation of the objective $\calL(\bA^0;\theta)$}
\label{deria}
To obtain the objective $\calL(\bA^0;\theta)$, we first need to derive the posterior of $\bA^{t-1}$ that conditions on the introduced latent variables $\bs^t$
\begin{align}
    q(\bA^{t-1}|\bA^t,\bs^t,\bA^0) &=\frac{q(\bA^{t-1},\bA^t, \bs^t|\bA^0)}{q(\bA^t, \bs^t|\bA^0)}\\
    &= \frac{q(\bA^t|\bA^{t-1},\bA^0)q(\bs^t|\bA^t,\bA^{t-1},\bA^0)q(\bA^{t-1}|\bA^0)}{q(\bA^t,\bs^t|\bA^0)}\nonumber\\
    &=  \frac{q(\bA^t|\bA^{t-1})q(\bs^t|\bA^t,\bA^{t-1})q(\bA^{t-1}|\bA^0)}{q(\bA^t|\bA^0)q(\bs^t|\bA^t, \bA^0)}.\nonumber
\end{align}
By rearranging terms, we have 
\begin{align}
    q(\bA^t|\bA^{t-1})q(\bs^t|\bA^t,\bA^{t-1})=\frac{q(\bA^{t-1}|\bA^t,\bs^t,\bA^0)q(\bA^t|\bA^0)q(\bs^t|\bA^t, \bA^0)}{q(\bA^{t-1}|\bA^0)}.
\end{align}
The VLB $\calL(\bA^0;\theta)$ of $\log{p(\bA^0)}$ is derived as follow
\begin{align}
    &\calL(\bA^0;\theta)\\
    &= \mathbb{E}_{q}\big[\log{\frac{p_\theta(\bA^{0:T}, \bs^{1:T})}{q(\bA^{1:T}, \bs^{1:T}|\bA^0)}}\big]\nonumber\\
    &=\mathbb{E}_q\big[\log{\frac{p(\bA^T)\prod_{t=1}^{T}p_\theta(\bA^{t-1},\bs^t|\bA^t)}{\prod_{t=1}^{T}q(\bA^t,\bs^{t}|\bA^{t-1})}}\big]\nonumber\\
   &=\mathbb{E}_{q}\big[\log{p(\bA^T)} +\sum_{t=1}^T\log{\frac{p_\theta(\bA^{t-1}|\bA^t,\bs^t)p_\theta(\bs^t|\bA^t)}{q(\bA^t|\bA^{t-1})q(\bs^t|\bA^t,\bA^{t-1})}}\big]\nonumber\\
    &=\mathbb{E}_{q}\big[\log{p(\bA^T)} + \log{\frac{p_\theta(\bA^0|\bA^1,\bs^1)p_\theta(\bs^1|\bA^1)}{q(\bA^1|\bA^0)q(\bs^1 |\bA^1,\bA^0)}} + \sum_{t=2}^T\log{\frac{p_\theta(\bA^{t-1}|\bA^t,\bs^t)p_\theta(\bs^t|\bA^t)}{\frac{q(\bA^{t-1}|\bA^t,\bs^t,\bA^0)q(\bA^t|\bA^0)q(\bs^t|\bA^t, \bA^0)}{q(\bA^{t-1}|\bA^0)}}}\big]\nonumber\\
    &=\mathbb{E}_{q}\big[\log{p(\bA^T)} + \log{\frac{p_\theta(\bA^0|\bA^1,\bs^1)p_\theta(\bs^1|\bA^1)}{q(\bA^1|\bA^0)q(\bs^1|\bA^1,\bA^0)}} + \sum_{t=2}^T\log{\frac{p_\theta(\bA^{t-1}|\bA^t,\bs^t)p_\theta(\bs^t|\bA^t)q(\bA^{t-1}|\bA^0)}{q(\bA^{t-1}|\bA^t,\bs^t,\bA^0)q(\bs^t|\bA^t,\bA^0)q(\bA^t|\bA^0)}}\big]\nonumber\\
    &=\mathbb{E}_{q}\big[\log{\frac{p(\bA^T)\cancel{q(\bA^T|\bA^0)}}{q(\bA^T|\bA^0)}}\!+\! \log{\frac{p_\theta(\bA^0|\bA^1,\bs^1)p_\theta(\bs^1|\bA^1)}{\cancel{q(\bA^1|\bA^0)}q(\bs^1|\bA^1,\bA^0)}}\!+\! \sum_{t=2}^T\log{\frac{p_\theta(\bA^{t-1}|\bA^t,\bs^t)p_\theta(\bs^t|\bA^t)\cancel{q(\bA^{t-1}|\bA^0)}}{q(\bA^{t-1}|\bA^t,\bs^t,\bA^0)q(\bs^t|\bA^t,\bA^0)\cancel{q(\bA^t|\bA^0)}}}\big]\nonumber\\
    &=\mathbb{E}_q\big[\log{\frac{p(\bA^T)}{q(\bA^T|\bA^0)}}+\underbrace{\log{p_\theta(\bA^0|\bA^1,\bs^1)}}_{\text{reconstruction term}~\calL_\mathrm{rec}}+\sum_{t=2}^T\underbrace{\log{\frac{p_\theta(\bA^{t-1}|\bA^t,\bs^t)}{q(\bA^{t-1}|\bA^t,\bs^t,\bA^0)}}}_{\text{edge prediction term}~\calL_\text{edge}(t)}+\sum_{t=1}^T\underbrace{\log{\frac{p_\theta(\bs^t|\bA^t)}{q(\bs^t|\bA^t,\bA^0)}}}_{\text{node selection term}~\calL_\text{node}(t)}\big].\nonumber
\end{align}
The objective requires modeling two latent variables: $\bA^{1:T}$ and $\bs^{1:T}$. Learning to predict $\bs^t$ from $\bA^t$ can be difficult since it involves capturing the dynamic interaction between nodes and the global structure of the current graph $\bA^t$. In Section~\ref{obj2}, we demonstrate a new objective which can avoid learning $p_\theta(\bs^t|\bA^t)$ by instead learning the node degree distribution $p_\theta(\bd^0)$.

\subsection{Derivation of the objective $\calL(\bA^0,\bd^0;\theta)$}
\label{obj2}
Since $p_\theta(\bA^0)=p_\theta(\bA^0,\bd^0)$, we have
\begin{align}
\log{p_\theta(\bA^0)}
=\log{p_\theta(\bA^0,\bd^0)}&\geq\calL(\bA^0,\bd^0;\theta)\\
&=\mathbb{E}_q\big[\log{\frac{p_\theta(\bd^0)p_\theta(\bA^0|\bd^0)}{q(\bd^0|\bA^0)q(\bA^{1:T}|\bA^0)}}\big]\nonumber\\
&=\mathbb{E}_q\big[\log{\frac{p_\theta(\bd^0)p_\theta(\bA^{0:T},\bs^{1:T}|\bd^0)}{q(\bd^0|\bA^0)q(\bA^{1:T},\bs^{1:T}|\bA^0)}}\big]\nonumber\\
&=\underbrace{\mathbb{E}_q\big[\log{\frac{p_\theta(\bd^0)}{q(\bd^0|\bA^0)}}\big]}_{\calL(\bd^0;\theta)}+\underbrace{\mathbb{E}_q\big[\log{\frac{p_\theta(\bA^{0:T},\bs^{1:T}|\bd^0)}{q(\bA^{1:T},\bs^{1:T}|\bA^0)}}\big]}_{\calL(\bA^0|\bd^0;\theta)}.\nonumber
\end{align}
Optimizing $\calL(\bd^0;\theta)$ is equivalent to fitting $p_\theta(\bd^0)$ to the node degree data distribution $p_\mathrm{data}(\bd^0)$ as $\bd^0$ is obtained from $\bA^0$. The full decomposition of $\calL(\bA^0|\bd^0;\theta)$ has the following form:
\begin{align}
\calL(\bA^0 |\bd^0;\!\theta)\!=\mathbb{E}_q\big[\log{\frac{p(\bA^T)}{q(\bA^T|\bA^0)}}+\underbrace{\log{p_\theta(\bA^0|\bA^1,\bs^1,\bd^0)}}_{\text{reconstruction term}~\calL_\mathrm{rec}}+\sum_{t=2}^T\underbrace{\log{\frac{p_\theta(\bA^{t-1}|\bA^t,\bs^t,\bd^0)}{q(\bA^{t-1}|\bA^t,\bs^t,\bA^0)}}}_{\text{edge prediction term}~\calL_\text{edge}(t)}+\sum_{t=1}^T\underbrace{\log{\frac{p_\theta(\bs^t|\bA^t,\bd^0)}{q(\bs^t|\bA^t,\bA^0)}}}_{\text{node selection term}~\calL_\text{node}(t)}\!\!\!\big].
\end{align}
Here $\bA^T$ is independent from $\bd^0$ so $p(\bA^T|\bd^0)=p(\bA^T)$. And as mentioned before, we choose to parameterize $p_\theta(\bs^t|\bA^t,\bd^0):=q(\bs^t|\bA^t,\bA^0)$, resulting in the KL divergence $\calL_\text{node}(t)=0$ for all $t$. The objective is further simplified to
\begin{align}
    \calL(\bA^0 |\bd^0;\theta)=\mathbb{E}_q\big[\log{\frac{p(\bA^T)}{q(\bA^T|\bA^0)}}+\log{p_\theta(\bA^0|\bA^1,\bs^1,\bd^0)}+\sum_{t=2}^T\log{\frac{p_\theta(\bA^{t-1}|\bA^t,\bs^t,\bd^0)}{q(\bA^{t-1}|\bA^t,\bs^t,\bA^0)}}\big].
\end{align}

\section{Detailed Implementations of the Denoising Networks and Training}

\label{app:arch}
\subsection{Parameterization of the edge prediction distribution}
\label{parameterazation_of_edge_prediction}
As we consider modeling the upper triangle of the adjacency matrix, the edge prediction distribution $p_\theta(\bA^{t-1}|\bA^t,\bs^t,\bd^0)$ is parameterized as:
\begin{align}
    &p_\theta(\bA^{t-1}|\bA^t, \bs^t,\bd^0)=\prod_{i,j:i<j}p_\theta(\bA^{t-1}_{i,j}|\bA^t_{i,j}, \bs^t_{i,j},\bd^0),~\text{with}\\
    &p_\theta(\bA^{t-1}_{i,j}|\bA^t_{i,j}, \bs^{t}_{i,j},\bd^0)=\calB(\bA_{i,j}^{t};\bs^{t}_{i,j}\ell_{i,j}^{t-1}+(1-\bs^{t}_{i,j})\bA_{i,j}^t),\nonumber\\
    &\text{where } \bs^t_{i,j}=\bs^t_{i}\bs^t_{j},~\ell^{t-1}=\mathrm{gnn}_\theta(\bA^t, \bs^t,\bd^0, t).\nonumber
\end{align}
Note that only when nodes $i,j$ are both selected, i.e., $\bs^t_{i,j}=1$, the corresponding Bernoulli parameter $\ell_{i,j}^{t-1}$ effectively decides the edge distribution. Below we elaborate on the architecture of the used network and analyze the runtime complexity. 

\begin{figure*}[t]
    \centering
    \begin{tabular}{c|c}
    \includegraphics[width=0.38\textwidth]{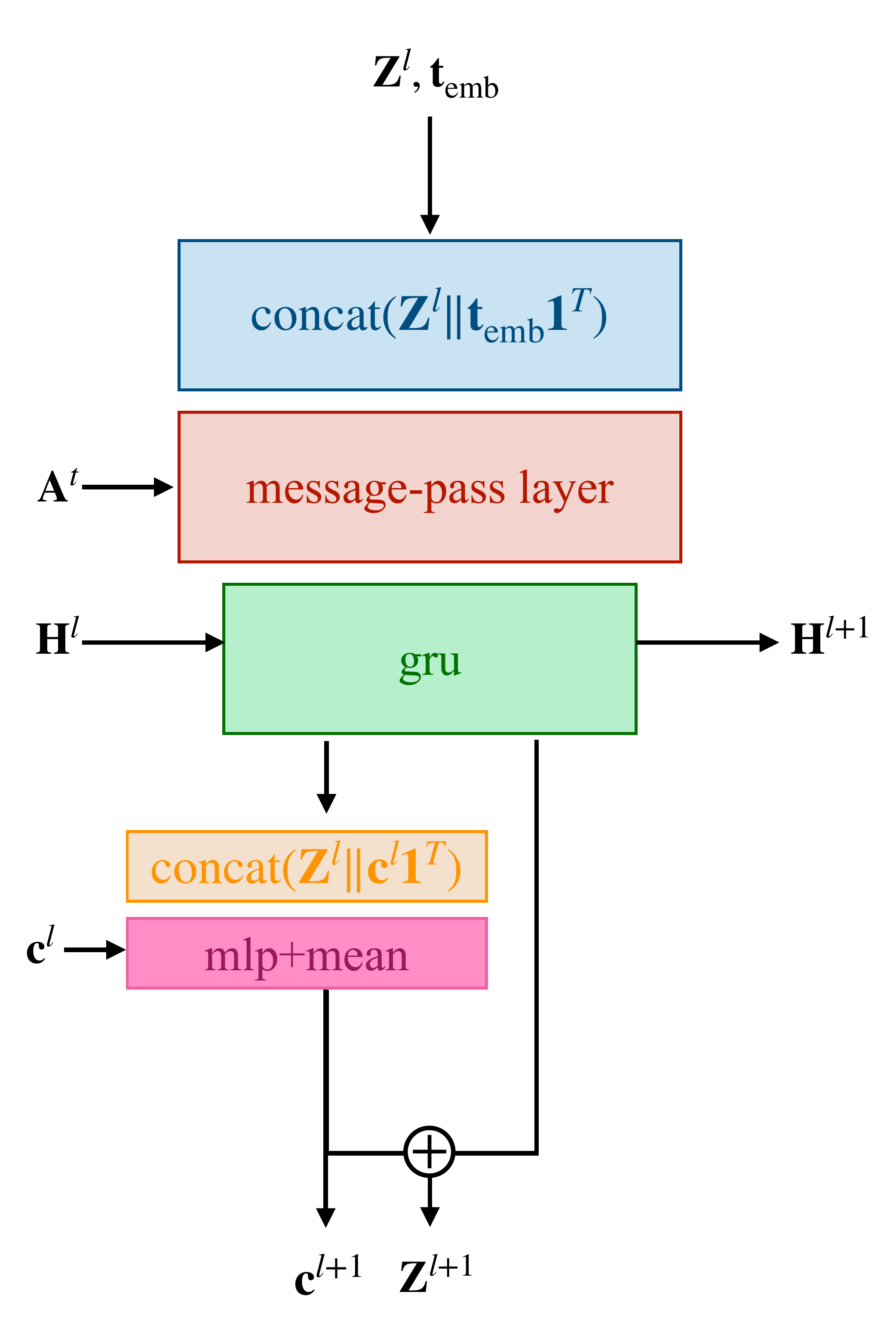} &
    \includegraphics[width=0.6\textwidth]{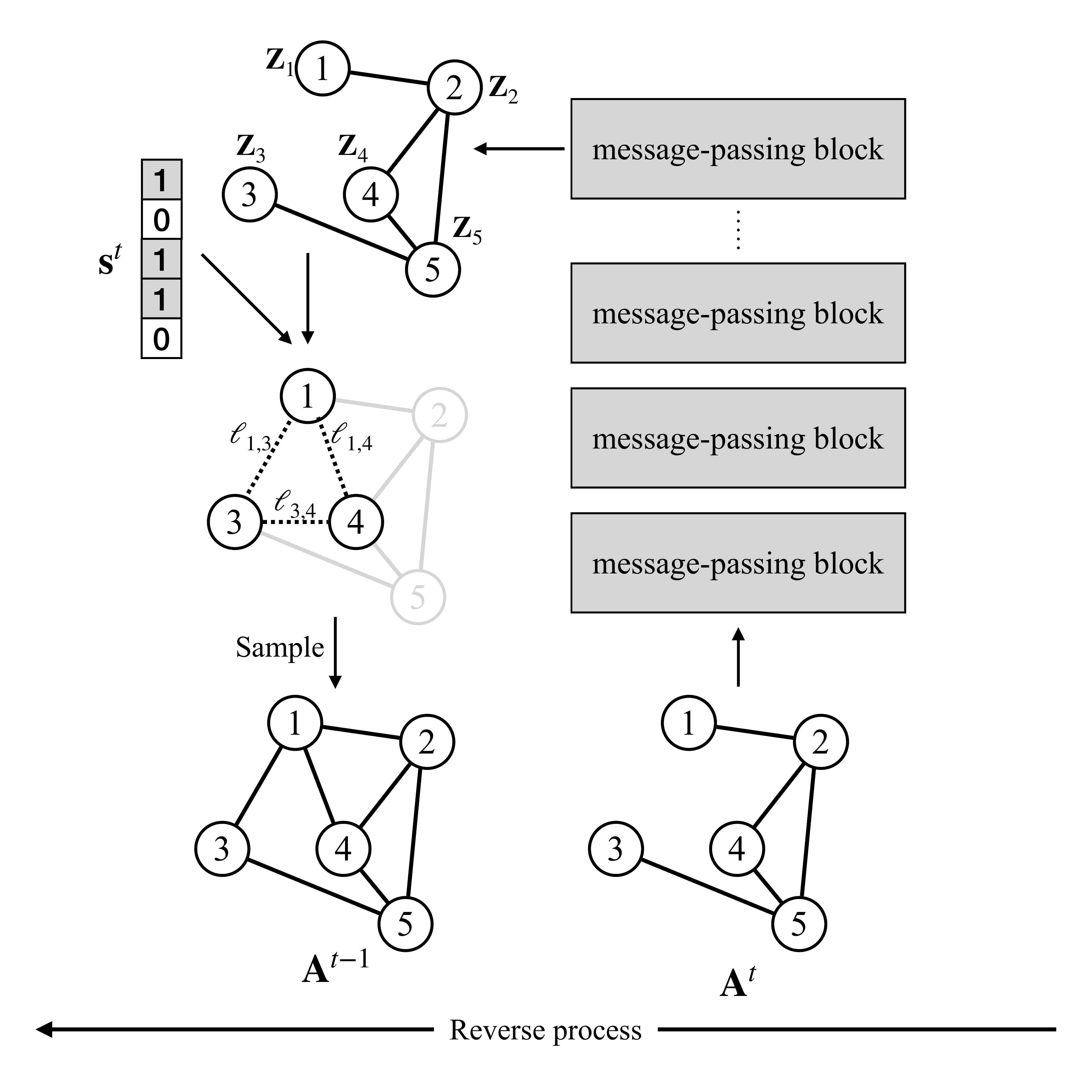} \\
    (a) Message-passing blocks & (b) Edge prediction of one step
    \end{tabular}
    \caption{Parameterization and inference of the edge prediction network. (a) shows the architecture of an MPB, which updates the node features $\bZ$ and context feature $\bc$ via message-passing. The state vectors $\bH$ are used for keeping geometric information at different levels of layers. (b) shows how the model infers edges during the reverse process. The node features are computed using a stacked MPB. The model only predicts edges between active nodes indicated by $\bs^t$.}
    \label{fig:impl-edge-network}
\end{figure*}

\paragraph{Architecture design and complexity analysis.} Figure~\ref{fig:impl-edge-network} shows the parameterization and the inference procedure of the edge prediction model. The main component of the parameterized network consists of $L$ message-passing blocks (MPB). After constructing the inputs, we iteratively update the node features using MPBs, and then compute the Bernoulli parameter $\ell_{i,j}^{t-1}$ for node pairs $(i,j)$ using a Multilayer perceptron(MLP). The computation flow is shown below
\begin{align}
    &\bZ^0 = \mathrm{concat}(\mathrm{emb}(\bd^t)\|\mathrm{emb}(\bd^0)),~\bt_\mathrm{emb} = \mathrm{emb}(t),~\bc^0=\mathrm{mean}(\bZ^0);\nonumber \\
    &\bZ^{l},\bc^l,\bH^l = \mathrm{block}_{l}(\bZ^{l-1}, \bt_\mathrm{emb},~\bc^{l-1}, \bH^{l-1}),~\text{for}~l=1,\ldots, L;\nonumber\\
    &\ell_{i,j}^{t-1} = \mathrm{mlp}(\bZ_i^L + \bZ_j^L),~\text{where}~ (i,j) \in\{(i,j)|\bs_{i,j}^{t}=1, 1\leq i<j \leq N\}.\nonumber
\end{align}

Here $\bZ^0$ is the node features initialized using degree sequence $\bd^0$ and $\bd^t$; $\bc$ is the global context features; $\bH$ is the node hidden state features. We follow \citet{dhariwal2021diffusion} and use the sinusoidal position embedding for the diffusion timestep $t$. $\bZ$, $\bc$, and $\bH$ are iteratively updated by MPBs. Finally, the edge prediction parameter $\ell_{i,j}^{t-1}$ is computed from the two node embeddings.

Each MPB contains a message-passing layer (MPL) and a Gated Recurrent Unit (GRU)~\citep{cho2014properties}. We experimented on various MPLs and found Unified Message Passaging Model~\citep{shi2020masked} demonstrates better expressivity for our tasks. The details for the $l$-th MBP can be represented as follow
\begin{align}
    &\bZ^l=\mathrm{concat}(\bZ^l \| \bt_\mathrm{emb}\bone^T)\nonumber\\
    &\bZ^l=\mathrm{mpl}(\bZ^l, \bA^{t})\nonumber\\
    &\bZ^l,\bH^{l+1}=\mathrm{gru}(\bZ^l,\bH^l)\nonumber\\
    &\bc^{l+1}=\mathrm{mlp}(\mathrm{mean}(\mathrm{concat}(\bZ^l\|\bc^l\bone^T)))\nonumber\\
    &\bZ^{l+1}=\bZ^{l}+\bc^{l+1}\bone^T\nonumber
\end{align}
Note that in MPB, the message-passing operation has runtime complexity $O(M)$, and all other operations have runtime complexity $O(N)$ as they operate on node features. The total runtime complexity of an MPB module is $O(M)$. Another major computation comes from the edge probability predictions, which are decided by the size of the node-pair set $|\{(i,j)|\bs_{i,j}^{t}=1, i<j\}|=(\sum_i\bs_i^t)^2\leq K^2$.

\subsection{Modeling the degree sequence distribution}
\label{degree_distribution_modeling}
We want the generative model $p_\theta(\bd^0)$ is invariant on node permutation $\pi$, i.e., for all $\pi$, we have $p_\theta(\bd^0)=p_\theta(\pi(\bd^0))$.
Let $\bu\in\{0,\ldots,N_\mathrm{max}\}^{d_\mathrm{max}}$ be a histogram such that $\bu_k=\sum_{i=1}^{N}\bone[\bd^0_i=k]$ for $k=1,\ldots, d_\mathrm{max}$, and $d_\mathrm{max}$ is the maximum node degree in the graph training set. We consider modeling $q(\bu|\bA^0)$ instead of $q(\bd^0|\bA^0)$ as $\bu$ is invariant to node permutation. The training objective is then to fit $p_\theta(\bu)$ to the empirical data distribution $p_\mathrm{data}(\bu)$. We consider it a sequence modeling task. The model $p_\theta(\bu)$ has the following decomposition
\begin{align}
p_\theta(\bu)=p_\theta(\bu_1)\prod_{k=2}^{d_\mathrm{max}}p_\theta(\bu_k|\bu_{<k}).
\end{align}

We parameterize $p_\theta(\bu_k|\bu_{<k})$ with a RNN~\citep{schuster1997bidirectional}, where the RNN produces $N_\mathrm{max}$-dimensional softmax logits at each step. Once a sequence $\bu$ is sampled, the graph's corresponding size is known. Note that we only model the degree distribution for the generic graph generation task. We only have one graph for large network generation for each dataset. We use the node degrees of the training graph when sampling new graphs. We describe the sampling algorithm in Alg.~\ref{alg:sampling-degree}. 

\begin{algorithm}[H]
  \caption{Sampling the degree sequence $\bd^0$}
  \label{alg:sampling-degree}
  \begin{algorithmic}
    \STATE {\bfseries Input:} Maximum degree $d_\mathrm{max}$, degree sequence model $p_\theta(\bu)$.
    \STATE Draw $\bu_1\sim p_\theta(\bu_1)$, initialize $\bd^0 = \bone_{\bu_1}$.
    \FOR{$k=2\ldots,d_\mathrm{max}$}
        \STATE Initialize $\bu_k$-dimensional all-ones vector $\bone_{\bu_k}$, set $\bd^0 = \mathrm{concat}(\bd^0\|k\bone_{\bu_k})$.
        \STATE Draw $\bu_k\sim p_\theta(\bu_k|\bu_{<k})$
    \ENDFOR
    \STATE {\bfseries Output:} Degree sequence $\bd^0$
  \end{algorithmic}
\end{algorithm}

\subsection{Training}
\label{training}
We demonstrate the training of the edge prediction model $p_\theta(\bA^{t-1}|\bA^t,\bs^t, \bd^0)$. Since we do not have a closed-form computation of the expectation, the gradient $\nabla_\theta \calL(\bA^0;\theta,\bd^0)$ is estimated via Monte Carlo samples:
\begin{align}
    \nabla_\theta\calL(\bA^0;\theta,\bd^0) = \mathbb{E}_{t\sim\calU[1,T]}\Big[\mathbb{E}_{q(\bA^{t-1}|\bA^0)}\mathbb{E}_{q(\bA^{t}|\bA^{t-1})}\mathbb{E}_{q(\bs^{t}|\bA^{t-1},\bA^t)}\big[\log{p_\theta(\bA^{t-1}|\bA^{t},\bs^{t},\bd^0)}\big]\Big].
\end{align}

We describe the procedure of training the edge prediction model in Alg.~\ref{alg:training}

\begin{algorithm}[H]
  \caption{Training procedure for edge prediction model}
  \label{alg:training}
  \begin{algorithmic}
    \STATE {\bfseries Input:} Data distribution $p_\mathrm{data}(\bA^0)$, diffusion steps $T$, forward transition distributions $q(\bA^t|\bA^{t-1})$ and $q(\bA^t|\bA^0)$.
    \WHILE{training}
    \STATE Draw $\bA^0\sim p_\mathrm{data}(\bA^0)$, compute node degree $\bd^0 = \mathrm{deg}(\bA^0)$.
    \STATE Sample $t\sim\calU[1,T]$, and have $t-1$ immediately.
    \STATE Draw $\bA^{t-1} \sim q(\bA^{t-1}|\bA^0)$, then draw $\bA^{t}\sim q(\bA^t|\bA^{t-1})$.
    \STATE Obtain active node variable $\bs^t$ from $\bA^{t-1}, \bA^t$.
    \STATE Perform gradient descent over $\theta$ with gradient $\nabla_\theta\log{p_\theta(\bA^{t-1}|\bA^{t},\bs^{t},\bd^0)}$
    \ENDWHILE
  \end{algorithmic}
\end{algorithm}

    

\section{Further Discussion of Current Diffusion Graph Models}
\subsection{Technical differences of existing diffusion-based graph models}
Diffusion-based generative models~\citep{sohl2015deep, ho2020denoising, hoogeboom2021argmax, austin2021structured} have gained prominent attention in the graph generation community. Current diffusion graph models adopt continuous or discrete-variable diffusion. For example, GDSS~\citep{jo2022score} and EDP-GNN~\citep{niu2020permutation} use Gaussian transition kernels for continuous-variable diffusion~\citep{ho2020denoising}, but GDSS additionally considers node and edge features and uses a score-matching diffusion approach. On the other hand, DiscDDPM~\citep{haefeli2022diffusion} and DiGress~\citep{vignac2022digress} employ discrete-time, discrete-variable diffusion models~\citep{hoogeboom2021argmax, austin2021structured}, with the former being featureless and the latter including a denoising process for node and edge attributes.  

One limitation shared by all current diffusion graph models is suffering from the scalability issue: they all need $O(TN^2)$ running time because they need to make predictions for each node pair in each diffusion step. This limitation prevents diffusion graph models from generating large networks. The main advantage of EDGE is to reduce this complexity to $O(\min(K^2, M))$, with $K$ being the number of active nodes and $M$ being the number of edges. It greatly reduces the computation in the generation process and has shown promising results in large network generation tasks.
We further highlight the limitation and technical differences of different models in Table~\ref{tab:app-related-work}.
\begin{table}[H]
    \centering
    \small
    \begin{tabular}{lcccccc}\toprule
& Diffusion type& \makecell{Convergent \\ distribution}& \makecell{Conditional \\ generation}& \makecell{Featured graph \\ generation} & Runtime & Scalability \\\midrule
EDP-GNN& \makecell{Disc. time, \\Cont. var.} & $\calN(0,1)$ &&&$O(TN^2)$&\\
GDSS& \makecell{Cont. time, \\Cont. var.} & $\calN(0,1)$& &{\checkmark}& $O(TN^2)$&\\
DiscDDPM   & \makecell{Disc. time, \\Disc. var.} & $G(N,0.5)$&&& $O(TN^2)$&\\
DiGress    & \makecell{Disc. time, \\Disc. var.} & \makecell{Empirical \\ distribution} & \makecell{Gradient from \\a classifier} & {\checkmark}                  &$O(TN^2)$&\\
EDGE (ours) & \makecell{Disc. time, \\Disc. var.} & $G(N,0)$& degree sequence&& $O(T\max(M,K^2))$ & {\checkmark} \\\bottomrule
\end{tabular}
    \caption{Technical differences of different diffusion graph models. Here $T$ is the number of diffusion steps, $N$ is the number of nodes in a graph, $M$ is the number of edges in a graph, and $K$ is the maximum number of active nodes during the diffusion process.}
    \label{tab:app-related-work}
\end{table}




\section{Generalizing to Tasks of Generating Attributed Graphs}
\label{app:feature-graph-generation}
\subsection{Hierarchical generation}
Generation of attributed graphs has a broad class of applications, such as molecule generation~\citep{du2021graphgt}. While EDGE is developed to generate graph structure only, here we briefly discuss how it can be incorporated into a hierarchical procedure to generate graphs with node and edge attributes. Here we consider the case where node and edge attributes are both categorical.

The attributes of nodes and edges are represented as one-hot vectors. For node attributes, we have a matrix $\bX \in \{0,1\}^{N\times \calC_\text{node}}$, while edge attributes are described by the matrix $\bA_\text{attr} \in \{0, 1\}^{N\times (\calC_\text{edge}+1)}$. In this context, $\calC_\text{node}$ and $\calC_\text{edge}$ denote the number of classes for node types and edge types, respectively. For a node pair $(i,j)$, the extra dimension indicates whether the edge exists or not. The graph structure is still denoted by $\bA$. Inspired by~\citet{lippe2020categorical}, we consider the following joint model:
\begin{align}
    p(\bX,\bA_\text{attr},\bA) = p(\bX)p(\bA|\bX)p(\bA_\text{attr}|\bX,\bA), \label{eq:joint}
\end{align}
which can be considered a hierarchical generation scheme that first samples node attributes, then samples the graph structure via EDGE conditioned on node attributes, and finally samples edge attributes conditioned on the graph and node attributes. 

\subsection{Model Details}
We consider modeling each component in Eqn.~(\ref{eq:joint}) separately.
For $p(\bX)$, we employ a similar approach as with the node degree sequence modeling, but we use the sequence length $\calC_\text{node}$ instead of $d_\mathrm{max}$. For $p(\bA|\bX)$, we apply the EDGE framework, incorporating node features from $\bX$ during both the training and generation phases. For $p(\bA_\text{attr}|\bX,\bA)$, we utilize a diffusion model that starts by randomly assigning edge types to edges in $\bA$ and iteratively refines edge labels, relying  on the information given by $\bX$ and $\bA$. It is important to note that we only refine labels for edges already specified by $\bA$, allowing us to use an MPNN to calculate edge features. We adopt the framework outlined in Appendix~\ref{app:arch}, and only perform prediction for   edges existing in $\bA$.

\section{Additional Details for Experimental Setups}
\label{detail_experiment}
We described the details of the experiments of generic graph generation and large network generation tasks. We provide the hyperparameters used in the experiments in Table~\ref{tab:hyperparmeters}. We do not augment the data input with extra features for all generation tasks except for the current node degrees $\bd^t$ and the node degrees $\bd^0$, which are both computation-free. Moreover, we set $p=10^{-12}$ in our implementation to maintain numerical stability.
\begin{table}[h]
\centering
\begin{tabular}{lcccccc}\toprule
                              \multicolumn{1}{c}{}       & Community       & Ego         & Polblogs   & Cora   & Road-Minnesota   & PPI       \\\midrule
                              \textbf{Diffusion}\\
                              Diffusion steps T          & 128             & 128         & 256        & 64      &64  & 512       \\
                              Noise scheduling           & \multicolumn{6}{c}{Linear}                                         \\
                              $\beta_0$    & \multicolumn{2}{c}{$7.8125\times10^{-4}$} & $3.9063\times10^{-4}$ & \multicolumn{2}{c}{$1.5625\times10^{-3}$} & $1.9531\times10^{-4}$\\
                              $\beta_T$    & \multicolumn{2}{c}{$1.5625\times10^{-1}$} & $7.8125\times10^{-2}$  & \multicolumn{2}{c}{$3.1250\times10^{-1}$}  & $3.9063\times10^{-2}$ \\
                              Sample time method         & \multicolumn{6}{c}{Importance sampling}                            \\\midrule
                              \textbf{Optimization}\\
                              Learning rate              & \multicolumn{6}{c}{$10^{-4}$}                                         \\
                              Optimizer                  & \multicolumn{6}{c}{Adam~\citep{kingma2014adam}}                                           \\
                              weight decay               & \multicolumn{5}{c}{$10^{-4}$}                                           \\
                              Batch size                 & 64              & 64          & 4          & 4     &4    & 1         \\
                              Number of epochs/iteration & 30000           & 10000       & 50000      & 50000   & 50000  & 50000     \\
                              \midrule
                              \textbf{Architecture}\\
                              Number of MPBs             & \multicolumn{6}{c}{5}                                              \\
                              Hidden dimension of MPL    & \multicolumn{6}{c}{64}                                             \\
                              Hidden dimension of GRU    & \multicolumn{6}{c}{64}                                             \\
                              Activation function         & \multicolumn{6}{c}{SiLU~\citep{elfwing2018sigmoid}}                                          \\
                              Time embedding             & \multicolumn{6}{c}{Sinusoidal positional embedding~\citep{devlin2018bert}}       \\
                              Dropout rate               & \multicolumn{6}{c}{0.1}                                            \\\midrule
                              \textbf{Evaluation}\\
                              Number of generated graphs & 128             & 128         & 5          & 5   & 5      & 5         \\
                              $d_\mathrm{max}$           & 40              & 100         & 351        & 168    & 5   & 593       \\
                              Number of attention heads            & 8               & 8           & 8          & 8      &8   & 8        \\\bottomrule
\end{tabular}
\caption{Hyperparameters}\vspace{-1em}
\label{tab:hyperparmeters}
\end{table}

\subsection{Generic graph generation}
\label{generic graph generation}
We follow \citet{you2018graphrnn} to generate the Community and Ego datasets and use the same data splitting strategy. Recent works~\citep{o2021evaluation, thompson2022evaluation} have suggested better metrics for evaluating the quality of the generated graphs. To make a fair comparison, we reproduce all baselines and follow ~\citet{thompson2022evaluation} to re-evaluate their generative performance. All the baselines are reproduced using their default hyperparameter setting except for GraphCNF and DiGress. For GraphCNF, we use the same model configuration of its molecule generation task for the Community dataset and a smaller model for the Ego dataset due to the limited capacity of the GPU memory. For DiGress, we do not augment the graphs with the structural features to ensure a fair comparison is made.

\subsection{Large network generation}
\label{large graph generation}
We consider the single network for each large network dataset as the training dataset. Since the evaluation metrics do not require referring to the test graphs, we do not include validation/test sets in this task. All models are trained until the network statistics converge, and the models of the final epoch are used to generate samples. For GraphRNN, we use the default BFS ordering to generate adjacency matrices for the model training. We train the model for 30000 iterations for all datasets and report the model performance using the checkpoint from the last epoch.

\paragraph{Computing the Edge Overlap.} Since GraphRNN and our model are edge non-independent models, \citet{chanpuriya2021power} suggests reporting the maximum edge overlap between the generated graphs and the training graph to ensure the models do not simply memorize the data. However, finding the maximum edge overlap requires searching over the node permutation space, which is impractical as there are $N!$ permutations. Instead, we obtain the node degree ascending permutation and use it to permute both the generated and training graphs. We observe that such a permutation scheme yields a much higher EO value than a random permutation. For instance, when a model can generate graphs with desiring statistics, degree-based permutation yields 15\% EO on average for the Poblogs dataset, while a random permutation yields an EO value that is almost 0.

\subsection{Computational Resources}
\label{computational resource}
We use PyTorch~\citep{paszke2019pytorch} and PyTorch Geometric~\citep{fey2019fast} to implement our framework. We train our models on Tesla A100, Tesla V100, or NVIDIA QUADRO RTX 6000 GPU and 32 CPU cores for all experiments.  For generic graph generation tasks, all models are trained within 72 hours. For large network generation tasks, model training is finished within 24 hours. The sampling speed reported in Figure~\ref{fig:speed} of all baselines and our approach is tested on Tesla A100 GPU.

\section{Extended Results}
\subsection{Comparison between the node degrees from generated graphs and the node degrees $\bd^0$}
\label{node degree comparison}
\begin{figure}[h]
    \centering
    \begin{tabular}{cc}
        \includegraphics[width=0.4\textwidth]{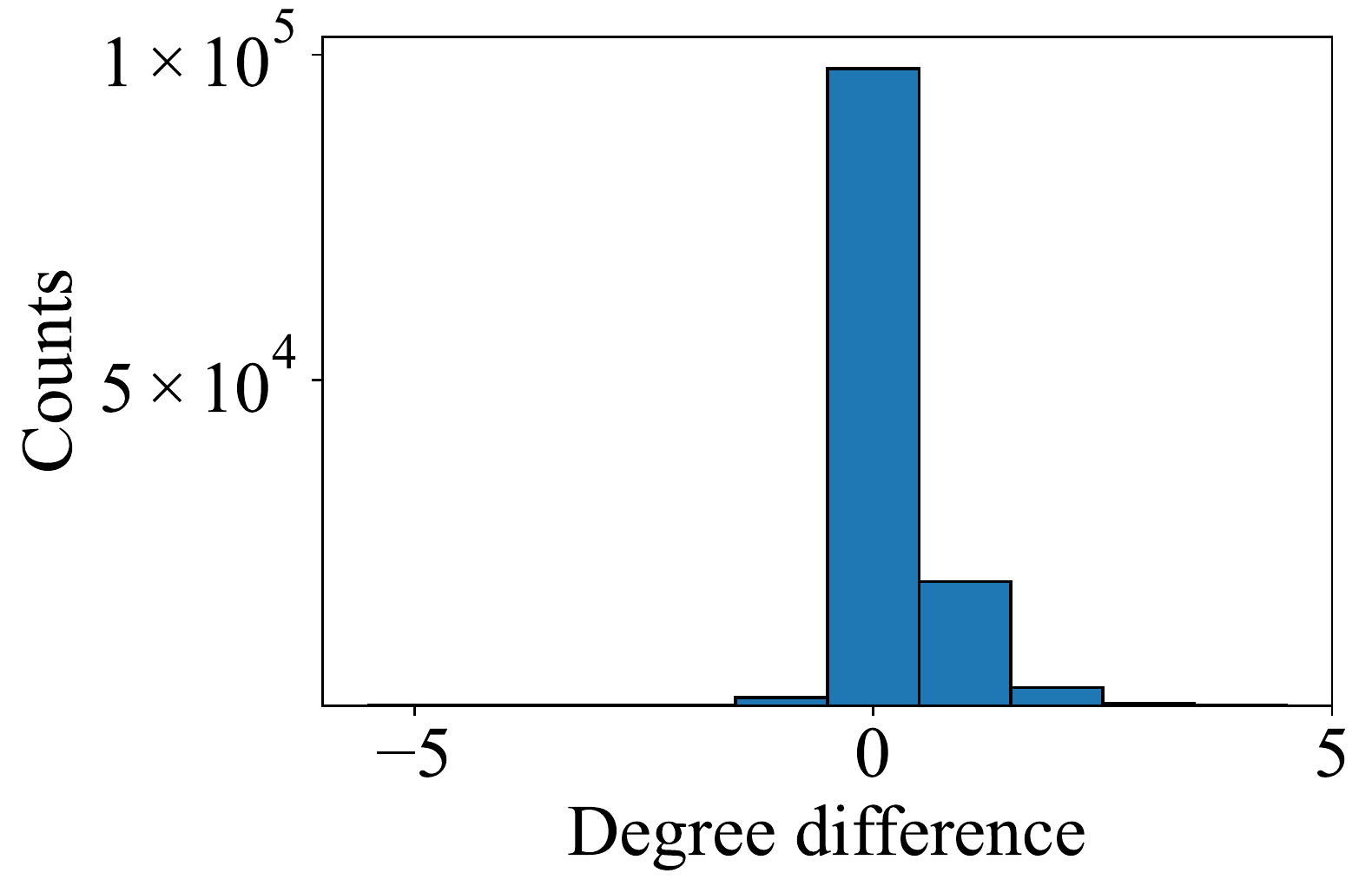} & \includegraphics[width=0.4\textwidth]{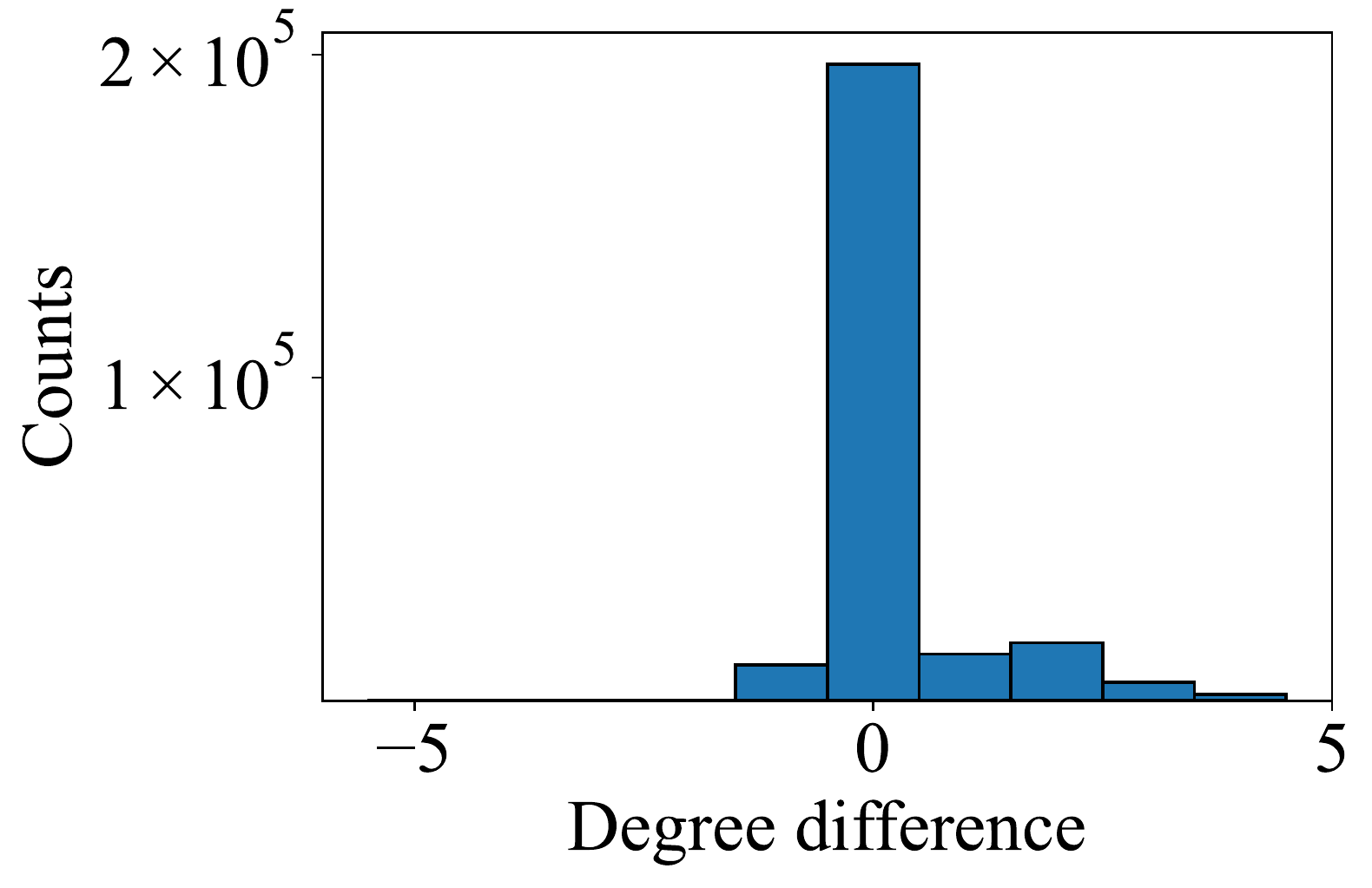} \\ 
         Polblogs & Cora \\
         \includegraphics[width=0.4\textwidth]{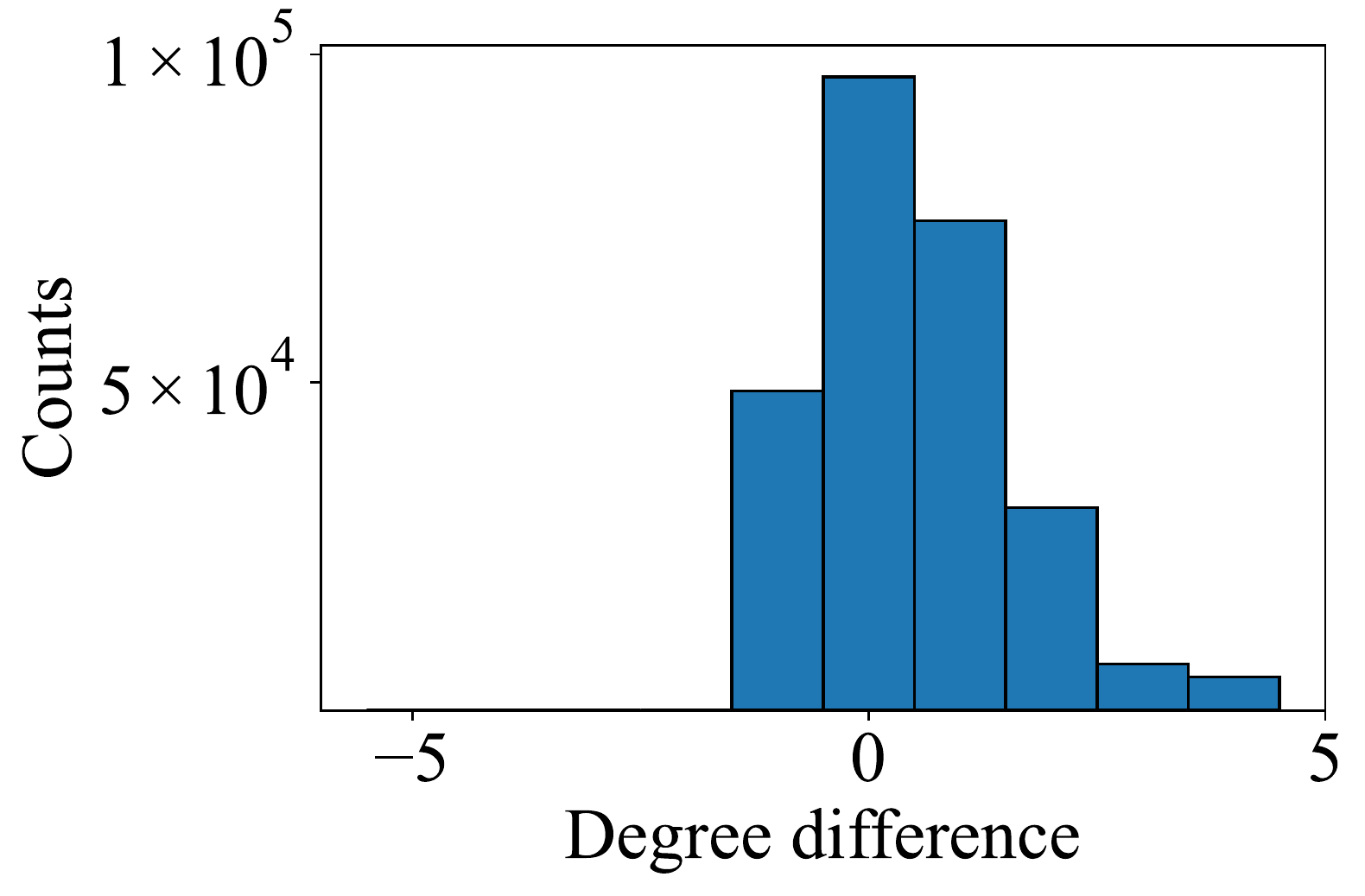} & \includegraphics[width=0.4\textwidth]{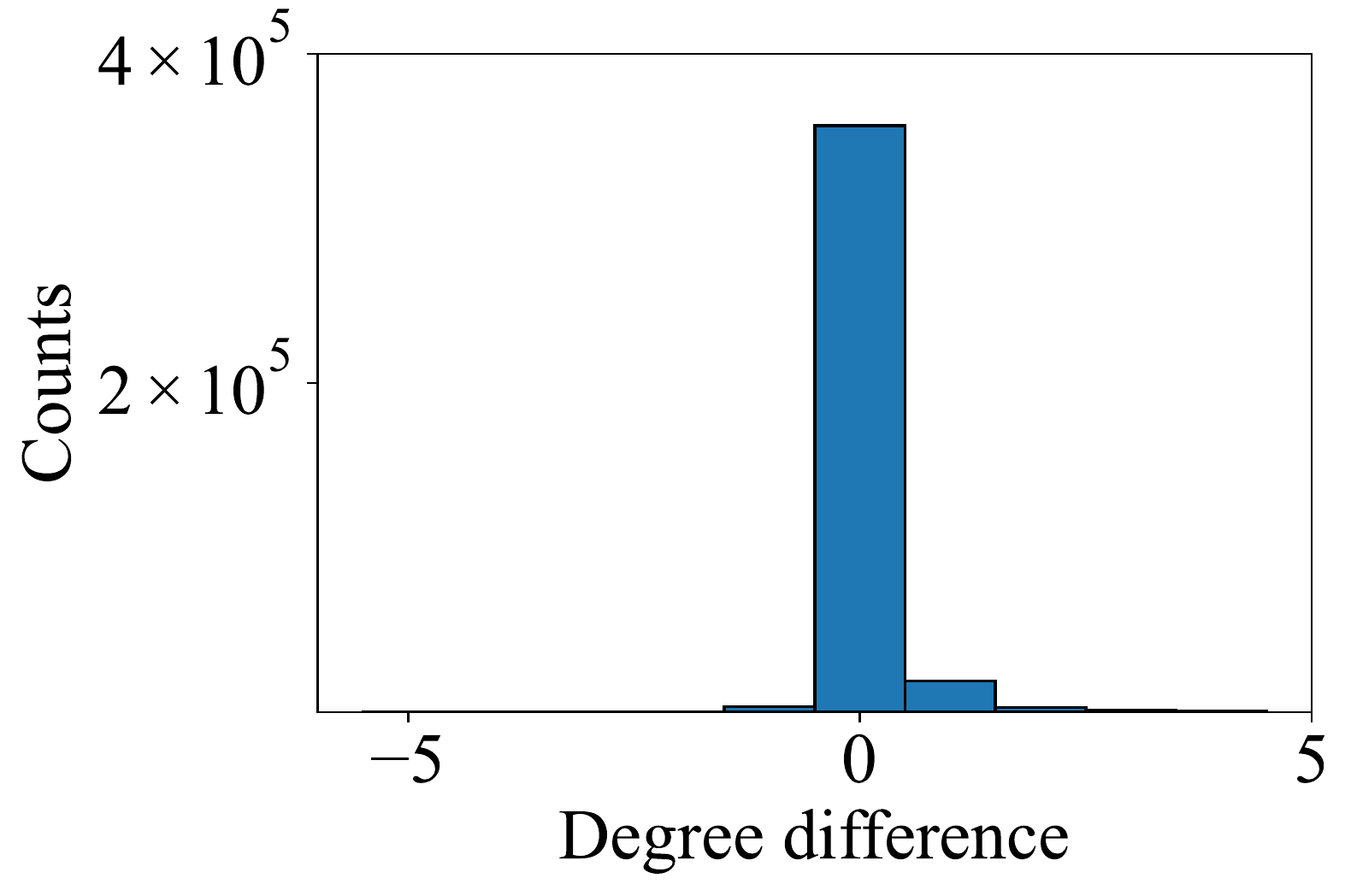} \\
         Road-Minnesota& PPI
    \end{tabular}
    \caption{Degree difference. Given specific node degrees $\bd^0$, the actual node degrees of the generated graphs are fairly accurate.}
    \label{fig:degree-approx}
\end{figure}

We show that the generated graphs' node degrees accurately approximate the given node degrees $\bd^0$. The node degrees in the generated graphs are compared to the node degrees $\bd^0$ by counting the number of nodes whose degree deviates from the given one. The degree difference is computed by subtracting the given degree from the actual degree. The histograms in Figure~\ref{fig:degree-approx} display the degree difference for each dataset, indicating the accuracy of the generated graph's node degrees in approximating the given node degrees.

\subsection{Further justification the use of $G(N,0)$ as the convergent distribution}

In addition to the desirable properties we described in Section \ref{sec:g(n,0)}, we demonstrate the potential benefit of using $G(N,0)$ as the convergent distribution in terms of generative performance. When using $G(N,0)$ as the convergent distribution, our proposed framework can be considered as a type of absorbing diffusion process~\citep{austin2021structured}. Similar to \citep{austin2021structured}, we observe that the generative performance of $G(N,0)$ is superior to $G(N,0.5)$. Table~\ref{tab:caveman-small} report the generative performance of $G(N,0.5)$ and $G(N,0)$ on the Community-small dataset~\citep{you2018graphrnn}. This demonstrates the superiority of using an absorbing state as the convergent distribution, which further justifies why one should consider $G(N,0)$ as the convergent distribution.

\begin{table}[H]
    \centering
    \begin{tabular}{lccccc}\toprule
    &\multicolumn{5}{c}{Community-small}\\
    &\multicolumn{3}{c}{Structure-based metrics (MMD)} & \multicolumn{2}{c}{Neural-based metrics}\\
& Deg. &Clus. &	Orb.& FID & RBF MMD\\\midrule
$G(N,0.5)$ & 0.0274 & 0.0249 & \textbf{0.0234}& 3.4121 &0.0243\\
$G(N,0)$ & \textbf{0.0081} & \textbf{0.0112}& 0.0262 & \textbf{1.5642} & \textbf{0.0204}\\\bottomrule
    \end{tabular}
    \caption{Vanilla discrete diffusion with $G(N,0.5)$ and $G(N,0)$ as the convergent distributions. $G(N,0)$ exhibits better generative performance than $G(N,0.5)$. }
    \label{tab:caveman-small}
\end{table}

\subsection{Full results on graph generation tasks}
We provide the mean and the standard derivation of metrics reported in the generic graph generation and large network generation tasks in Table~\ref{tab:app-generic-graph} and Table~\ref{tab:app-large-network}, respectively.

\begin{table*}[h]
    \centering 
    \begin{tabular}{lccccc}\toprule
& \multicolumn{5}{c}{Community}
 \\
 & \multicolumn{3}{c}{Structure-based metrics (MMD)} & \multicolumn{2}{c}{Neural-based metrics}
 \\
&Deg.	& Clus. & Orb. & FID	& RBF MMD
\\\midrule
GRNN & 0.1440 $\pm$ 0.0025 &	\underline{0.0535 $\pm$ 0.0264} &	\textbf{0.0198 $\pm$ 0.0003} &	8.3869 $\pm$ 1.5429 &	0.1591 $\pm$ 0.0104
\\
GRAN & 0.1022 $\pm$ 0.0185 & 0.0894 $\pm$ 0.0082& \textbf{0.0198 $\pm$ 0.0005} & 64.1145 $\pm$ 12.0927 & 0.0749 $\pm$
 0.0097\\\midrule					
GraphCNF & 0.1129 $\pm$ 0.0295 & 1.2882 $\pm$ 0.1918 & \textbf{0.0197 $\pm$ 0.0005} & 29.1526 $\pm$ 3.1900 & 0.1341 $\pm$ 0.0241
\\
GDSS & {0.0535 $\pm$ 0.0095} & 0.2072 $\pm$ 0.0520 & \textbf{0.0196 $\pm$ 0.0003} & {6.5531 $\pm$ 0.9418} & \underline{0.0443 $\pm$ 0.0058}
\\
DiscDDPM &0.1238 $\pm$ 0.0068 & 0.6549 $\pm$ 0.0463 & \underline{0.0246 $\pm$ 0.0004} & 8.6321 $\pm$ 1.1961 & 0.0840 $\pm$ 0.0099
\\
DiGress &\underline{0.0409 $\pm$ 0.0041} & \textbf{0.0167 $\pm$ 0.0169} & 0.0298 $\pm$ 0.0002 & \underline{3.4261 $\pm$ 0.4549} & \underline{0.0460 $\pm$ 0.0069}
\\\midrule
EDGE &	\textbf{0.0175 $\pm$ 0.0056} &	\underline{0.0689 $\pm$ 0.0197} &	\textbf{0.0198 $\pm$ 0.0002} &	\textbf{2.2378 $\pm$ 0.5111} &	\textbf{0.0227 $\pm$ 0.0097}
\\\bottomrule\toprule
& \multicolumn{5}{c}{Ego} 
 \\
 & \multicolumn{3}{c}{Structure-based metrics (MMD)} & \multicolumn{2}{c}{Neural-based metrics}
 \\
&Deg.	& Clus. & Orb. & FID	& RBF MMD
\\\midrule
GRNN & \underline{0.0768 $\pm$ 0.0142} & 1.1456 $\pm$ 0.0910 &	0.1087 $\pm$ 0.0442 &	90.5655 $\pm$ 19.2041 &	0.6827 $\pm$ 0.1181
\\
GRAN & 0.5778 $\pm$ 0.1415 & {0.3360 $\pm$ 0.0948} & \textbf{0.0406 $\pm$ 0.0112} & 489.9598	$\pm$ 42.1109 & {0.2633 $\pm$ 0.0911}
\\\midrule		
GraphCNF& 0.1010 $\pm$ 0.0421 &	0.7654 $\pm$ 0.0510 & {0.0820 $\pm$ 0.0334} &\underline{18.7929 $\pm$ 3.5102}	& \underline{0.0896 $\pm$ 0.0125}
\\
GDSS& 0.8189 $\pm$ 0.0691&	0.6032 $\pm$ 0.2114 &	0.3315 $\pm$ 0.0591&	60.6100 $\pm$ 8.1208 & 0.4331 $\pm$ 0.0982
\\
DiscDDPM & 0.4613 $\pm$ 0.1042 & \underline{0.1681  $\pm$ 0.0735} & \underline{0.0633 $\pm$ 0.0156} & 42.7994 $\pm$ 5.6312& 0.1561 $\pm$ 0.0224
\\
DiGress & \underline{0.0708 $\pm$ 0.0127} & \textbf{0.0092 $\pm$ 0.0062} & 0.1205 $\pm$ 0.0669 & \underline{18.6794 $\pm$ 4.6395} & \textbf{0.0489 $\pm$ 0.0232}
\\\midrule
EDGE &	\textbf{0.0579 $\pm$ 0.0101} & \underline{0.1773 $\pm$ 0.0521} & \textbf{0.0519 $\pm$ 0.0216} & \textbf{15.7614 $\pm$ 2.5021} &	\textbf{0.0658 $\pm$ 0.0199} 
\\\bottomrule
    \end{tabular}
    \caption{Generation performance on generic graphs with standard derivation.}
    \label{tab:app-generic-graph}
\end{table*}
\begin{table}[h]
    \centering
    \begin{tabular}{lcccccc}\toprule
    \multicolumn{7}{c}{Polblogs}\\
&EO&PLE&NTC&CC& CPL& AC\\\midrule
TRUE&	100&	1.414	&1&	0.226	&2.736	&-0.221\\\midrule
OPB	& 24.5 $\pm$ 0.4	&\underline{1.395 $\pm$ 0.002}&	0.667 $\pm$ 0.013&	0.150 $\pm$ 0.001&	2.524 $\pm$ 0.005&	-0.143 $\pm$ 0.003\\
HDOP&16.4 $\pm$ 0.3	&1.393 $\pm$ 0.003&	0.687 $\pm$ 0.021&	0.153 $\pm$ 0.002&	2.522 $\pm$ 0.009&	-0.131 $\pm$ 0.006\\
CELL&26.8 $\pm$ 0.2&	1.385 $\pm$ 0.001&	0.810 $\pm$ 0.011&	\underline{0.211 $\pm$ 0.002}&	2.536 $\pm$ 0.006&	\underline{-0.230 $\pm$ 0.002}\\
CO  &20.1 $\pm$ 0.2&	1.975 $\pm$ 0.107&	0.045 $\pm$ 0.002&	0.028 $\pm$ 0.001&	2.502 $\pm$ 0.008&	0.068 $\pm$ 0.009\\
TSVD&32.0 $\pm$ 0.2&	1.373 $\pm$ 0.001&	\underline{0.872 $\pm$ 0.023}&	0.205 $\pm$ 0.004&	2.533 $\pm$ 0.005&	\textbf{-0.216 $\pm$ 0.005}\\
VGAE & 3.6 $\pm$ 0.2 & 1.723 $\pm$ 0.010 & 0.05 $\pm$ 0.006 & 0.001 $\pm$ 0.001 & 2.531 $\pm$ 0.063 & -0.086 $\pm$ 0.009\\\midrule
GRNN&9.6 $\pm$ 0.5	&1.334 $\pm$ 0.013&	0.355 $\pm$ 0.048&	0.095 $\pm$ 0.008&	\underline{2.566 $\pm$ 0.056}&	0.096 $\pm$ 0.065\\\midrule
EDGE&16.5 $\pm$ 0.3&	\textbf{1.398 $\pm$ 0.002}&	\textbf{0.977 $\pm$0.079}	&\textbf{0.217 $\pm$ 0.005}	&\textbf{2.647 $\pm$ 0.028}	&\textbf{-0.214 $\pm$ 0.015}\\\bottomrule\toprule
    \multicolumn{7}{c}{Cora}\\
&EO&PLE&NTC&CC& CPL& AC\\\midrule
TRUE &  100 & 1.885 & 1 & 0.090 & 6.311 & -0.071\\\midrule
OPB	    &10.9 $\pm$ 0.2 & \textbf{1.852 $\pm$ 0.008} & 0.097 $\pm$	0.019 & 0.008 $\pm$ 0.001 & 4.476 $\pm$ 0.046 & \underline{-0.037 $\pm$ 0.009}\\
HDOP	&0.9 $\pm$ 0.1 & \textbf{1.849 $\pm$ 0.011} & 0.113 $\pm$	0.003 & 0.009 $\pm$ 0.001 & 4.477 $\pm$ 0.030 & \underline{-0.030 $\pm$ 0.004}\\
CELL	&10.3 $\pm$ 0.2 & 1.774 $\pm$ 0.001 & 0.009 $\pm$	0.003 & 0.002 $\pm$ 0.001 & \underline{5.799 $\pm$ 0.012} & -0.018 $\pm$ 0.013\\
CO      &9.7 $\pm$ 0.5 & 1.776 $\pm$ 0.007 & 0.009 $\pm$ 0.002 & 0.002 $\pm$ 0.000 & 5.653 $\pm$ 0.044 & 0.010 $\pm$ 0.012\\
TSVD	&6.7 $\pm$ 0.2 & \textbf{1.858 $\pm$ 0.012} & \underline{0.349 $\pm$	0.029} & \underline{0.028 $\pm$ 0.001} & 4.908 $\pm$ 0.052 & -0.006 $\pm$ 0.005\\
VGAE & 1.5 $\pm$ 0.5 & 1.717 $\pm$ 0.005 & 0.120 $\pm$ 0.012 & 0.220 $\pm$ 0.012 & 4.934 $\pm$ 0.069 & 0.002 $\pm$ 0.010 \\\midrule
GRNN	&0.4 $\pm$ 0.1 & \underline{1.822 $\pm$ 0.008} & 0.043 $\pm$	0.007 & 0.011 $\pm$ 0.002 & \textbf{6.146 $\pm$ 0.065} & 0.043 $\pm$ 0.025\\\midrule
EDGE	&0.9 $\pm$ 0.0 & 1.755 $\pm$ 0.005 & \textbf{0.446 $\pm$	0.029} & \textbf{0.034 $\pm$ 0.002} & 4.995 $\pm$ 0.048 & \textbf{ -0.046 $\pm$ 0.008}\\\bottomrule\toprule

    \multicolumn{7}{c}{Road-Minnesota}\\
&EO&PLE&NTC&CC& CPL& AC\\\midrule
TRUE    & 100  & 2.147 & 1 & 0.028 & 35.349 & -0.187\\\midrule
OPB	    & 29.7 $\pm$ 0.3 & \textbf{2.188 $\pm$ 0.016} & 0.083 $\pm$ 0.036 & 0.002 $\pm$ 0.001 & 8.036 $\pm$ 0.051& 0.009 $\pm$ 0.011\\
HDOP    & 13.2 $\pm$ 1.1 & \textbf{2.192 $\pm$ 0.065} & \underline{0.208 $\pm$ 0.111} & 0.004 $\pm$ 0.001 & 8.274 $\pm$ 0.032& \underline{-0.024 $\pm$ 0.006}\\
CELL	& 30.7 $\pm$ 1.3 & 2.267 $\pm$ 0.011 & 0.053 $\pm$ 0.069 & 0.001 $\pm$ 0.001 & 10.219 $\pm$ 0.096& \textbf{-0.082 $\pm$ 0.00}4\\
CO      & 19.8 $\pm$ 0.9 & \underline{2.044 $\pm$ 0.049} & 2.845 $\pm$ 0.916 & \textbf{0.040 $\pm$ 0.003} & \underline{11.478 $\pm$ 0.075}& -0.012 $\pm$ 0.008\\
TSVD	& 19.4 $\pm$ 0.6 & \textbf{2.172 $\pm$ 0.041} & 0.060 $\pm$ 0.046 & 0.001 $\pm$ 0.000 & 8.431 $\pm$ 0.130& 0.006 $\pm$ 0.009\\
VGAE	& 1.3 $\pm$ 0.3 & {1.678 $\pm$ 0.091} & 0.096 $\pm$ 0.031 & 0.009 $\pm$ 0.001 & 11.120 $\pm$ 0.075& \underline{-0.027 $\pm$ 0.001}\\\midrule
GRNN    & 0.6 $\pm$ 0.1 & 1.570 $\pm$ 0.017 & 0.099 $\pm$ 0.023 & 0.007 $\pm$ 0.002 & \textbf{11.695 $\pm$ 0.059} & 0.006 $\pm$ 0.009\\\midrule
EDGE    & 0.8 $\pm$ 0.1 & {1.910 $\pm$ 0.023} & \textbf{0.962 $\pm$ 0.101} & \underline{0.011 $\pm$ 0.001} & {9.125 $\pm$ 0.088}& \underline{-0.063 $\pm$ 0.006}
\\\bottomrule\toprule

    \multicolumn{7}{c}{PPI}\\
&EO&PLE&NTC&CC& CPL& AC\\\midrule
TRUE    &100 & 1.462	&1    &0.092 &  3.095 &-0.099\\\midrule
OPB	    &16.3 $\pm$ 0.2&	\underline{1.443 $\pm$ 0.001}	&0.640 $\pm$ 0.007&   0.058 $\pm$ 0.000&   2.914 $\pm$ 0.005&  \textbf{-0.089 $\pm$ 0.003}\\
HDOP    &6.9 $\pm$ 0.1&	\underline{1.444 $\pm$ 0.001}	&0.638 $\pm$ 0.007&   0.058 $\pm$ 0.001&   2.917 $\pm$ 0.008&  \underline
{-0.086 $\pm$ 0.003}\\
CELL	&6.7 $\pm$ 0.2&	1.400 $\pm$ 0.000	&0.248 $\pm$ 0.005&	 0.040 $\pm$ 0.001 &  \textbf{3.108 $\pm$ 0.003} & 0.176 $\pm$ 0.004\\
CO      &9.9 $\pm$ 0.1&	1.754 $\pm$ 0.071	&0.016 $\pm$ 0.001&   0.006 $\pm$ 0.000&   \underline{3.046 $\pm$ 0.002} &	0.043 $\pm$ 0.004\\
TSVD	&13.2 $\pm$ 0.1&  {1.426 $\pm$ 0.001}&	\underline{0.848 $\pm$ 0.015}&	 \underline{0.077 $\pm$ 0.001}&   2.867 $\pm$ 0.004&	\textbf{-0.089 $\pm$ 0.004}\\
VGAE & 0.5 $\pm$ 0.0 & 1.362 $\pm$ 0.006 & 0.091 $\pm$ 0.009 & 0.012 $\pm$ 0.005 & 2.991 $\pm$ 0.063 & 0.054 $\pm$ 0.007 \\\midrule
GRNN	&OOM&OOM&OOM&OOM&OOM&OOM\\\midrule
EDGE & 7.5 $\pm$ 0.4 & \textbf{1.449 $\pm$ 0.003} &\textbf{0.981 $\pm$ 0.003} & \textbf{0.091 $\pm$ 0.031} &	\underline{3.028 $\pm$ 0.044} & \textbf{-0.107 $\pm$ 0.023}\\\bottomrule
    \end{tabular}
    \caption{Generation performance on large networks with standard derivation.}
    \label{tab:app-large-network}
\end{table}

\subsection{Visualizations}
\begin{figure}[h]
    \centering
    \begin{tabular}{cccccc}
\includegraphics[width=0.13\textwidth]{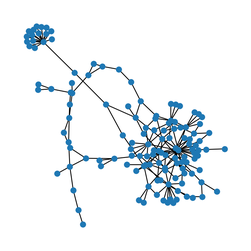} & 
\includegraphics[width=0.13\textwidth]{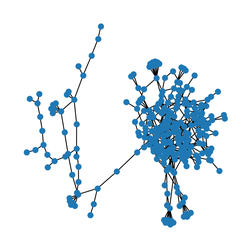} & 
\includegraphics[width=0.13\textwidth]{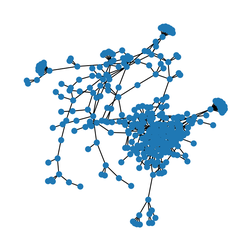} & 
\includegraphics[width=0.13\textwidth]{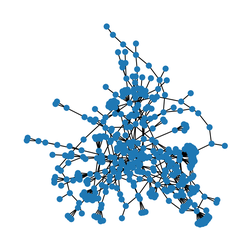} & 
\includegraphics[width=0.13\textwidth]{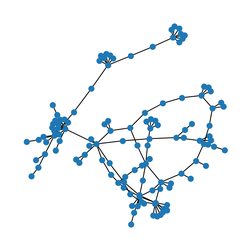} & 
\includegraphics[width=0.13\textwidth]{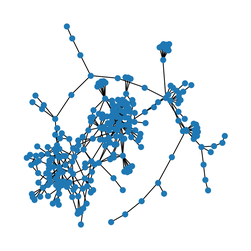} \\
\multicolumn{6}{c}{GRNN}\\

\includegraphics[width=0.13\textwidth]{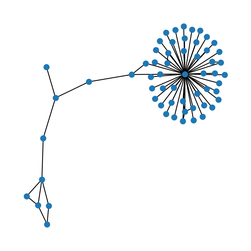} & 
\includegraphics[width=0.13\textwidth]{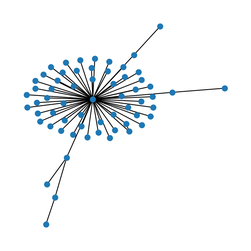} & 
\includegraphics[width=0.13\textwidth]{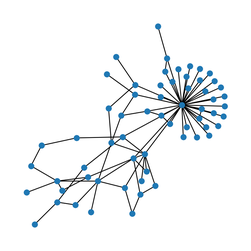} & 
\includegraphics[width=0.13\textwidth]{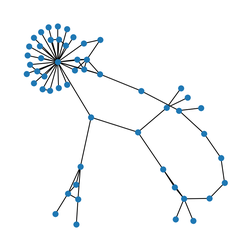} & 
\includegraphics[width=0.13\textwidth]{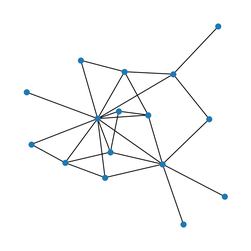} & 
\includegraphics[width=0.13\textwidth]{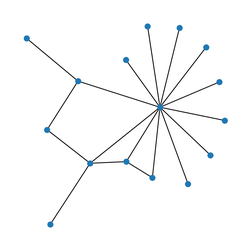} \\
\multicolumn{6}{c}{GRAN}\\

\includegraphics[width=0.13\textwidth]{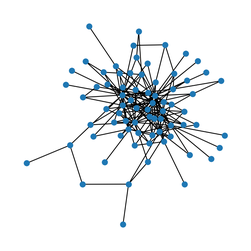} & 
\includegraphics[width=0.13\textwidth]{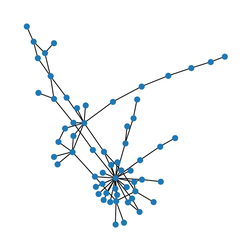} & 
\includegraphics[width=0.13\textwidth]{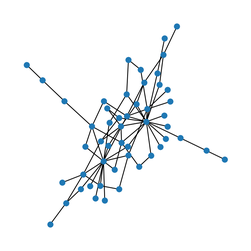} & 
\includegraphics[width=0.13\textwidth]{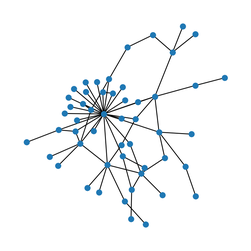} & 
\includegraphics[width=0.13\textwidth]{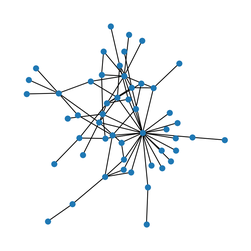} & 
\includegraphics[width=0.13\textwidth]{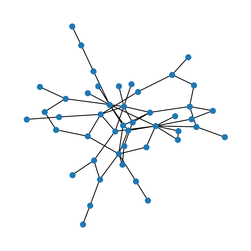} \\
\multicolumn{6}{c}{GraphCNF}\\

\includegraphics[width=0.13\textwidth]{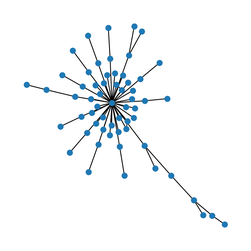} & 
\includegraphics[width=0.13\textwidth]{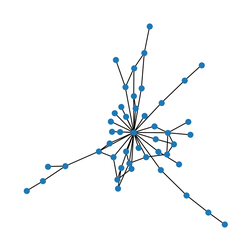} & 
\includegraphics[width=0.13\textwidth]{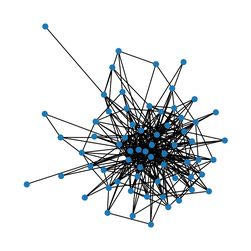} & 
\includegraphics[width=0.13\textwidth]{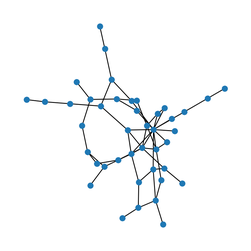} & 
\includegraphics[width=0.13\textwidth]{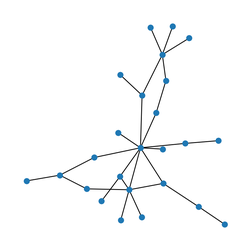} & 
\includegraphics[width=0.13\textwidth]{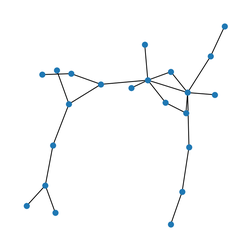} \\
\multicolumn{6}{c}{GDSS}\\
\includegraphics[width=0.13\textwidth]{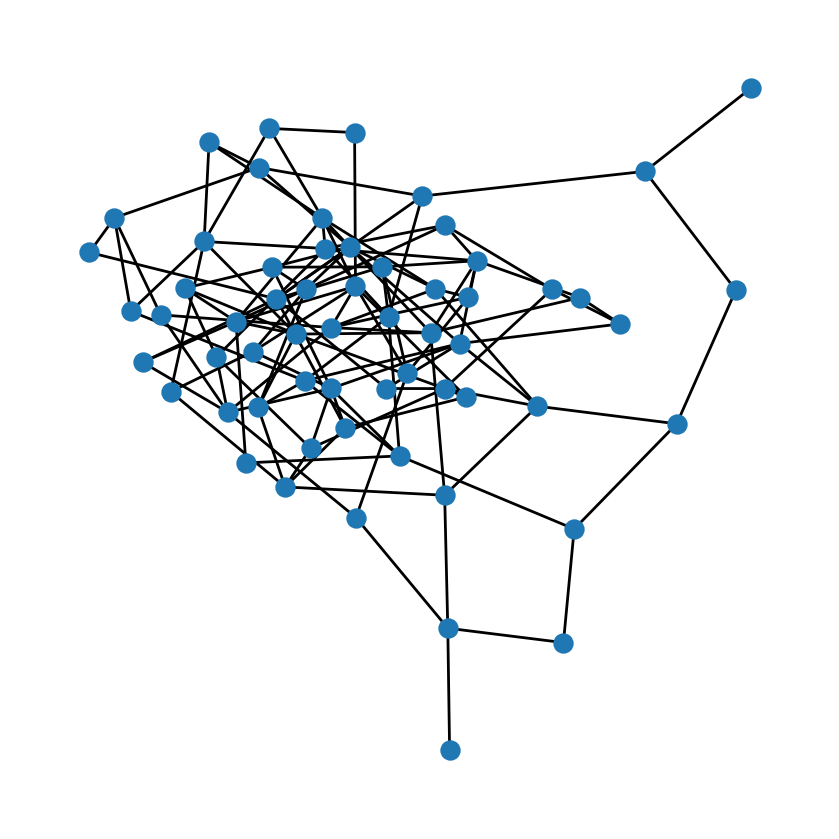} & 
\includegraphics[width=0.13\textwidth]{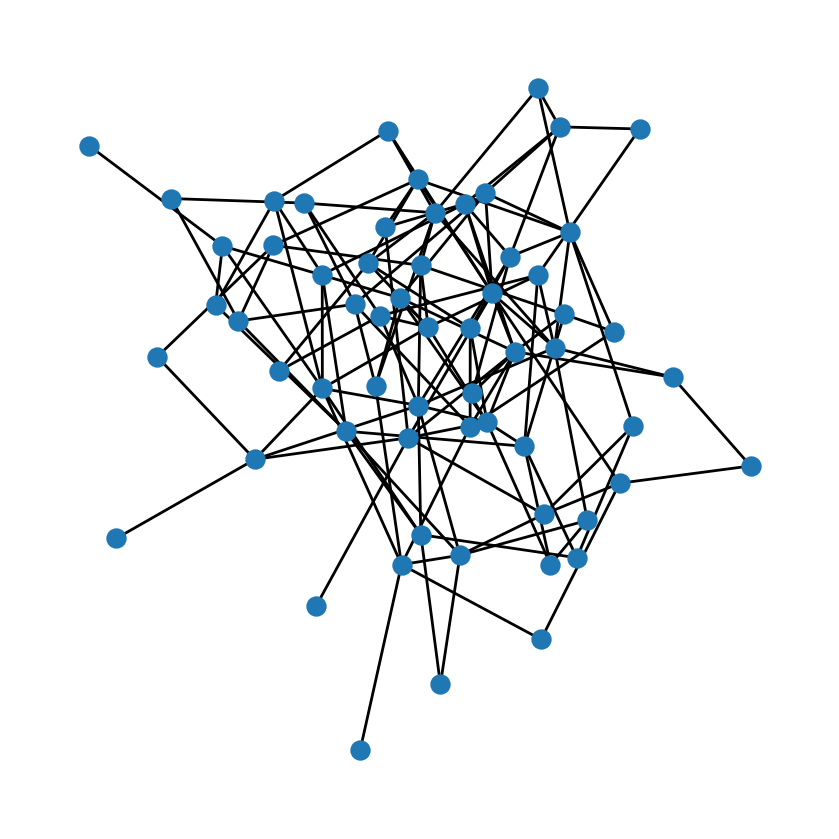} & 
\includegraphics[width=0.13\textwidth]{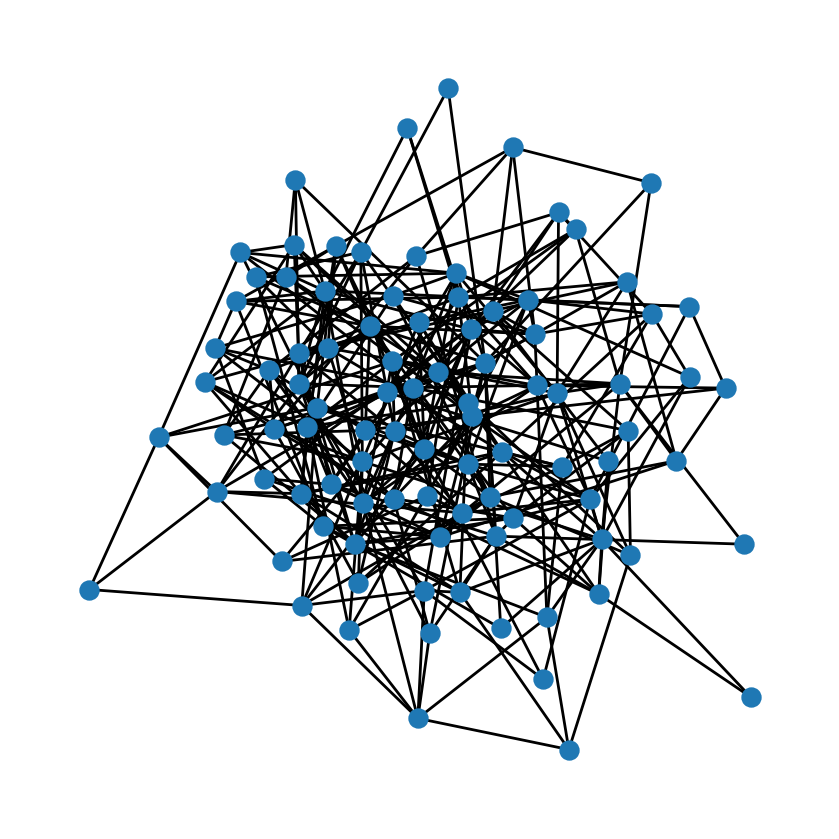} & 
\includegraphics[width=0.13\textwidth]{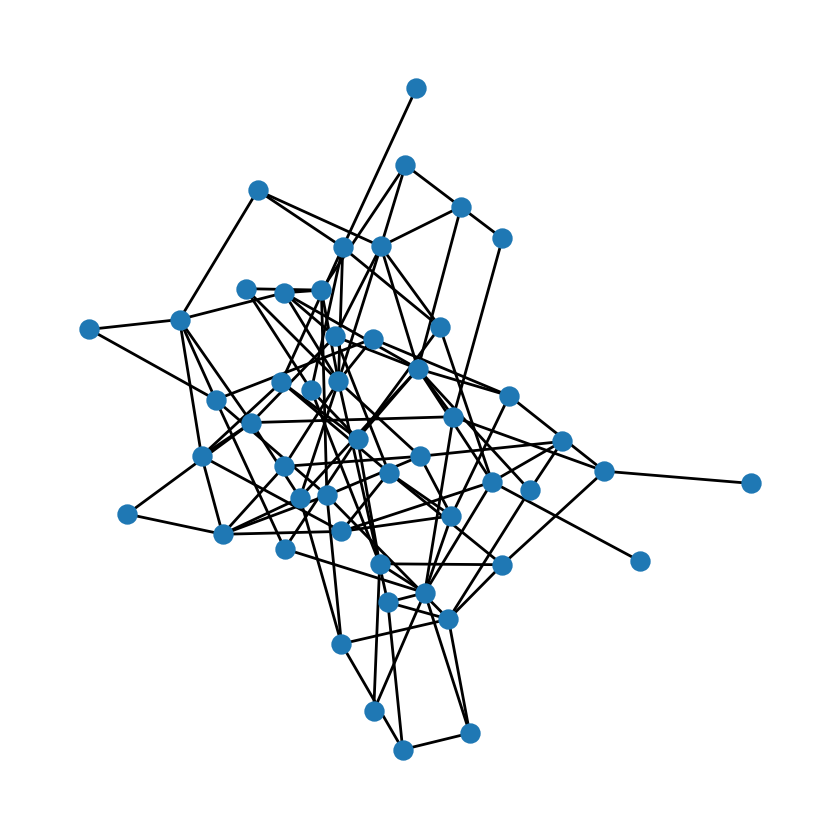} & 
\includegraphics[width=0.13\textwidth]{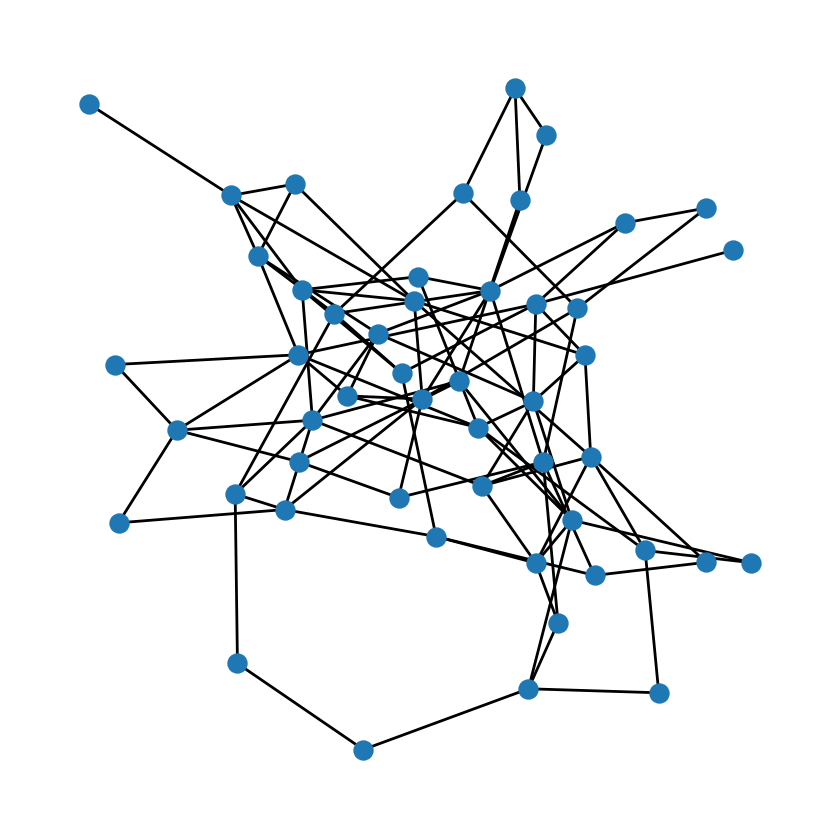} & 
\includegraphics[width=0.13\textwidth]{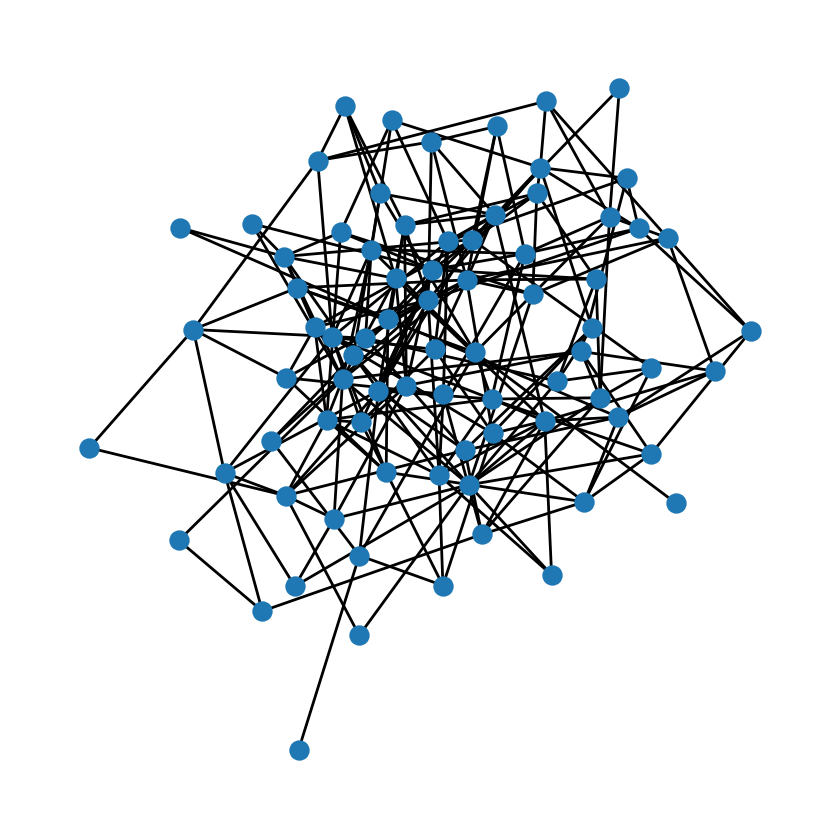}\\
\multicolumn{6}{c}{DiscDDPM}\\
\includegraphics[width=0.13\textwidth]{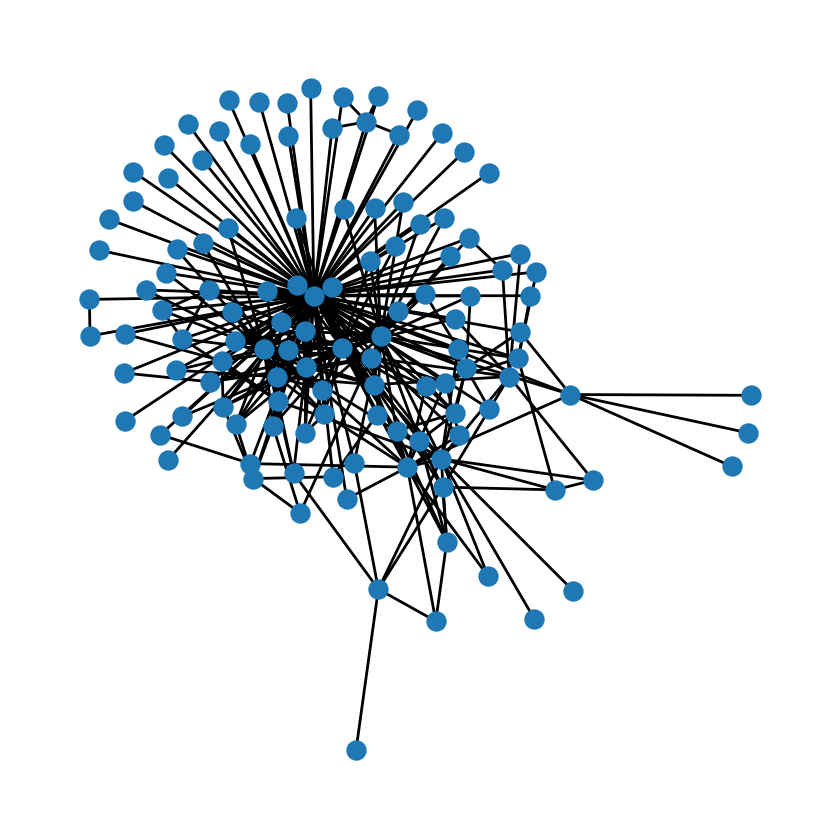} & 
\includegraphics[width=0.13\textwidth]{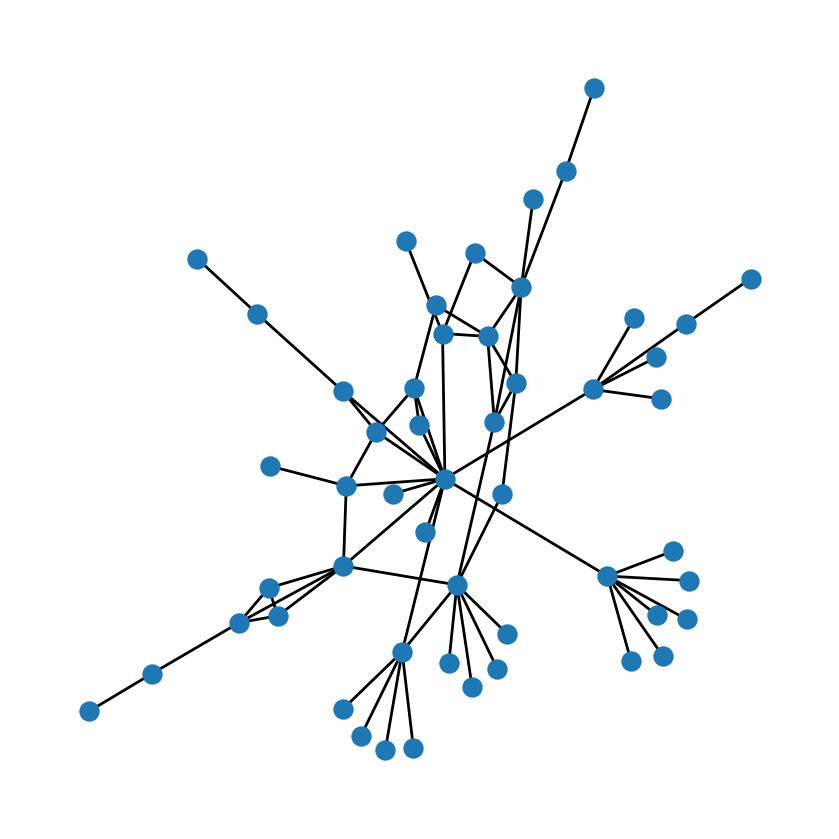} & 
\includegraphics[width=0.13\textwidth]{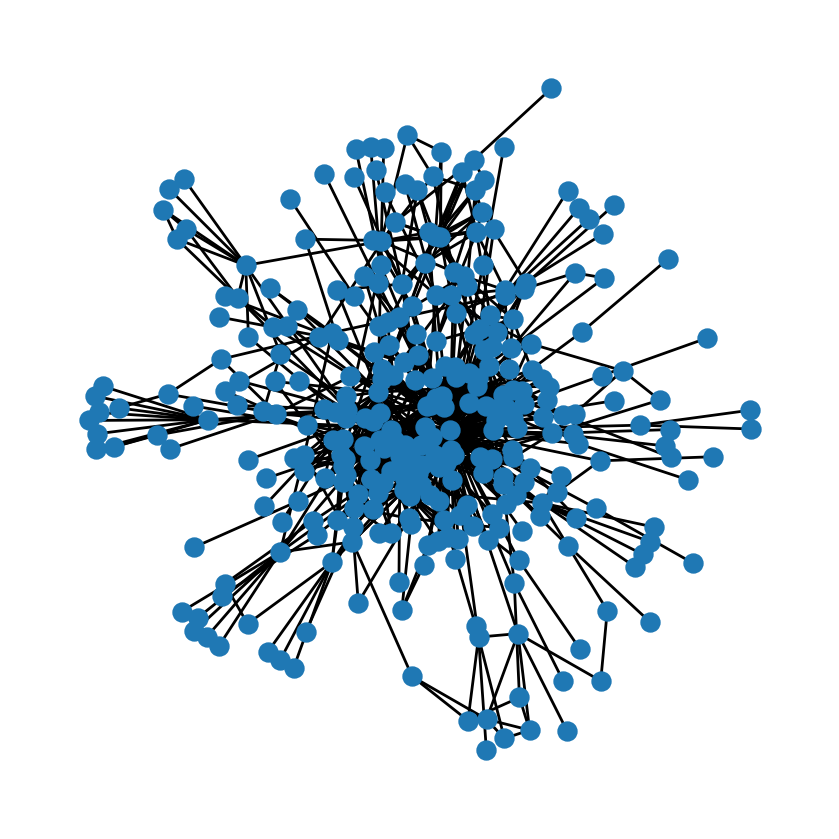} & 
\includegraphics[width=0.13\textwidth]{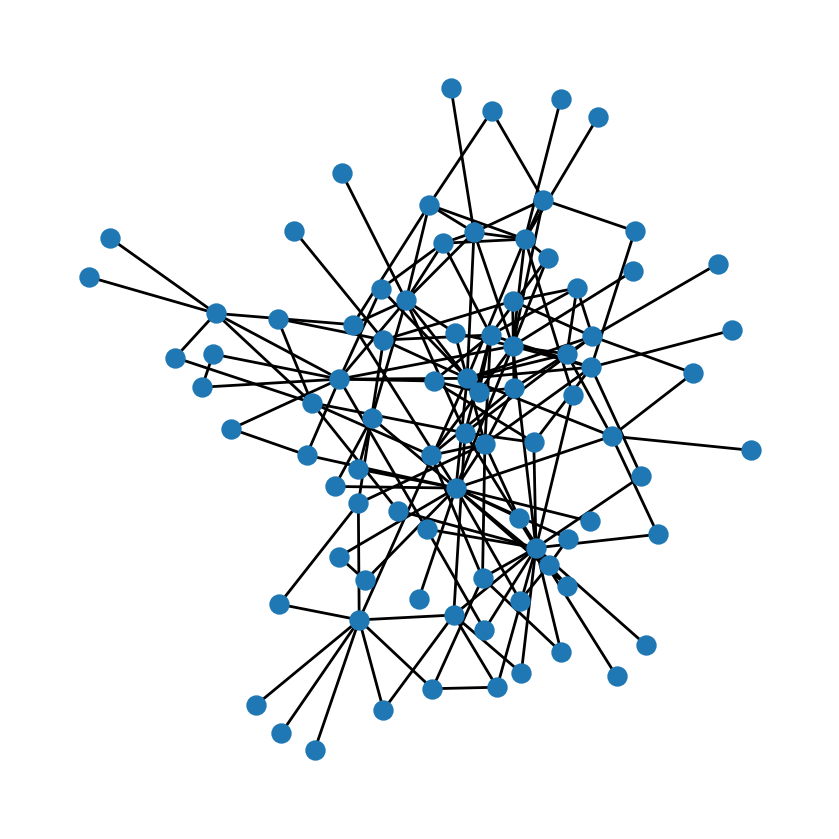} & 
\includegraphics[width=0.13\textwidth]{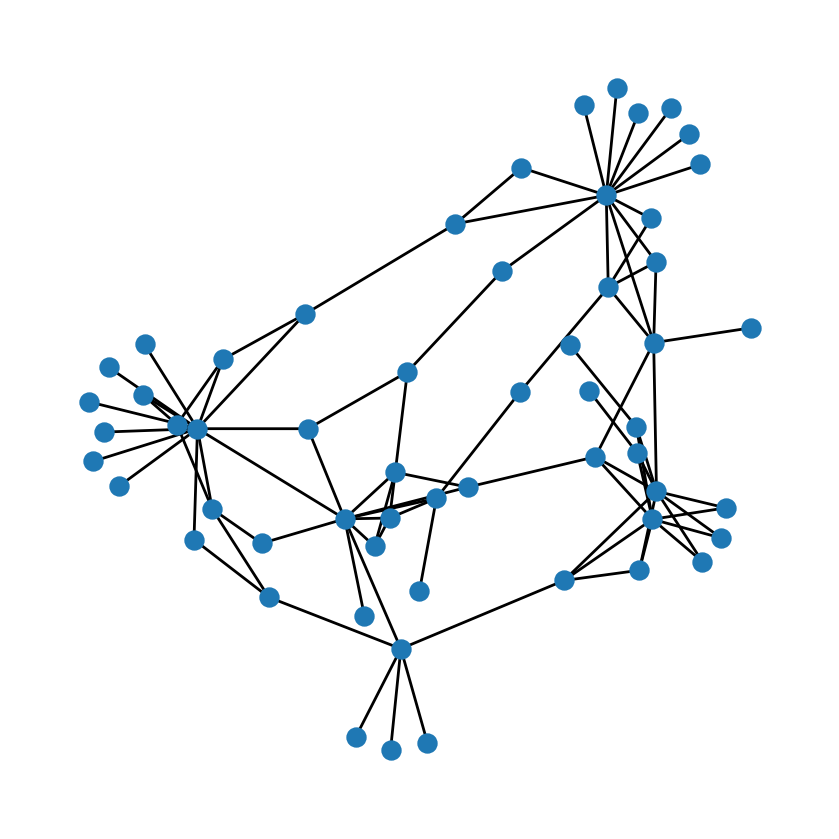} & 
\includegraphics[width=0.13\textwidth]{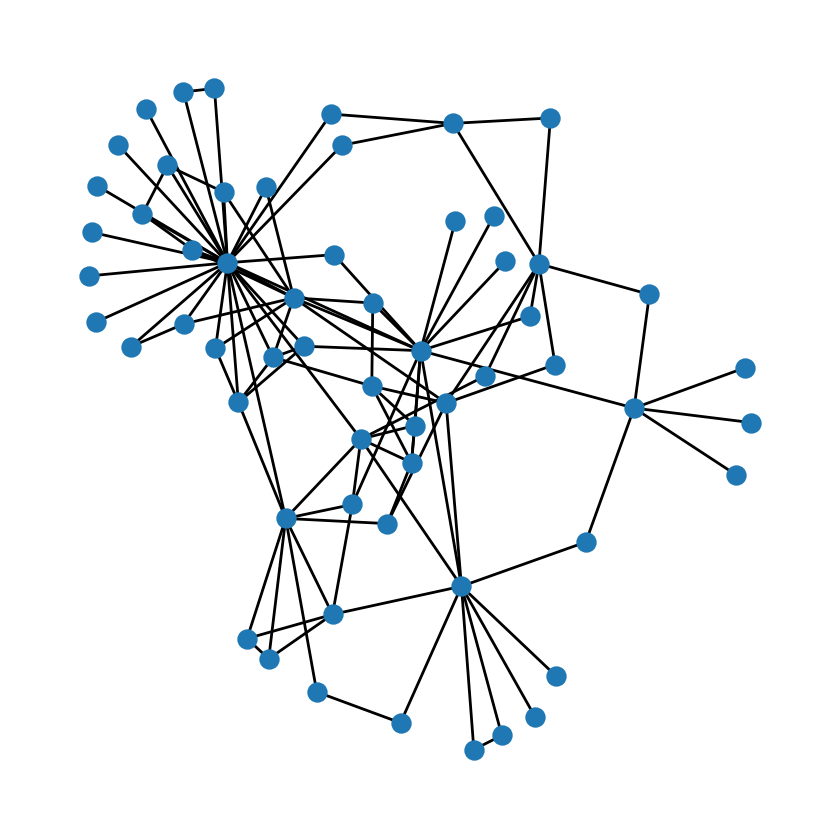}\\
\multicolumn{6}{c}{DiGress}\\
\includegraphics[width=0.13\textwidth]{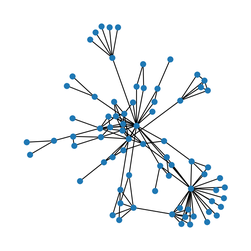} & 
\includegraphics[width=0.13\textwidth]{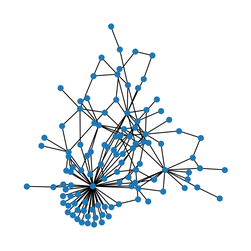} & 
\includegraphics[width=0.13\textwidth]{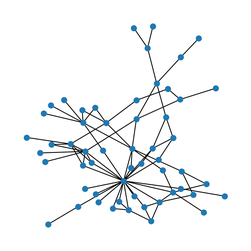} & 
\includegraphics[width=0.13\textwidth]{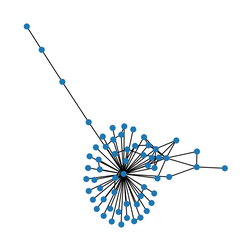} & 
\includegraphics[width=0.13\textwidth]{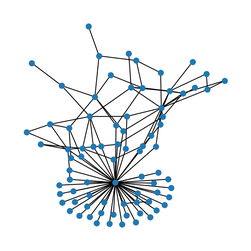} & 
\includegraphics[width=0.13\textwidth]{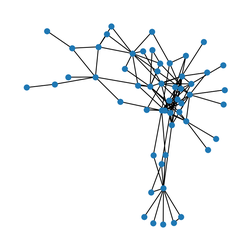} \\

\multicolumn{6}{c}{EDGE}\\
\includegraphics[width=0.13\textwidth]{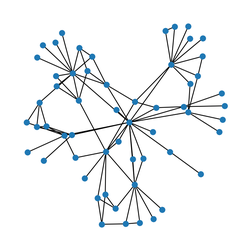} & 
\includegraphics[width=0.13\textwidth]{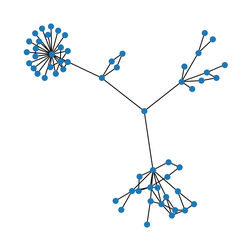} & 
\includegraphics[width=0.13\textwidth]{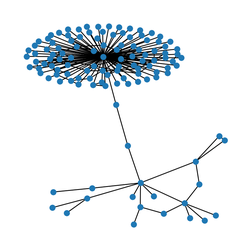} & 
\includegraphics[width=0.13\textwidth]{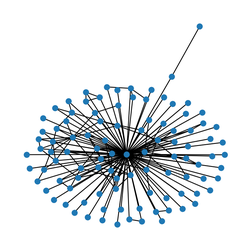} & 
\includegraphics[width=0.13\textwidth]{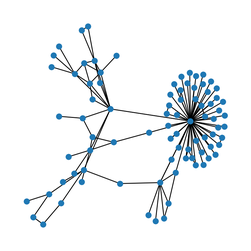} & 
\includegraphics[width=0.13\textwidth]{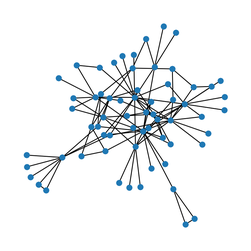} \\
\multicolumn{6}{c}{Test graphs}
    \end{tabular}
    \caption{Visualization of the Ego dataset}
    \label{fig:vis-ego}
\end{figure}

\begin{figure}[h]
    \centering
    \begin{tabular}{cccccc}
\includegraphics[width=0.13\textwidth]{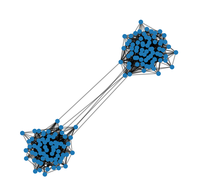} & 
\includegraphics[width=0.13\textwidth]{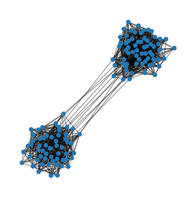} & 
\includegraphics[width=0.13\textwidth]{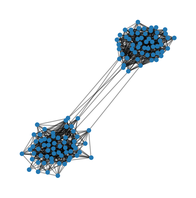} & 
\includegraphics[width=0.13\textwidth]{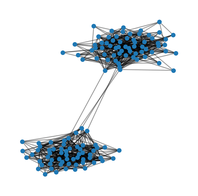} & 
\includegraphics[width=0.13\textwidth]{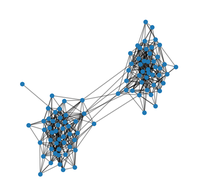} & 
\includegraphics[width=0.13\textwidth]{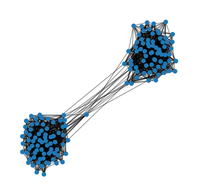} \\
\multicolumn{6}{c}{GRNN}\\

\includegraphics[width=0.13\textwidth]{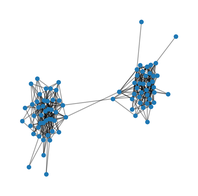} & 
\includegraphics[width=0.13\textwidth]{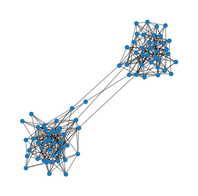} & 
\includegraphics[width=0.13\textwidth]{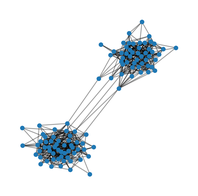} & 
\includegraphics[width=0.13\textwidth]{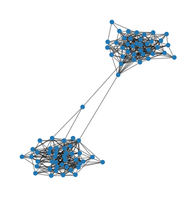} & 
\includegraphics[width=0.13\textwidth]{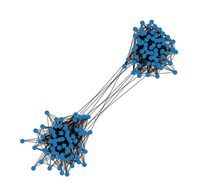} & 
\includegraphics[width=0.13\textwidth]{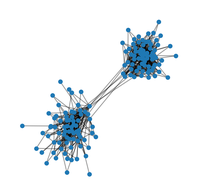} \\
\multicolumn{6}{c}{GRAN}\\

\includegraphics[width=0.13\textwidth]{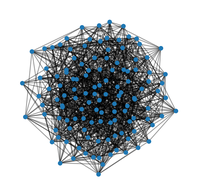} & 
\includegraphics[width=0.13\textwidth]{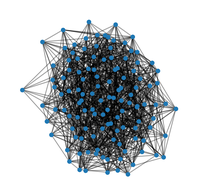} & 
\includegraphics[width=0.13\textwidth]{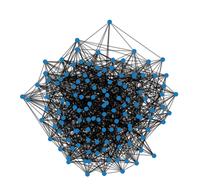} & 
\includegraphics[width=0.13\textwidth]{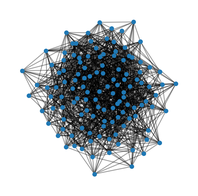} & 
\includegraphics[width=0.13\textwidth]{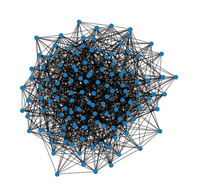} & 
\includegraphics[width=0.13\textwidth]{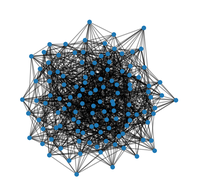} \\
\multicolumn{6}{c}{GraphCNF}\\

\includegraphics[width=0.13\textwidth]{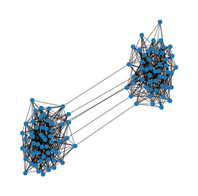} & 
\includegraphics[width=0.13\textwidth]{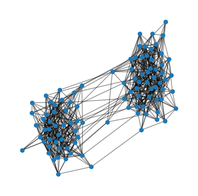} & 
\includegraphics[width=0.13\textwidth]{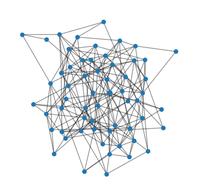} & 
\includegraphics[width=0.13\textwidth]{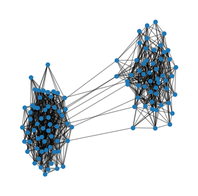} & 
\includegraphics[width=0.13\textwidth]{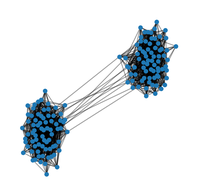} & 
\includegraphics[width=0.13\textwidth]{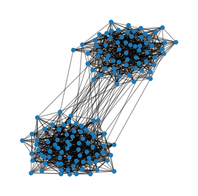} \\
\multicolumn{6}{c}{GDSS}\\
\includegraphics[width=0.13\textwidth]{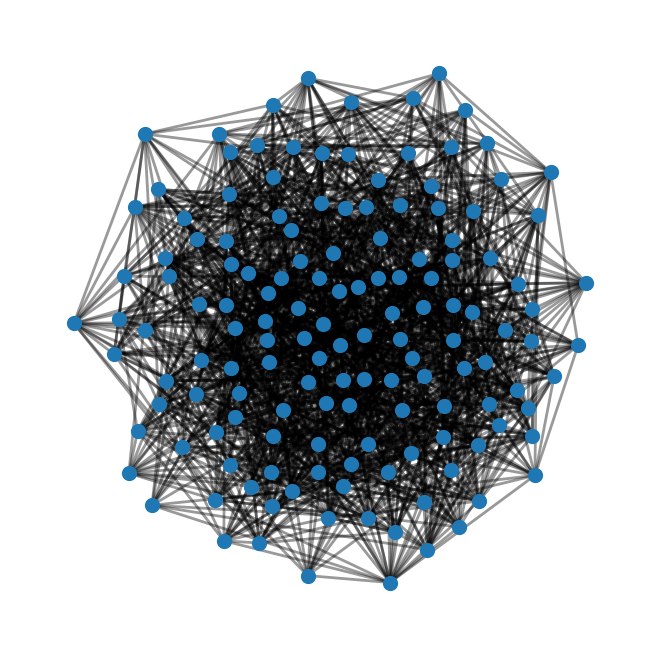} & 
\includegraphics[width=0.13\textwidth]{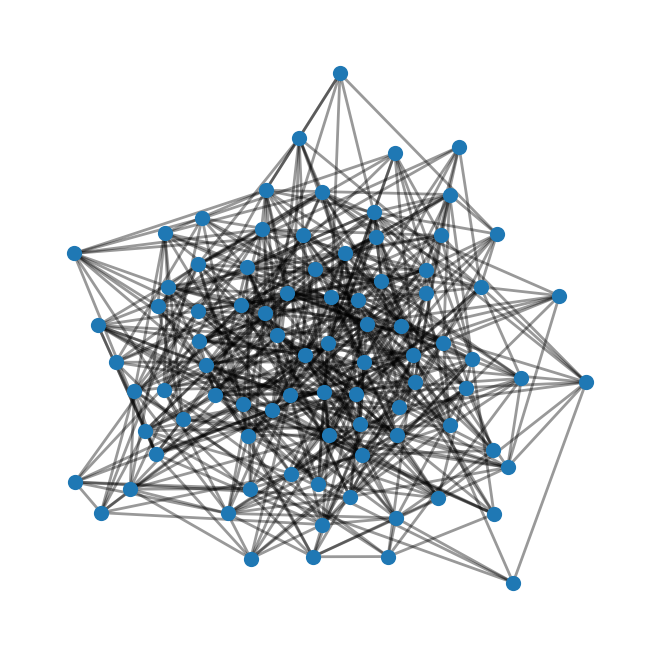} & 
\includegraphics[width=0.13\textwidth]{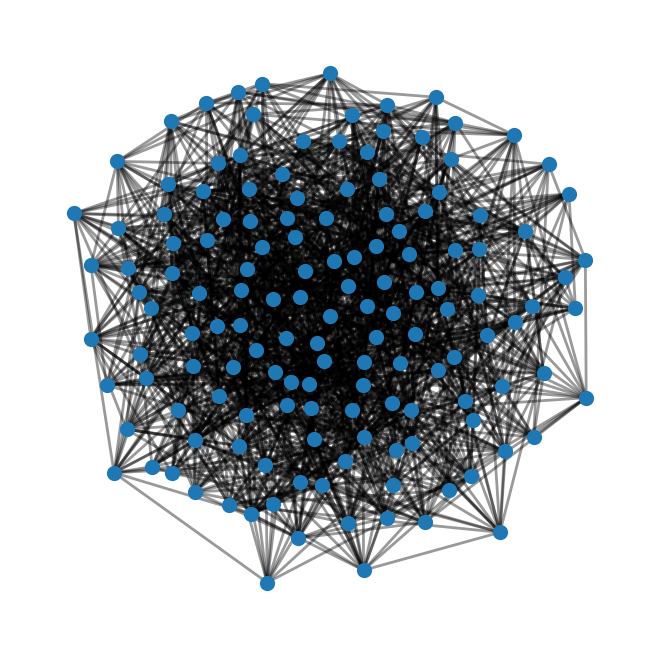} & 
\includegraphics[width=0.13\textwidth]{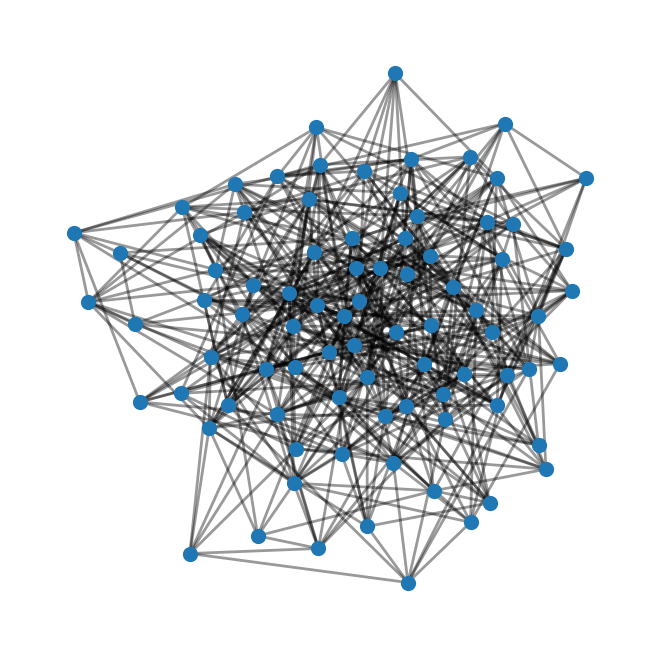} & 
\includegraphics[width=0.13\textwidth]{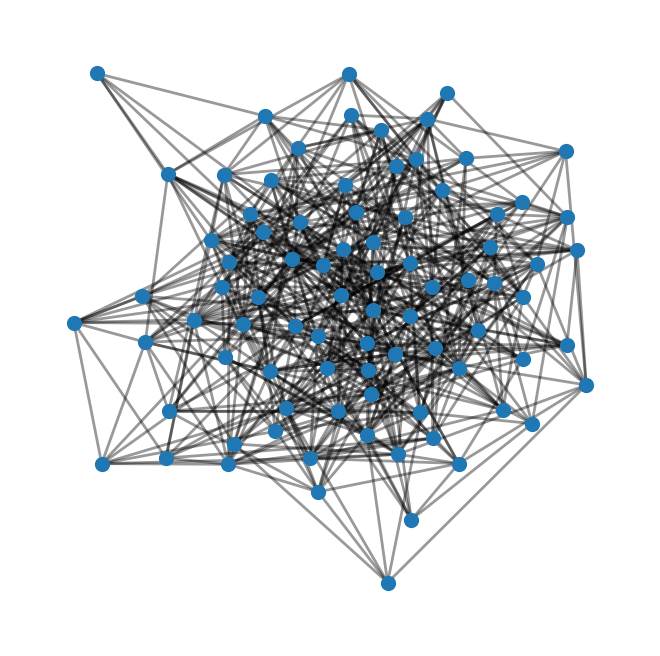} & 
\includegraphics[width=0.13\textwidth]{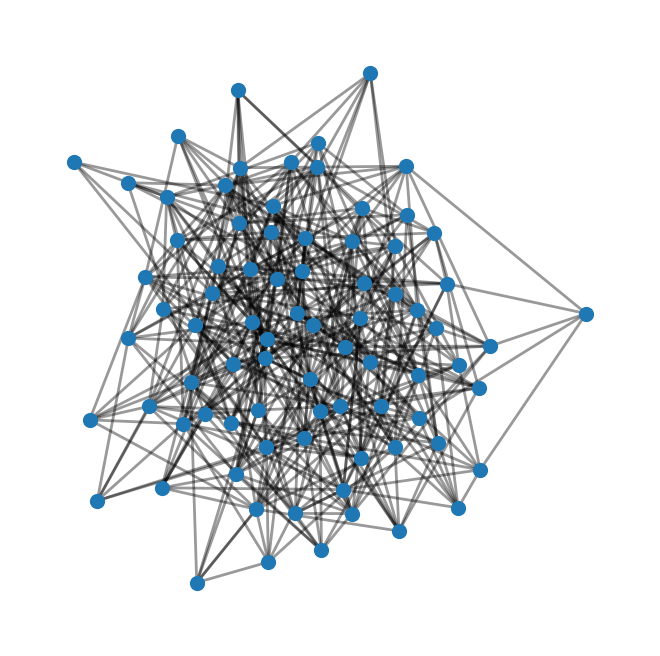}\\
\multicolumn{6}{c}{DiscDDPM}\\
\includegraphics[width=0.13\textwidth]{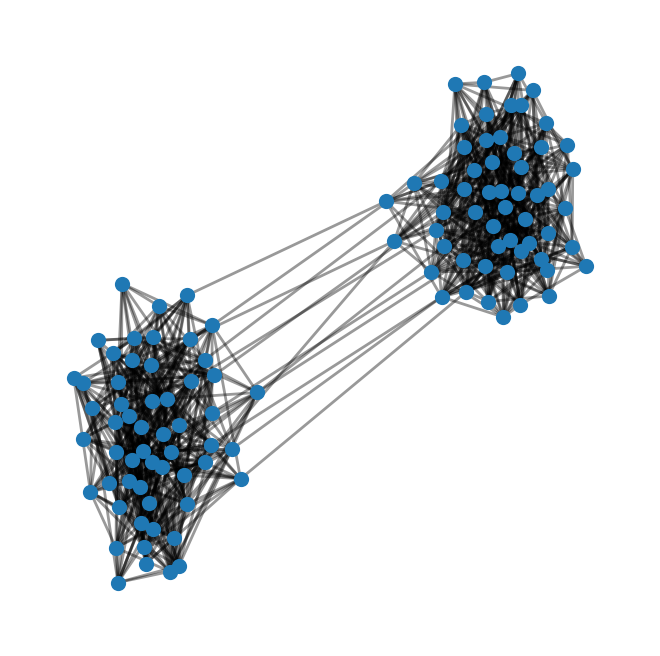} & 
\includegraphics[width=0.13\textwidth]{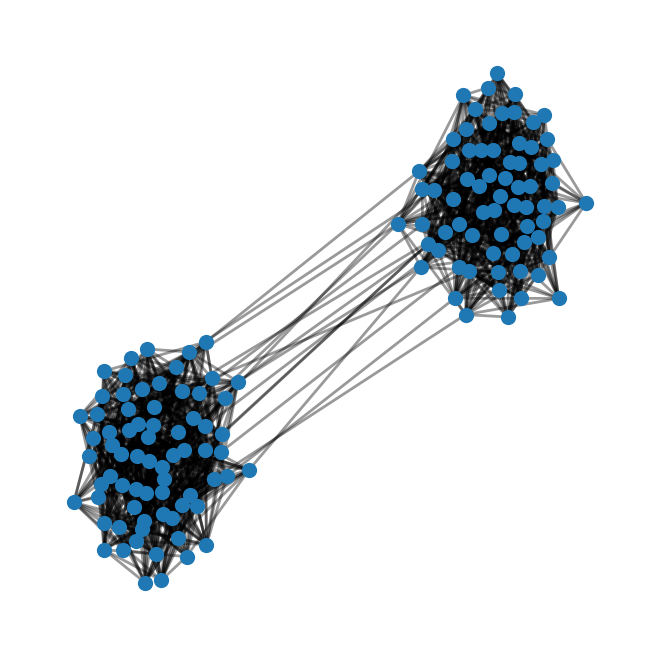} & 
\includegraphics[width=0.13\textwidth]{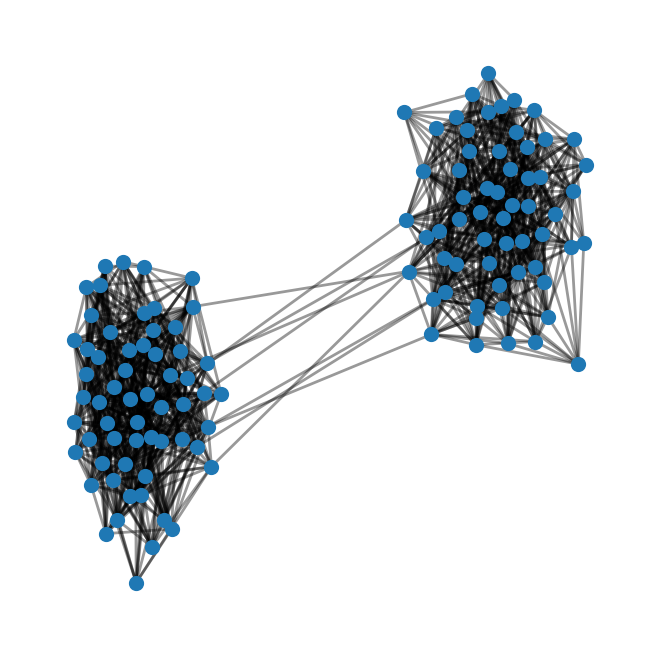} & 
\includegraphics[width=0.13\textwidth]{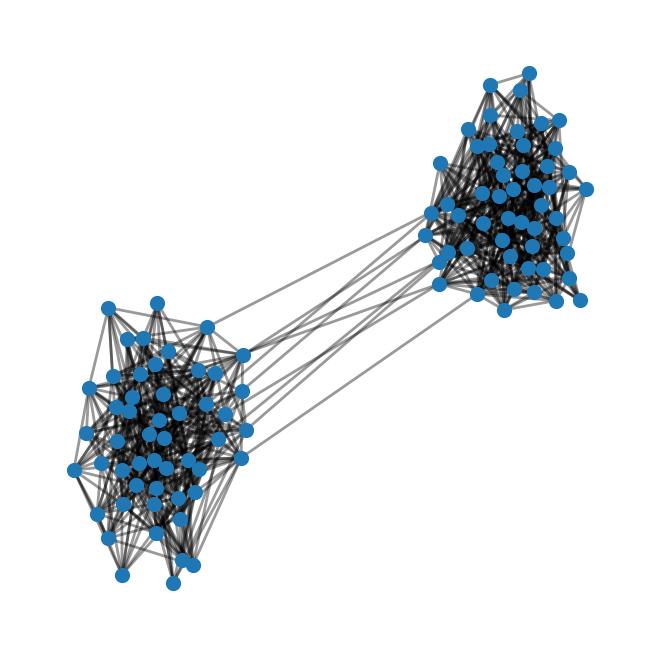} & 
\includegraphics[width=0.13\textwidth]{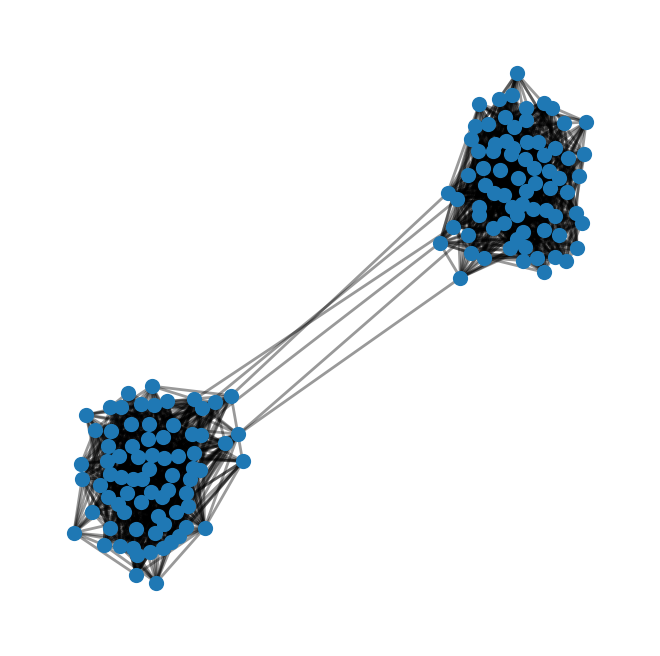} & 
\includegraphics[width=0.13\textwidth]{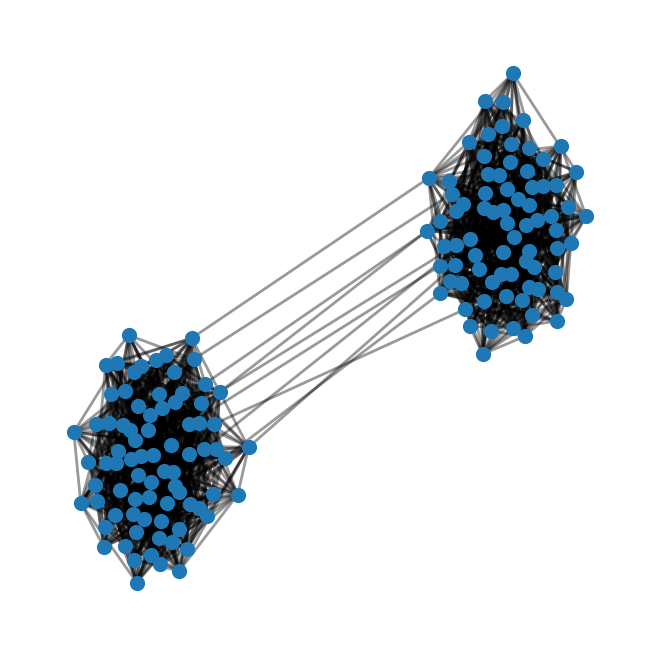}\\
\multicolumn{6}{c}{DiGress}\\
\includegraphics[width=0.13\textwidth]{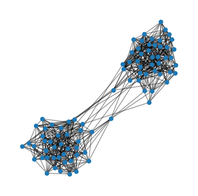} & 
\includegraphics[width=0.13\textwidth]{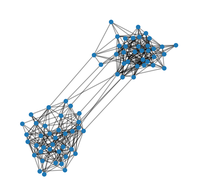} & 
\includegraphics[width=0.13\textwidth]{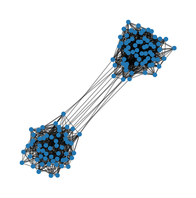} & 
\includegraphics[width=0.13\textwidth]{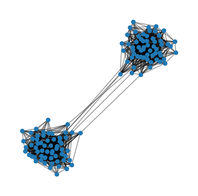}& 
\includegraphics[width=0.13\textwidth]{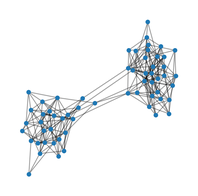} & 
\includegraphics[width=0.13\textwidth]{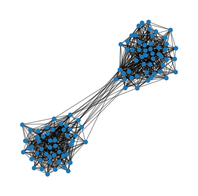} \\
\multicolumn{6}{c}{EDGE}\\
\includegraphics[width=0.13\textwidth]{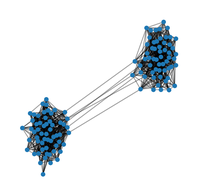} & 
\includegraphics[width=0.13\textwidth]{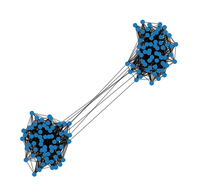} & 
\includegraphics[width=0.13\textwidth]{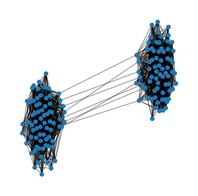} & 
\includegraphics[width=0.13\textwidth]{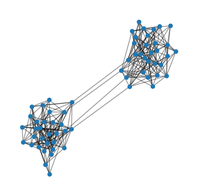} & 
\includegraphics[width=0.13\textwidth]{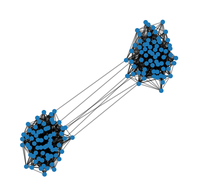} & 
\includegraphics[width=0.13\textwidth]{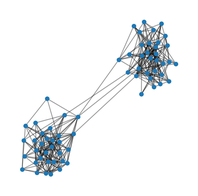} \\
\multicolumn{6}{c}{Test graphs}
    \end{tabular}
    \caption{Visualization of the Community dataset}
    \label{fig:vis-comm}
\end{figure}

\paragraph{Visualization of generated generic graphs.} We visualize six generic graphs from the test data and the generated graphs for each dataset in Figure~\ref{fig:vis-ego} and~\ref{fig:vis-comm}. The visualized graphs are randomly selected from the test data and the generated samples. 

\paragraph{Visualization of generated molecules.} We visualize 16 molecules generated from GDSS, DiGress, and our approach in Figure~\ref{fig:vis-qm9}. 
\begin{figure}[h]
    \centering
    \begin{tabular}{cccccccc}
\includegraphics[width=0.1\textwidth]{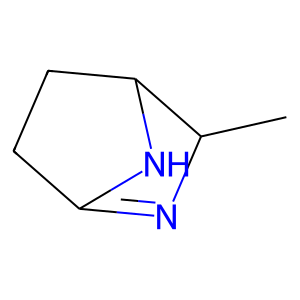} & 
\includegraphics[width=0.1\textwidth]{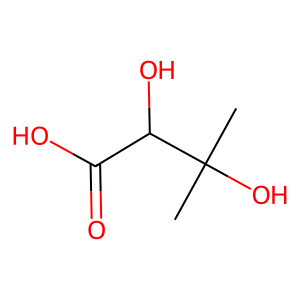} & 
\includegraphics[width=0.1\textwidth]{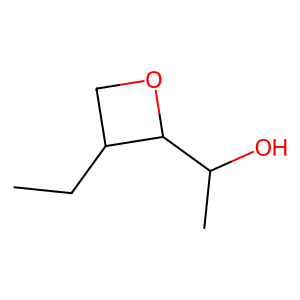} & 
\includegraphics[width=0.1\textwidth]{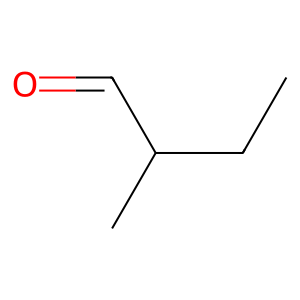} & 
\includegraphics[width=0.1\textwidth]{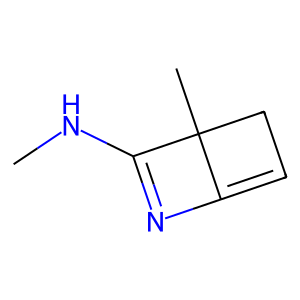} & 
\includegraphics[width=0.1\textwidth]{figures/qm9_gdss/mol_1.png} & 
\includegraphics[width=0.1\textwidth]{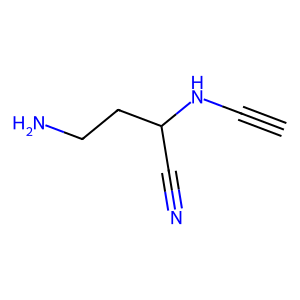} & 
\includegraphics[width=0.1\textwidth]{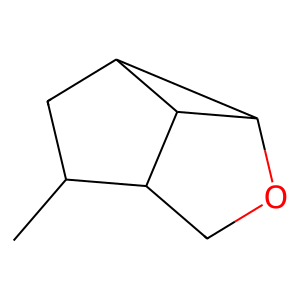} \\
\includegraphics[width=0.1\textwidth]{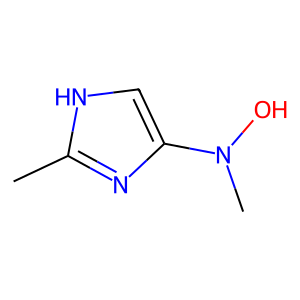} &
\includegraphics[width=0.1\textwidth]{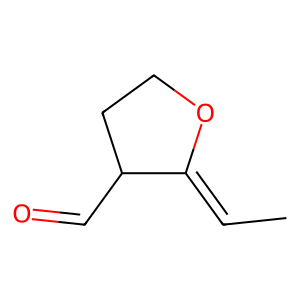} &
\includegraphics[width=0.1\textwidth]{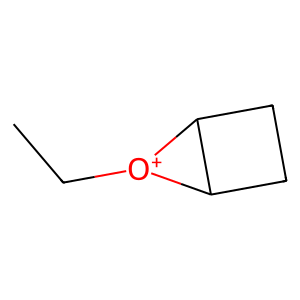} & 
\includegraphics[width=0.1\textwidth]{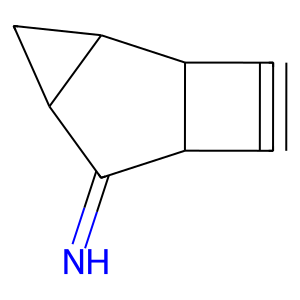} &
\includegraphics[width=0.1\textwidth]{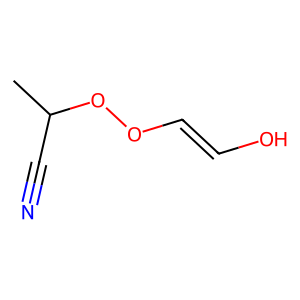} & 
\includegraphics[width=0.1\textwidth]{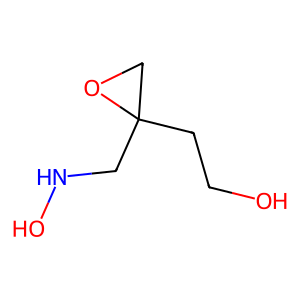} &
\includegraphics[width=0.1\textwidth]{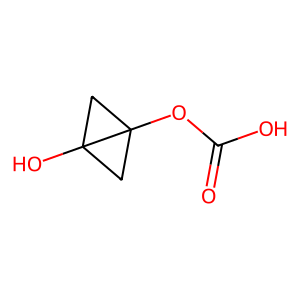} & 
\includegraphics[width=0.1\textwidth]{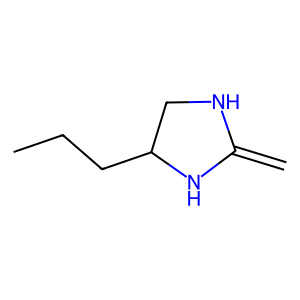} \\
\multicolumn{8}{c}{GDSS}\\\\
\includegraphics[width=0.1\textwidth]{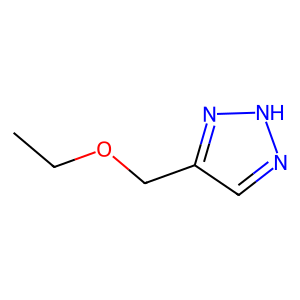} & 
\includegraphics[width=0.1\textwidth]{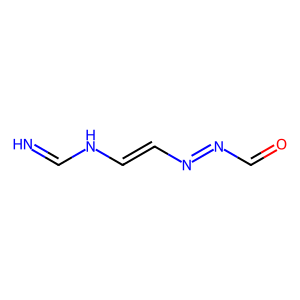} & 
\includegraphics[width=0.1\textwidth]{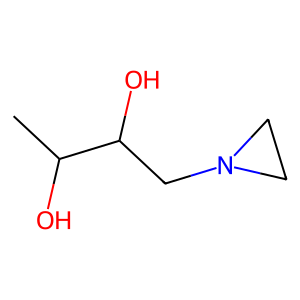} & 
\includegraphics[width=0.1\textwidth]{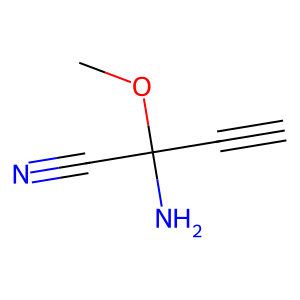} & 
\includegraphics[width=0.1\textwidth]{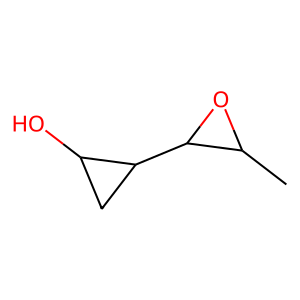} & 
\includegraphics[width=0.1\textwidth]{figures/qm9_digress/mol_1.png} & 
\includegraphics[width=0.1\textwidth]{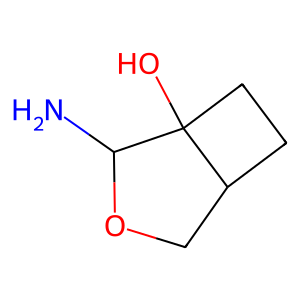} & 
\includegraphics[width=0.1\textwidth]{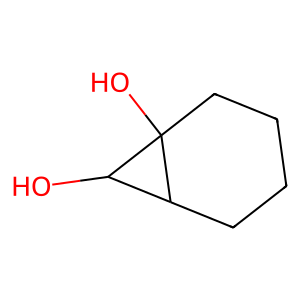} \\
\includegraphics[width=0.1\textwidth]{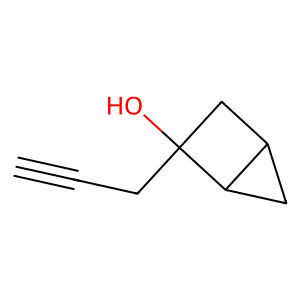} &
\includegraphics[width=0.1\textwidth]{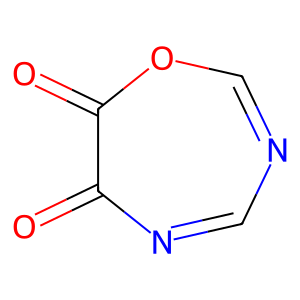} &
\includegraphics[width=0.1\textwidth]{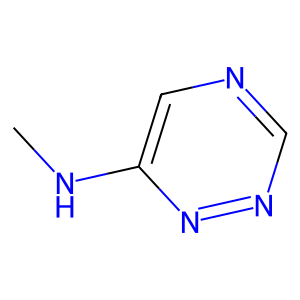} & 
\includegraphics[width=0.1\textwidth]{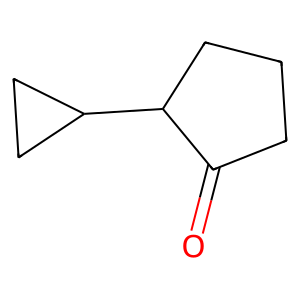} &
\includegraphics[width=0.1\textwidth]{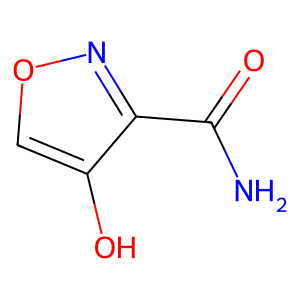} & 
\includegraphics[width=0.1\textwidth]{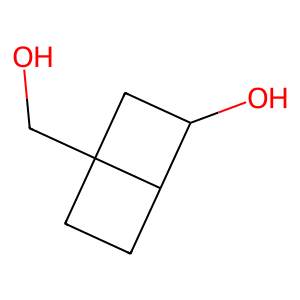} &
\includegraphics[width=0.1\textwidth]{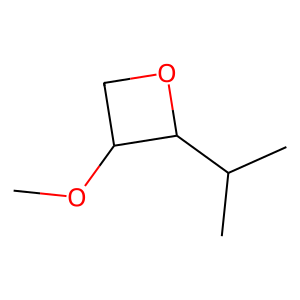} & 
\includegraphics[width=0.1\textwidth]{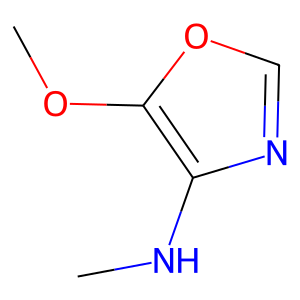} \\
\multicolumn{8}{c}{DiGress}\\\\
\includegraphics[width=0.1\textwidth]{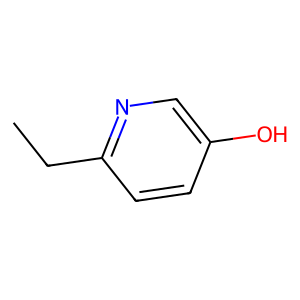} & 
\includegraphics[width=0.1\textwidth]{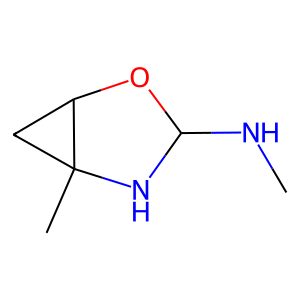} & 
\includegraphics[width=0.1\textwidth]{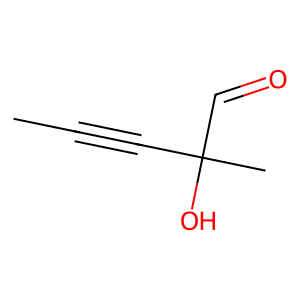} & 
\includegraphics[width=0.1\textwidth]{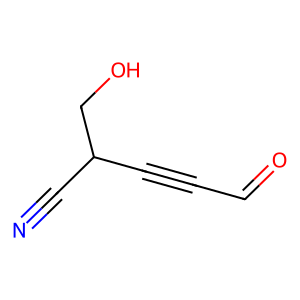} & 
\includegraphics[width=0.1\textwidth]{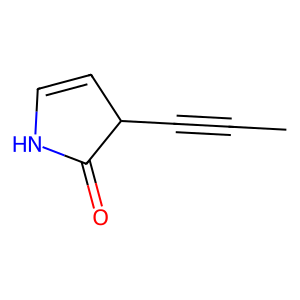} & 
\includegraphics[width=0.1\textwidth]{figures/qm9_edge/mol_1.png} & 
\includegraphics[width=0.1\textwidth]{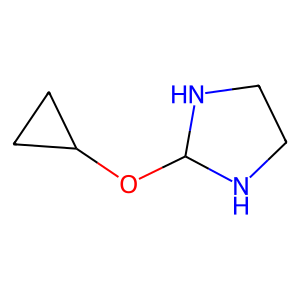} & 
\includegraphics[width=0.1\textwidth]{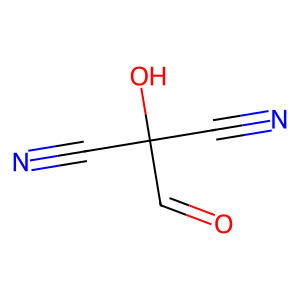} \\
\includegraphics[width=0.1\textwidth]{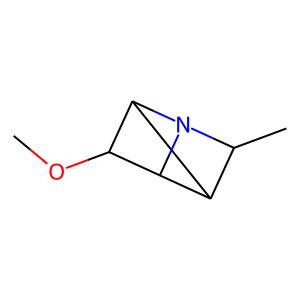} &
\includegraphics[width=0.1\textwidth]{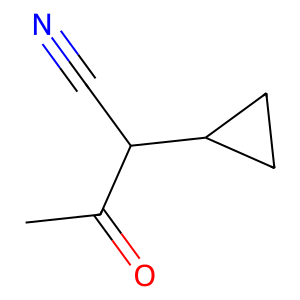} &
\includegraphics[width=0.1\textwidth]{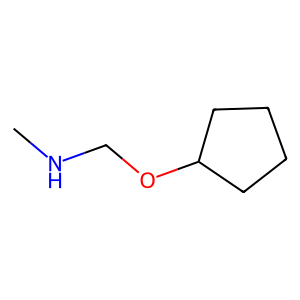} & 
\includegraphics[width=0.1\textwidth]{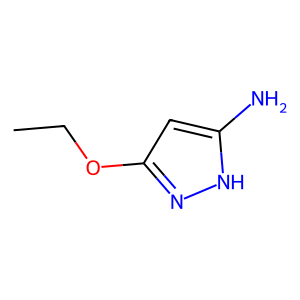} &
\includegraphics[width=0.1\textwidth]{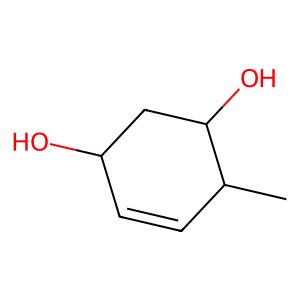} & 
\includegraphics[width=0.1\textwidth]{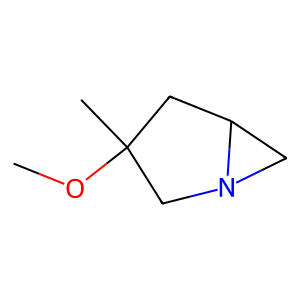} &
\includegraphics[width=0.1\textwidth]{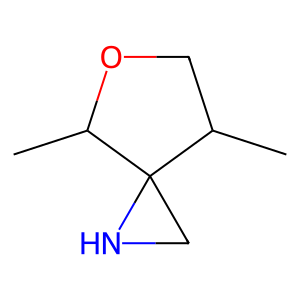} & 
\includegraphics[width=0.1\textwidth]{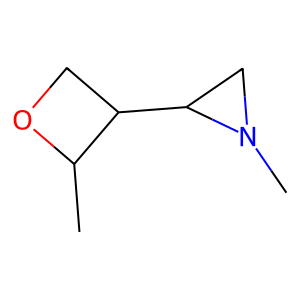} \\
\multicolumn{8}{c}{EDGE}\\\\
\includegraphics[width=0.1\textwidth]{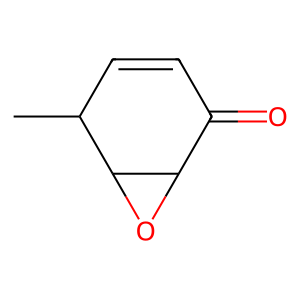} & 
\includegraphics[width=0.1\textwidth]{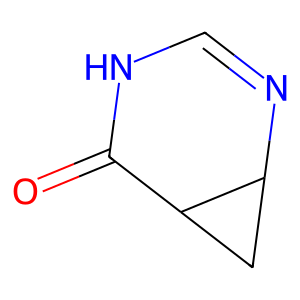} & 
\includegraphics[width=0.1\textwidth]{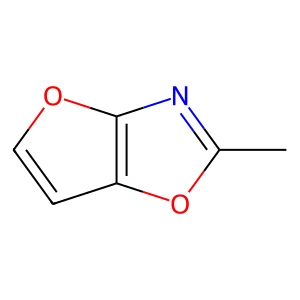} & 
\includegraphics[width=0.1\textwidth]{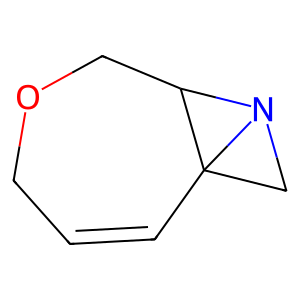} & 
\includegraphics[width=0.1\textwidth]{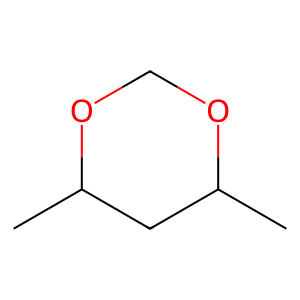} & 
\includegraphics[width=0.1\textwidth]{figures/qm9_gt/mol_1.png} & 
\includegraphics[width=0.1\textwidth]{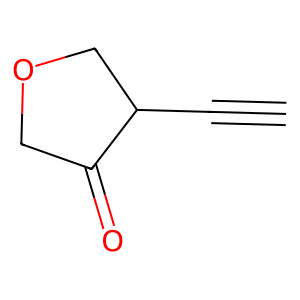} & 
\includegraphics[width=0.1\textwidth]{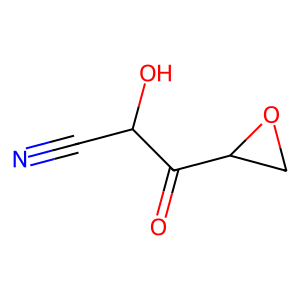} \\
\includegraphics[width=0.1\textwidth]{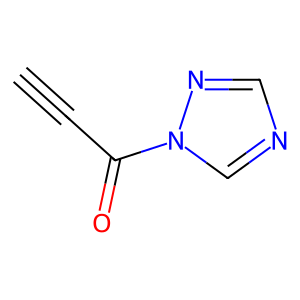} &
\includegraphics[width=0.1\textwidth]{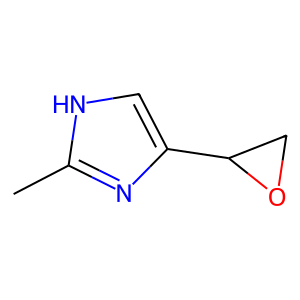} &
\includegraphics[width=0.1\textwidth]{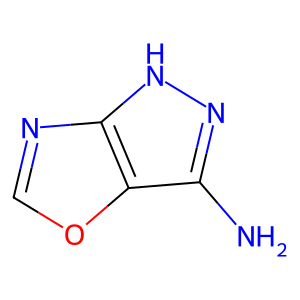} & 
\includegraphics[width=0.1\textwidth]{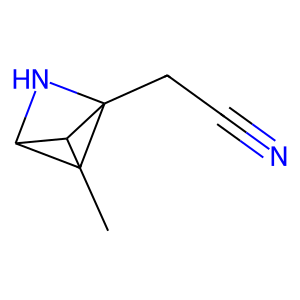} &
\includegraphics[width=0.1\textwidth]{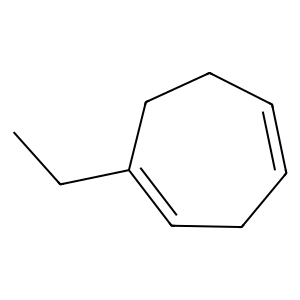} & 
\includegraphics[width=0.1\textwidth]{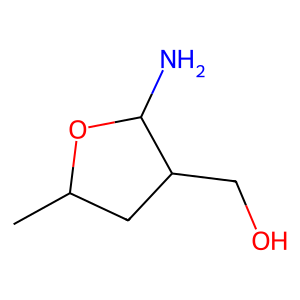} &
\includegraphics[width=0.1\textwidth]{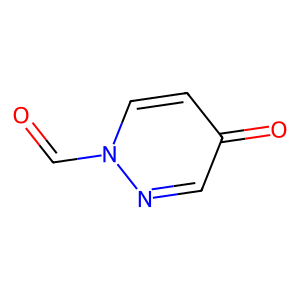} & 
\includegraphics[width=0.1\textwidth]{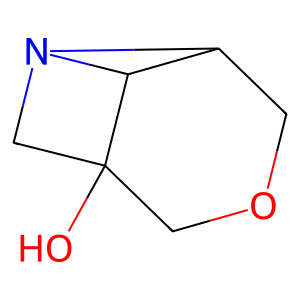} \\
\multicolumn{8}{c}{Test molecules}\\
    \end{tabular}
    \caption{Visualization of the QM9 dataset}
    \label{fig:vis-qm9}
\end{figure}




\end{document}